\documentclass[lettersize,journal]{IEEEtran}
\usepackage{amsmath,amsfonts}
\usepackage{algorithmic}
\usepackage{algorithm}
\usepackage{array}
\usepackage{textcomp}
\usepackage{stfloats}
\usepackage{url}
\usepackage{verbatim}
\usepackage{graphicx}
\usepackage{cite}
\usepackage{float}
\usepackage{booktabs} 
\usepackage{longtable}
\usepackage{stfloats}
\usepackage{xspace}
\usepackage{tikz}
\usepackage{flushend}
\usepackage{hyperref}

\hypersetup{
    colorlinks=true, 
    linkcolor=red,   
    filecolor=magenta, 
    urlcolor=blue,    
}

\definecolor{wood}{RGB}{0,139,0}
\definecolor{veget}{RGB}{0,255,0}
\definecolor{bare}{RGB}{139,0,0}
\definecolor{indus}{RGB}{255,0,0}
\definecolor{resid}{RGB}{205,173,0}
\definecolor{road}{RGB}{83,134,139}
\definecolor{paddy}{RGB}{0,139,139}
\definecolor{planting}{RGB}{139,105,20}
\definecolor{human}{RGB}{189,183,107}

\usepackage{subfigure}
\begin{document}

\title{
ClassWise-SAM-Adapter: Parameter Efficient Fine-tuning Adapts Segment Anything to SAR Domain for Semantic Segmentation
}

\author{Xinyang Pu,~\IEEEmembership{Graduate Student Member,~IEEE},~Hecheng Jia,~\IEEEmembership{Member,~IEEE},~Linghao Zheng,~\IEEEmembership{Graduate Student Member,~IEEE},~Feng Wang,~\IEEEmembership{Member,~IEEE},~Feng Xu,~\IEEEmembership{Senior Member,~IEEE}

\thanks{This work was supported in part by the Natural Science Foundation of China under Grant 61991421 and 61991422. (Corresponding author: Feng Wang and Feng Xu).

The authors are with the Key Laboratory for Information Science of Electromagnetic Waves(Ministry of Education), School of Information Science and Technology, Fudan University, Shanghai 200433, China. (fengwang@fudan.edu.cn, fengxu@fudan.edu.cn)}
\thanks{Manuscript received Nov 11, 2023; revised ***.}}

\markboth{Journal of \LaTeX\ Class Files,~Vol.~14, No.~8, October~2023}%
{Shell \MakeLowercase{\textit{et al.}}: A Sample Article Using IEEEtran.cls for IEEE Journals}

\IEEEpubid{0000--0000/00\$00.00~\copyright~2021 IEEE}
 
\maketitle

\begin{abstract}
In the realm of artificial intelligence, the emergence of foundation models, backed by high computing capabilities and extensive data, has been revolutionary. Segment Anything Model (SAM), built on the Vision Transformer (ViT) model with millions of parameters and vast training dataset SA-1B, excels in various segmentation scenarios relying on its significance of semantic information and generalization ability. Such achievement of visual foundation model stimulates continuous researches on specific downstream tasks in computer vision. The ClassWise-SAM-Adapter (CWSAM) is designed to adapt the high-performing SAM for landcover classification on space-borne Synthetic Aperture Radar (SAR) images. The proposed CWSAM freezes most of SAM's parameters and incorporates lightweight adapters for parameter efficient fine-tuning, and a classwise mask decoder is designed to achieve semantic segmentation task. This adapt-tuning method allows for efficient landcover classification of SAR images, balancing the accuracy with computational demand. In addition, the task specific input module injects low frequency information of SAR images by MLP-based layers to improve the model performance. Compared to conventional state-of-the-art semantic segmentation algorithms by extensive experiments, CWSAM showcases enhanced performance with fewer computing resources, highlighting the potential of leveraging foundational models like SAM for specific downstream tasks in the SAR domain. The source code is available at: \url{https://github.com/xypu98/CWSAM}.
\end{abstract}


\section{Introduction}
\IEEEPARstart{T}{he} emergence of foundation models in natural language processing and computer vision trigger the developments of universal architectures on extensive algorithms of various downstream tasks with high uniformity and strong generalization capabilities. Thanks to the release of Segment Anything Model (SAM) \cite{sam}, the progressions of foundation models in computer vision and the involved researches are booming, which bring available future of uniform large model in computer vision because of its zero-shot ability on natural scene segmentation task. SAM has impressive segmentation performance in extensive application fields benefiting from its elaborate architectures and massive training data (SA-1B) \cite{sam} with billions of mask label.
\par Although the great generalization and zero-shot ability on most of scenes show prominent advantages of SAM as a visual foundation model, the complicated requirements of downstream tasks and multiple application fields raise more complex requirements and further optimizations of SAM. Especially for Synthetic Aperture Radar (SAR) images \cite{xuSAR} \cite{DLmeetSAR}, the unique imaging mechanism of electromagnetic wave on SAR induces distinctive image characteristics that are inexperienced and indecipherable for SAM, and deteriorates the effectiveness and reliability of SAM on SAR images semantic segmentation task as a consequence. On the other hand, due to the scarce opportunity and high cost of accessing SAR images, it is difficult to construct a SAR foundation model from scratch with limited SAR image data resources. Utilizing the Segment Anything Model (SAM) for the SAR image interpretation presents a possibility worth exploring. However, it comes with challenges that we summarize to the following two main issues:
\begin{enumerate}
\item{SAM is hardly able to segment the SAR images for reasonable accuracy even though it has good generalization capability on natural images including optical aerial images as shown in Figure 1.a. Considering the electromagnetic wave imaging mechanism of signal echo on synthetic aperture radar, the appearances of SAR images are very different with natural scene images. To be specific, SAR images exhibit coherent speckle noise due to wave physics, side lobes and the complex scattering characteristics that can introduce false alarms, and layover and shadowing effects in regions with steep terrain or tall structures.}
\item{The category-agnostic property of SAM hinders its application on diverse visual downstream tasks. For semantic segmentation and landcover classification tasks on SAR images, \IEEEpubidadjcol each pixel needs a predicted category while SAM only generates several binary mask lack of classwise information under the guidance of prompt input. The segment results of SAR images from SAM represent almost no meaningful semantic information as shown in Figure 1.b because pixels in the same category are not classified into the same category. In addition, landcover classification task demand the model to output a multi-class predicted mask for each image while the quantity of output masks by SAM is variable.}
\end{enumerate}
\begin{figure}[htbp]
\subfigure[Optical aerial images with predicted masks]{
	\centering
    \begin{minipage}[t]{0.9\linewidth}
	   \centering
	   \includegraphics[width=.45\columnwidth]{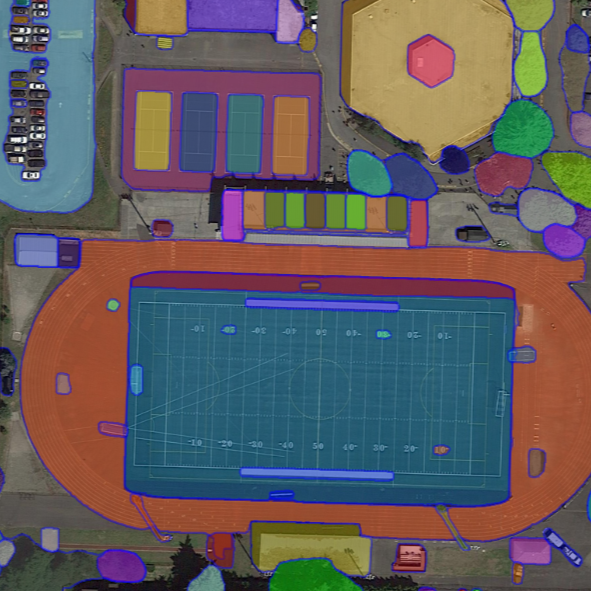}\hspace{3pt}
	   \includegraphics[width=.45\columnwidth]{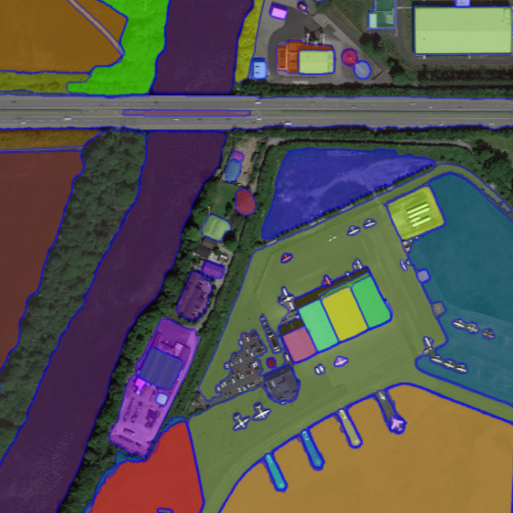}
	\end{minipage}
 }
 \subfigure[SAR images with predicted masks]{
    \begin{minipage}[t]{0.9\linewidth}
    \centering
	   \includegraphics[width=.45\columnwidth]{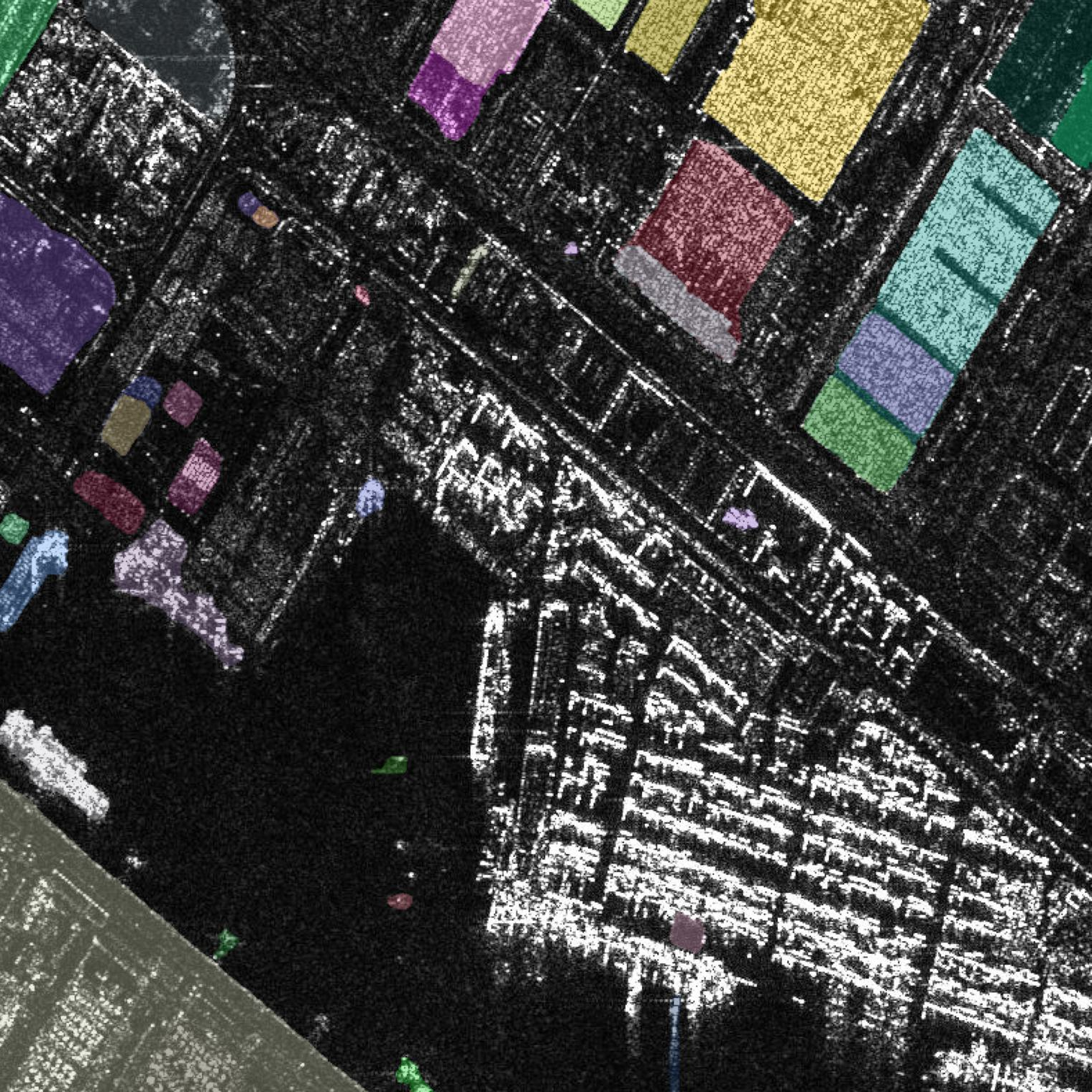}\hspace{3pt}
	   \includegraphics[width=.45\columnwidth]{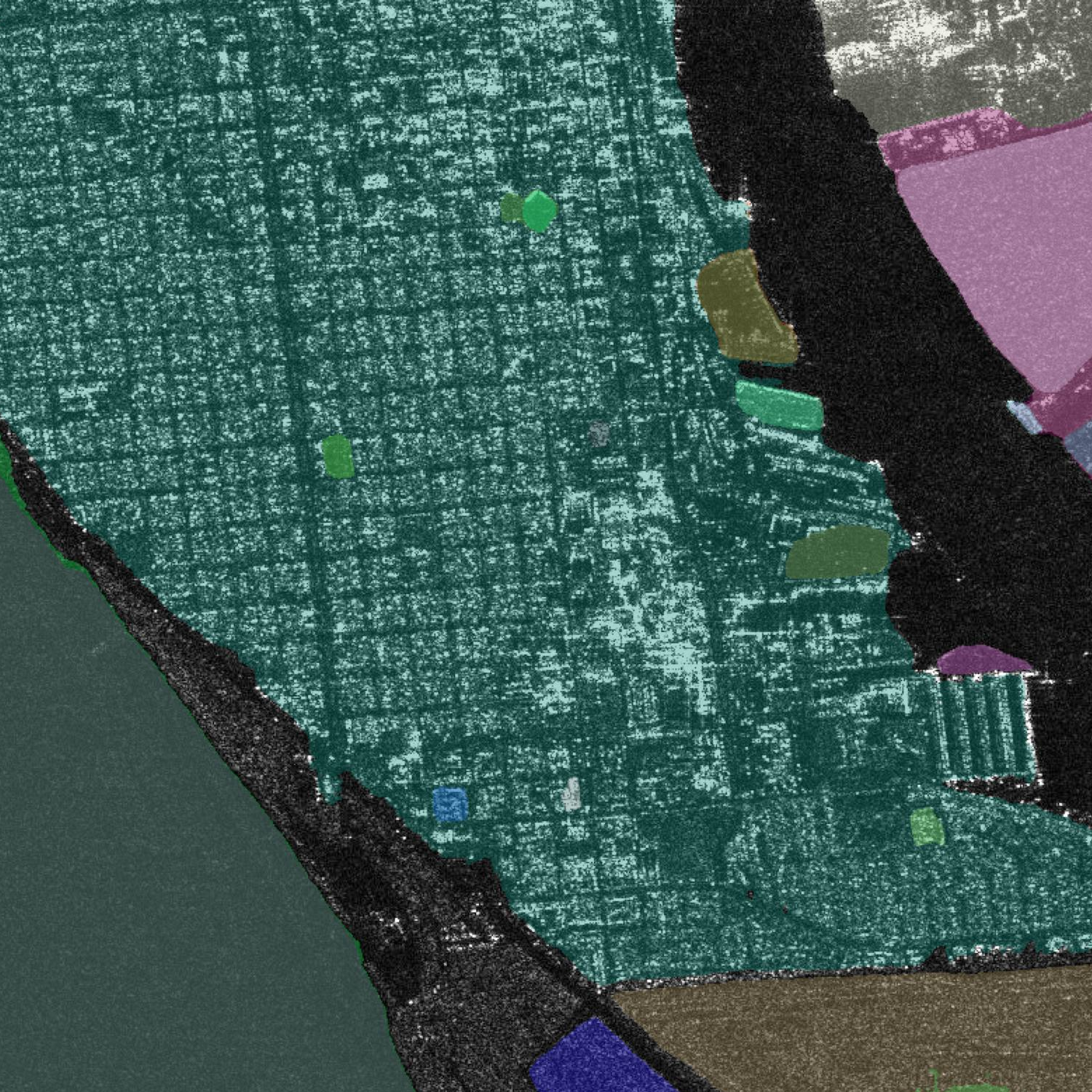}
	\end{minipage}
 }
	   \caption{The output masks of remote sensing images by Segment Anything}
\end{figure}
\par Therefore, we propose a method named ClassWise-SAM-Adapter (CWSAM) to achieve reasonable performance on pixel-level SAR landcover classification task by leveraging foundation model–Segment Anything Model and adding a classwise light-weighted mask decoder. A well-designed adapter fine-tuning mechanism is utilized. Freezing most parameters of the heavy Vision Transformer (ViT) \cite{ViT} image encoder of SAM and inserting a small number of trainable parameters by adapters, CWSAM shifts the natural scene domain of SAM to SAR domain, and explores the potential application of large model in SAR downstream tasks such as landcover classification. Since only limited parameters are trainable with gradient in the model weight, CWSAM demands low computing resources and makes training process efficient. 

We construct CWSAM as a preliminary attempt of visual foundation model application in SAR images and believe that such effective methods with viable computing cost will bring good future on the development of foundation model to SAR intelligent interpretation. Extensive experiments illustrate that the efficiency and reliability of the proposed method on both segmentation performance and training process. CWSAM outperforms multiple state-of-the-art semantic segmentation algorithms with fewer trainable parameters and less memory consumption. To our best knowledge, it is the first time that SAM is introduced for parameter efficient fine-tuning on SAR domain to realize semantic segmentation downstream task with category-aware ability.
\par In summary, the contributions of this paper can be summarized as follows.
\begin{enumerate}
\item{The proposed CWSAM introduces an end-to-end architecture with light-weighted adapters for efficient parameter fine-tuning on large models. The method transfers the natural scene domain of SAM to the SAR domain. By leveraging the visual foundation model, it achieves reliable performance on landcover classification tasks using SAR images.}
\item{A classwise mask decoder of CWSAM is designed to create classwise attribute to original category-agnostic SAM model for fine-grained semantic segmentation downstream task. For SAR images, the classwise mask decoder is utilized to identify multiple landcover categories on pixel level.}
\item{The task specific input module of SAR image low frequency information is conducted to provide sufficient semantic information of landcover characteristics through 2D image Fast Fourier transform for enhancing segmentation performance.}
\end{enumerate}

Comprehensive experiments demonstrate that CWSAM surpasses multiple existing semantic segmentation algorithms without the high requirements of computing costs. Even though meeting characteristic difference between SAR and natural images of SAM, the proposed method achieves to employ the advantages of foundation model to improve the segmentation performance on SAR images.

The remainder of this paper is organized as follows: Section II presents a comprehensive introduction of related work about CWSAM. Section III illustrates the exhaustive architecture of the proposed CWSAM algorithm. Section IV provides quantitative and qualitative evaluation results and comparison of multiple experiments as well as ablation study. Section V summarizes the whole paper with final conclusion. 

\section{Related works}
\subsection{Segment Anything}
Segment Anything Model(SAM) is released by Meta AI Research in the Segment Anything (SA) project for image segmentation and shows impressive zero-shot performance and powerful generalization on multiple application scenes \cite{zhang2023comprehensive}. SAM conducts a Vision Transformer-based image encoder, a light-weighted mask decoder and a flexible prompt encoder allowing inputs of points, bounding boxes, masks and texts. Inspired by the booming Large Language Model (LLM) in natural language processing (NLP), the image encoder of SAM adopts the basic architecture of Vision Transformer and has three different sizes of models, including ViT-Base, ViT-Huge and ViT-Large. MAE \cite{MAE} method is explored to pretrain the Vision Transformer that can process high resolution input images. SAM has powerful zero-shot abilities by prompt engineering that contains two types of prompts: sparse prompt (points, boxes and text) and dense prompt (mask). Positional encoding \cite{Positionalencoding} is responsible for extract prompt embedding of points and boxes, and CLIP \cite{CLIP} for text prompt. Dense prompts are embedded by convolution module and summed element-wise to fuse with image embedding. Multiple prompt methods enable the users to concentrate on various scenes and objects, and bring strong application flexibility relying on SAM zero-shot advantages. In addition, the mask decoder provides three output masks of diverse scopes for per image input to address the issue of ambiguous prompt.

\subsection{Parameter efficient fine-tuning of Segment Anything}
Leveraging the data engine and SA-1B dataset with billions of masks, SAM is regarded as a foundation model of segmentation in computer vision and simulates diverse exploration algorithms on utilizing SAM for downstream tasks on various application fields. References \cite{ma2023segment} \cite{Med-SAM-Adapter} \cite{SAM4Med} explore the potential of SAM on medical image analysis, \cite{tang2023sam} \cite{SAMFails} focus on camouflaged object detection and \cite{SAMCivil} on infrastructure detection. Mirror and transparent objects detection task based on SAM is also involved on \cite{han2023segment}. Furthermore, for aerial images, several researches \cite{ren2023segment} \cite{KnowledgedistillationSAM} robe the generalization of SAM under typical gap between natural scenes of SAM and remote sensing. \cite{ScalingupRSSAM} develops SAM as an efficient annotation system to produce the large-scale remote sensing segmentation dataset called SAMRS. \cite{Text2SegRS} \cite{RSPrompter} conduct prompt engineering to optimize SAM for downstream segmentation task on remote sensing images. 

Considering full fine-tuning a large model is difficult due to the high demand of computing resources and available data, most of these SAM fine-tuning researches follow the mainstream parameter efficient fine-tuning methods of Transformer architecture inspired by LLMs and NLP technologies, including adding additional parameters, selecting a part of parameters to update and introducing reparametrization. The methods of adding additional parameters can be roughly divided into two types: adapter-like \cite{AdaptFormer} and soft prompts \cite{Prefix-Tuning} \cite{Parameter-EfficientPT}.

\subsection{Semantic Segmentation and SAR landcover classification}
Semantic segmentation plays an essential role on computer vision and has made great achievements on proposing a variety of deep learning-based algorithms. Regarded as an extension task from image classification to pixel level image segmentation, semantic segmentation starts from a fundamental architecture of fully convolution networks named FCN \cite{FCN}, and emerges manifold model designs and different structures of mask decoders to improve the segmentation performance. Subsequently, various encoder-decoder architectures \cite{U-Net} \cite{BiSeNet} are optimized with attention mechanisms \cite{ACFNet} \cite{DualAttentionNetworSSk}, proper receptive filed \cite{PyramidSceneParsing} \cite{DenseASPP}, high-resolution feature pyramid \cite{AtrousConvolution} \cite{EncoderDecoderAtrous}, dilated convolutions \cite{DSSDCNFCRF} \cite{DeepLab} and so on. Transformer-based algorithms \cite{SegFormer} \cite{Segmenter} conduct great performance on semantic segmentation task. 

The requirements of pixel level semantic segmentation applications on SAR images tend to be urgent with the development of high-resolution SAR technologies \cite{chinafLiCHGLHZ23} \cite{deoSAR}. Several modified FCNs network are proposed for wetland semantic segmentation task on SAR images \cite{MOHAMMADIMANESH2019223} and PolSAR image semantic segmentation task \cite{PolSARSemantic} through transfer learning. \cite{Oilspillsegmentation} \cite{BuildingsDetectionVHR} \cite{RoadSegmentationSAR} utilize the FCNs networks as baseline and focus on the single category on segmentation tasks of SAR images, such as oil spill, road, and building. \cite{ShiFCWX21} \cite{Zheng2022LandCC} explored the landcover semantic segmentation tasks containing multiple categories on GF-3 SAR images via an encoder-decoder structure and a Transformer architecture. MCANet  \cite{MCANet} proposed a multimodal-cross attention network to achieve land use classification task by fusing the feature map of both optical and SAR images, as well as CFNet \cite{CFNet}. SCNN \cite{StatisticalCNNSAR} utilizes the statistical convolutional neural network to encode and analysis the statistical properties of the SAR image for landcover classification on the TerraSAR-X data.

\begin{figure*} [!ht]
\centering 
\includegraphics[width=\textwidth]{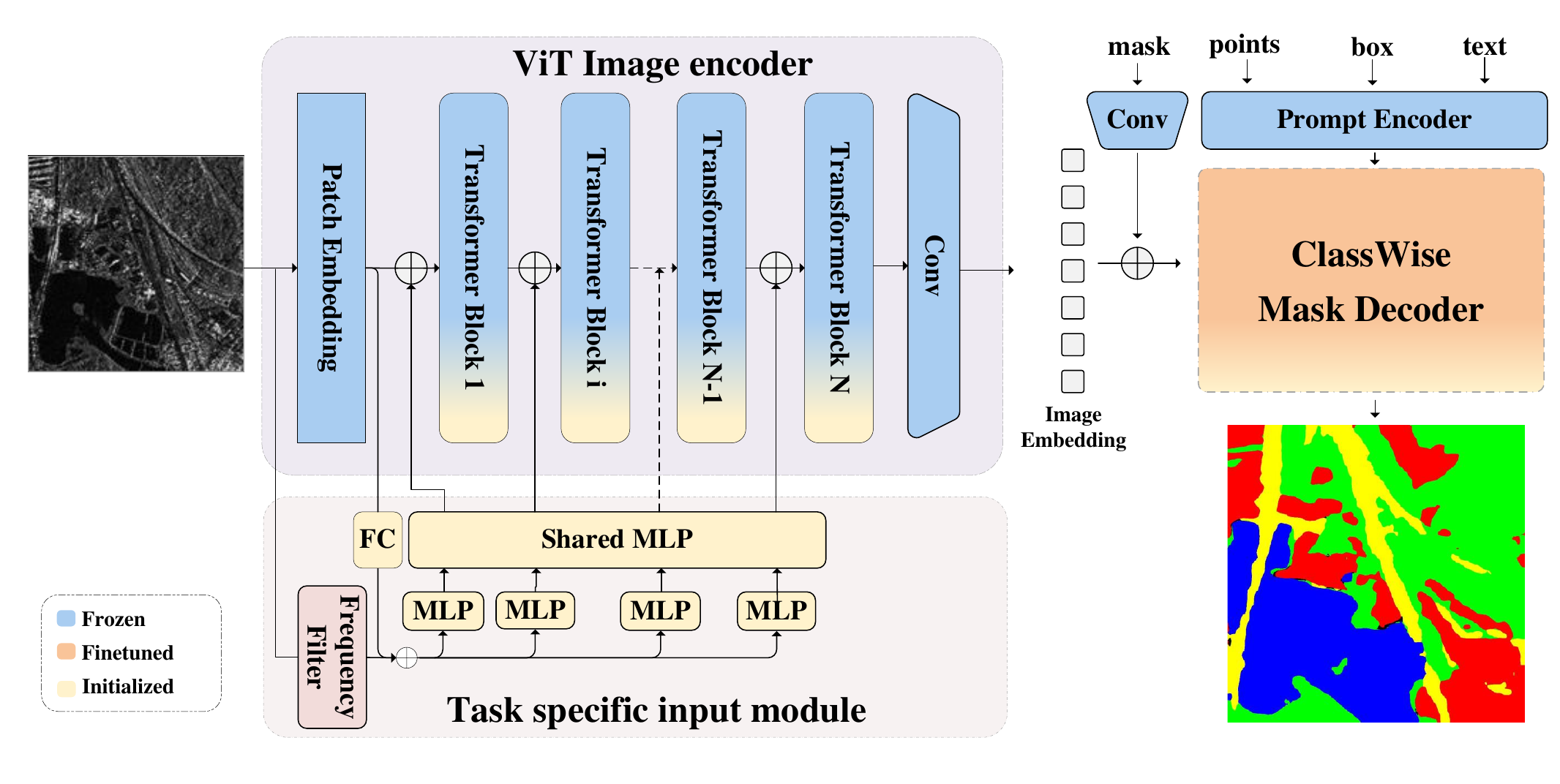}\\ 
\caption{The architecture of the proposed Classwise SAM Adapter.}\label{fig_all} 
\end{figure*}

\section{METHODOLOGY}
In this section, the proposed ClassWise-SAM-Adapter (CWSAM) is introduced by three parts according to its three independent structures. The section III-A presents the adapter fine-tuning algorithm of Vision Transformer image encoder. The classwise mask decoder is introduced in Section III-B to illustrate how to achieve pixel level classification of semantic segmentation task based on SAM. Task specific input module of SAR image low frequency information in Section III-C demonstrates the method of fusing low-frequency semantic information of SAR images into image encoder to benefit the segmentation performance. The figure 2 shows the whole architecture of the proposed CWSAM.
\subsection{Adapted Vision Transformer-based image encoder}
The image encoder of SAM is pretrained on the SA-1B dataset and containing rich semantic information of extensive natural scenes in the real world, so its well iterated parameters are beneficial for downstream task adaption. Therefore, as shown in the ViT image encoder at left part of the Figure 2, in order to leverage SAM as a fundamental backbone, we maintain the original architecture of SAM’s image encoder unchanged which is built from Vision Transformer, and freeze all of original the parameters during the training process. Inspired by AdaptMLP in AdaptFormer \cite{AdaptFormer}, only several simple and efficient adapters are inserted into individual blocks of Vision Transformer structure. At the beginning of image encoder, the patch embedding transforms the input images into sequential patches, and then the extracted patches meet the adapted transformer blocks. 
 
\textbf{Adapters in Vision Transformer blocks:} The detailed pipeline of each Transformer block and the structure of trainable adapter are visualized in Figure 3. 
\begin{figure}[!ht]
\centering
\includegraphics[width=3.5in]{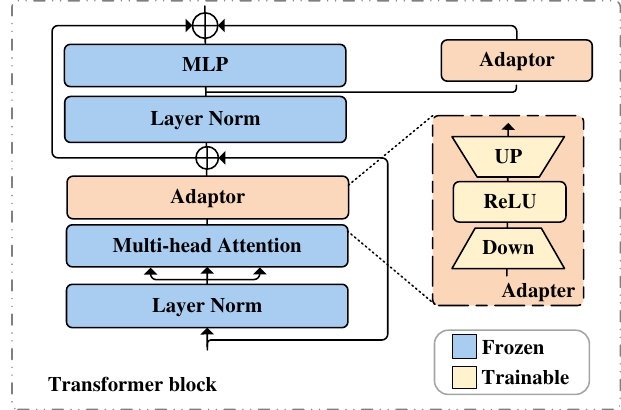}
\caption{Structure of Transformer block and adapter.}
\label{fig3}
\end{figure}
 
Every Transformer block composes two sub-blocks: a multi-head self-attention layer \cite{TransformerNIPS2017_3f5ee243} and an MLP layer, which are injected by adapter modules with initial parameters for training. In the first multi-head attention layer, going through the layerNorm function, the tokens are extracted by multi-head attention mechanism with input of $Q$, $K$ and $V$, which the self-attention function can be described as:
\begin{equation}
    {\rm Attention}({Q}, {K}, {V}) = {\rm Softmax}(\frac{{Q}{K}^\mathsf{T}}{\sqrt{\rm d_{head}}}){V}.
    \label{eqn:attention}
\end{equation}
A light-weighted adapter is following the attention block and meet a feature skip connection before the second MLP sub-block. In the second MLP block, another adapter parallels with original SAM MLP layer and generates output feature of current block. To be specific, all the adapters contain unified structures: a downscale fully connection layer $\rm Down$, a ReLU activation functions and an upscale fully connection layer $\rm Up$, so the function of adapter-tuning is formally formulated as follows:
\begin{equation}
    {\rm Adapter}(f_{input}) = {\rm Up}({\rm ReLU}({\rm Down}(f_{input}))).
    \label{eqn:adapter}
\end{equation}

Therefore, in the $i$th Transformer block, the feature extraction process is:

\begin{equation}
    x_{i}^{'} = {\rm Adapter}({\rm Attention}({\rm LN}(x_{i-1})))+x_{i-1}.
\end{equation}
\begin{equation}
    x_i = {\rm MLP_{sam}}({\rm LN}( x_{i}^{'}))+{\rm Adapter}({\rm LN}( x_{i}^{'}))+ x_{i}^{'} 
    \label{eqn:adapter_block}
\end{equation}
where  $x_i$ and $x_{i-1}$ are the output features of the $i$th and $i-1$th Transformer block respectively, and $x_{i}^{'}$ is the intermediate feature from the first sub-block in the $i$th Transformer block.

With a hand of parameters of these embedded adapters, the proposed method is capable of converting the SAM’s perspective from natural scenes to SAR images and provide reasonable features for subsequent mask decoder. 

\subsection{Classwise mask decoder and loss function}
To represent the multiple categories of every pixels, a classwise mask decoder is built on the original architecture of mask decoder in Segment Anything Model \cite{sam} and consists of a category prediction head to achieve landcover classification task. In the classwise mask decoder of the proposed CWSAM algorithm, the dynamic mask prediction head is ignored so that only one mask result is obtained and involved for computing the training losses.

\textbf{Architecture:} The framework of classwise mask encoder is explained on the Figure 4. At the beginning, Two two-way Transformer blocks are explored to extract features of all embeddings, including image embedding and prompt embedding. Multiple attention mechanisms are employed, such as self-attention for prompt embedding and cross-attention between image embedding and prompt embedding with two directions (image-to-prompt embedding and prompt-to-image embedding). A upscaling convolution block of SAM mask decoder prepares the final output binary mask.

However, with binary mask results, the original design of mask decoder of SAM is unable to figure out classification on pixel level for semantic segmentation task. Therefore, a classwise light-weighted convolution module is implemented to generate classwise masks for all categories. After the upscaling output convolution block of SAM mask decoder, the additional light-weighted upscaling convolution module is added with all randomly initialized parameters trained from scratch to finish final multi-classes mask prediction. The upscale convolution module composes of a deconvolution layer, a RELU activation layer and a convolution block to enlarge semantic feature to generate final mask prediction of $\rm N$ classes channels for separate category. 

\textbf{Feature Enhancement Module:} Besides, considering the semantic segmentation task with pixel level classification requires rich semantic information from feature map, we design a feature enhancement module in the proposed classwise mask decoder. A skip connection between image embedding and upscaling convolution block is parallel to the two-way Transformer block, the output features of the upscaling convolution block from the input of source image embedding and two-way Transformer are concatenated together and double the dimension of original scaled feature map. Consequently, the designed final classwise upscaling module extract the concatenated feature and prepare the pre-prediction mask feature. After the dot product between pre-prediction mask and tokens from MLP layers, the predicted multi-class mask is generated and each pixel of image is assigned with a specific category. 
\begin{figure}[!ht]
\centering
\includegraphics[width=3.5in]{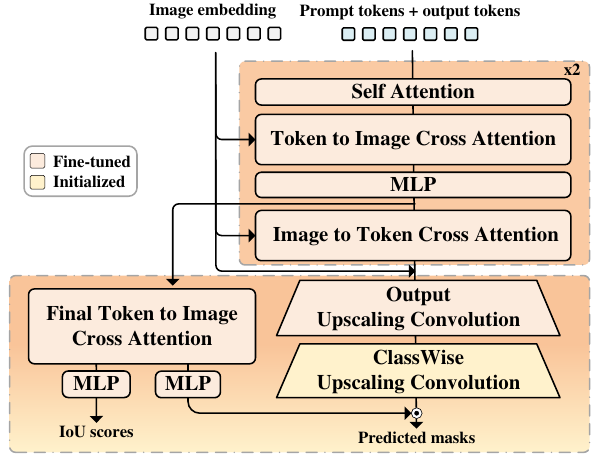}
\caption{Embedding pipeline of classwise mask decoder.}
\label{fig4}
\end{figure}

\textbf{Loss function}: For unbalanced data distribution, the loss function is Weighted Cross Entropy loss which is formulate as:
\begin{equation}
\begin{split}
{\rm Loss} = - \sum_i^{\rm N} {\rm {w_i}} (\cdot y_i \cdot \log(\sigma(logits_i)) + \\
    (1 - y_i) \cdot \log(1 - \sigma(logits_i)))
\end{split}
\end{equation}
where $logits$ refer to the original output by the model, N is the number of categories, $w_i$ is the weighted value of the $i$ class, $y_i$ is the ground truth and $\sigma(x)$ is $sigmoid$ function as:
\begin{equation}
\sigma(x) = \frac{1}{1 + e^{-x}}
\end{equation}

\subsection{Task specific input of low frequency SAR characteristics}
Although fine-tuning the parameters of adapters in image encoder transfer SAM to SAR image domain, the difference between natural sense and microwave signal continuously hinders the model performance. The feature extraction of SAM’s Vision Transformer image encoder still lack of SAR domain information because some semantic information will be lost through the frozen parameters. Therefore, to extend reasonable semantic information of SAR images for segmentation downstream task, we build an MLP-based architecture paralleled with VIT image encoder to maintain meaningful low frequency information of SAR images, such as land surface texture and pixel brightness depending on signal reflectivity. First, the low frequency characteristics of input images are derived by the processes of 2D Fast Fourier Transform, low pass filter and inverse Fast Fourier transform \cite{FFT}, which is described as:
\begin{equation}
    {\rm F}(u, v) = \sum_{x=0}^{\rm M-1} \sum_{y=0}^{\rm N-1} f(x, y) \cdot e^{-j2\pi \left( \frac{ux}{\rm M} + \frac{vy}{\rm N} \right)}
\end{equation}

where $f(x,y)$ is the image in the spatial domain, and ${\rm F}(u,v)$ is its representation in the frequency domain, $\rm M$ and $\rm N$ are the width and height of the image, respectively. $u$ and $v$ are the frequency coordinates. Then, in order to decrease computing cost, a rectangular low-pass filter in the frequency domain is employed as:
\begin{align}
{\rm H}(u, v) = 
\begin{cases} 
1 & \text{if } \frac{\rm M}{2} - \frac{\rm W}{2} \leq u \leq \frac{\rm M}{2} + \frac{\rm W}{2} \\
& \text{and } \frac{\rm N}{2} - \frac{\rm H}{2} \leq v \leq \frac{\rm N}{2} + \frac{\rm H}{2} \\
0 & \text{otherwise}
\end{cases}
\end{align}

where $\rm M$ and $\rm N$ are the number of rows and columns in the image, respectively. $\rm W$ and $\rm H$ are the width and height of the filter in the frequency domain. $u$ and $v$ are the frequency coordinates.

\begin{figure}[!ht]
\centering
\includegraphics[width=3.5in]{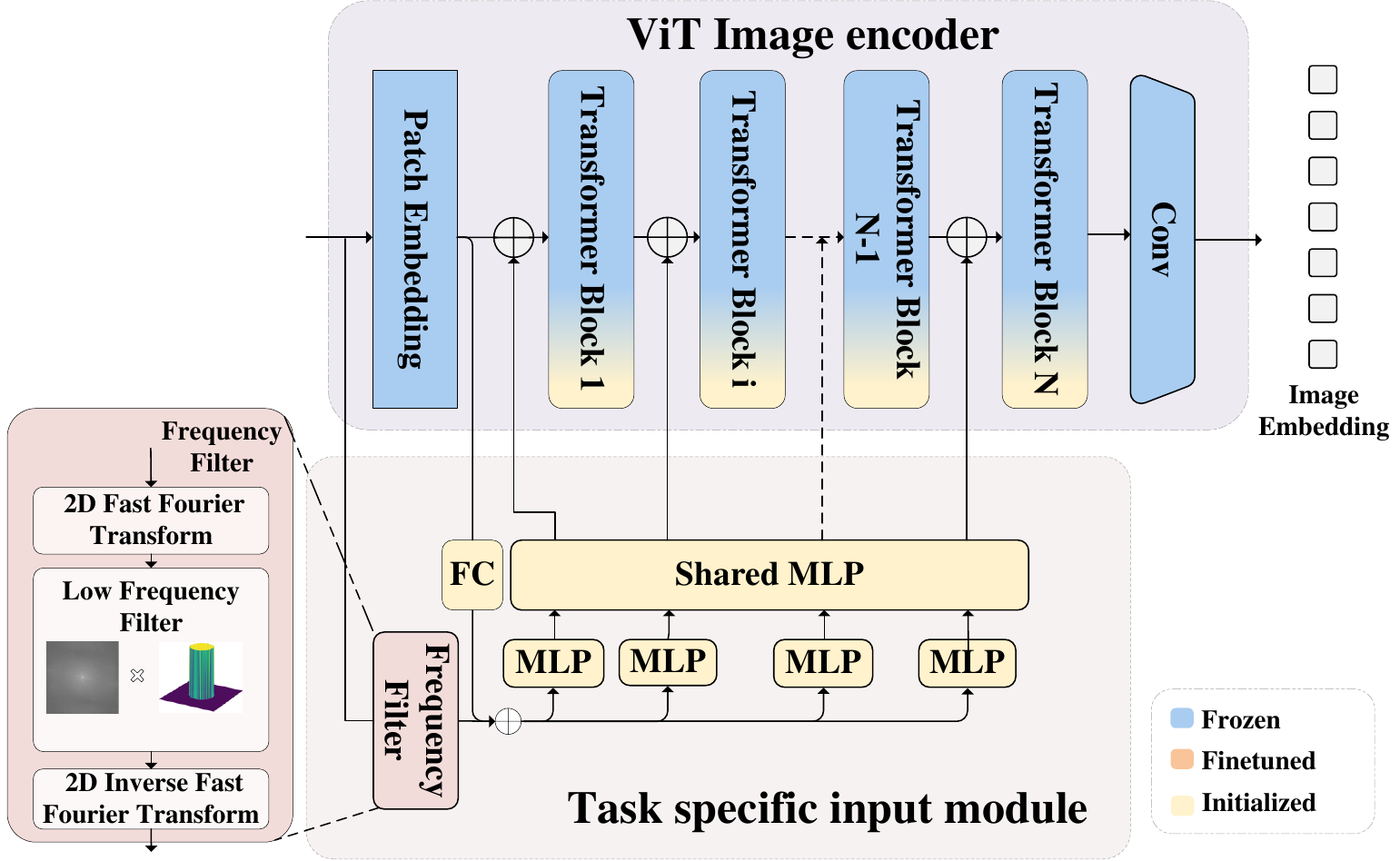}
\caption{the structure of task specific input module.}
\label{fig5}
\end{figure}
Refer to Figure 5, the patch embedding feature of input images in ViT encoder follows dimensional reduction and adds with low frequency feature. Afterwards multiple MLP blocks extract this feature for every Vision Transformer blocks respectively, and finally a parameter-shared MLP block inspried by \cite{SAMFails} is responsible for realizing each feature fusion between task specific input module and original SAM image encoder.

\section{EXPERIMENTAL RESULTS AND ANALYSIS}
\subsection{Experimental Dataset and Settings}
Two landcover classification datasets of SAR images are utilized for evaluation experiments of the proposed method, FUSAR-Map1.0 \cite{ShiFCWX21} and FUSAR-Map2.0. \cite{Zheng2022LandCC} Both datasets contain multiple categories of landcover annotations, and represent rich semantic information for SAR image interpretation, while FUSAR-Map2.0 raises higher requirements to algorithms by extreme data distribution and detailed label.

(1) FUSAR-Map1.0: The FUSAR-Map1.0 dataset comprises 610 GF-3 satellite SAR images of size 1024x1024 pixels from twelve scenes of six provinces in China. The annotations contain four landcover categories, building, vegetation, water and road. One scene of Wuhan city is selected for test dataset and cut into 80 slices. The other eleven senses generate total 530 SAR images of size 1024x1024 for training dataset. In addition, parts of pixels in images do not have category labels because of their illegible appearances and are excluded from both training and evaluation processes.

(2) FUSAR-Map2.0: The FUSAR-Map2.0 dataset has more fine-grained semantic information of landcover categories, including total 10 landcover types, i.e., water, woodland, vegetation, bare soil, industry, residence, road, paddy, planting and human built, and the quantity of each category is extremely unbalance which brings challenges to segmentation models. Total 10 scenes from various countries are included in this dataset, the proportion of training and testing scenes is 8:2. Cropped to the slides of size 1024x1024 pixels, the FUSAR-Map2.0 consists of total 738 image samples, and 549 for training and 189 for testing. The annotation distribution of every category between train and test dataset are set to be approximate for reasonable evaluation experiments.

\subsection{Evaluation Protocol and Metrics}
To analyze the effectiveness of the proposed method, diversified standard evaluation metrics of semantic segmentation are employed \cite{Reviewonss}. 

(1) mIOU (mean Intersection over Union): mIoU is widely utilized to evaluate the average segmentation performance of all categories, calculating the intersection over union (IoU) of predicted mask and ground-truth, which is described as: 
\begin{equation}
\text{mIoU} = \frac{1}{\rm k} \sum_{i=1}^{\rm k} \frac{\text{TP}}{\text{FN} + \text{FP} + \text{TP}}   
\end{equation}
where $\text {TP}$ represents the true positives. $\text{FN}$ represents the false negatives. $\text{FP}$ represents the false positives. The sum is taken over all classes, indexed from 1 to $\rm k$. The total number of categories is $\rm k$.

(2) OA (Overall Accuracy): OA represents the total percentage of the correctly classified pixels in the whole images as:
\begin{equation}
    \text{OA} = \frac{\sum_{i=1}^{\rm k} p_{ii}}{\sum_{i=1}^{\rm k} T_i}
\end{equation}

where $p_{ii}$ represents the number of correctly classified pixels for the $i_{th}$ class. $T_i$ represents the total number of pixels for the $i_{th}$ class. The total number of categories is $\rm k$. 

However, OA may be influenced by main category with the highest number of pixels in the dataset. Therefore, other metrics are also considered in the experiments, including Accuracy, precision, and mDice. In our experiments, we treat mIOU as the main evaluation metric to concentrate on the average performance on all the landcover categories, especially the difficult and ambiguous classes such as road and building. 

\subsection{Implementation Details}
Our experiments of the proposed method are conducted on the architecture of SAM ViT-Base backbone and compared with multiple the state-of-the art algorithms in semantic segmentation task under the unified training configuration and settings using Pytorch.

(1)	Training configurations: Two Nvidia RTX 3090 GPUs are employed to train the model for total 120 epochs, and only one for inference process. All the experiments adopt the AdamW optimizer and cosine annealing learning rate strategy with the initial learning rate of 0.0002. All the images have the resolution of 1024x1024 with the batch size of 1 in the training process to fully make use of computing resources. 

(2)	Model architecture: Based on the ViT-Base which is the smallest model of SAM, the main feature dimensions of the model maintain unchanged, including SAM image encoder, prompt encoder and most part of mask decoder. In Vision Transformer-based image encoder, the first module is a patch embedding block with the patch size of 16 and 14x14 windowed attention. Subsequently, 12 number of Transformer blocks with embedding dimension size of 768 extract the image features of the constant dimensions as 64x64x768. Four equally-spaced global attention blocks are set on layers 2, 5, 8, and 11 on the ViT image encoder, and a final convolution block generates the image embeddings with 256 output channels. 

Besides, the feature dimension of all adapters in the image encoder is 768, which is same as embedding dimension of ViT blocks, and the hidden dimension of adapters is one quarter of embedding dimension. For task-specific input, the channel dimension is downscaled to 24 by a fully connected layer for light-weighted computing process, and maintains invariant at individual light-weighted MLP blocks. A shared MLP block increase the channel dimension from 24 to the embedding dimension size of 768 in order to finish the feature fusion with ViT blocks.

For the classwise mask decoder with feature enhance module, the Transformer dimension is 256 and the SAM output upscaling block dimensions are 256 for input, 64 for hidden layer, and 32 for output features. After feature enhanced concatenation, the classwise upscaling block takes the input feature of 64 channels, and conduct the transposed convolution module of output dimension of 64, kernel size of 2 and stride of 2, to upscale the feature resolution from 64x64 to 256x256. Following a normalize layer and activation layer, a convolution block outputs the classwise upscaled embedding of 32 x number of classes channels which is reshaped to size of (1, 32, number of classes, 256, 256) for each ambiguous mask per batch size. Finally, four ambiguity-aware predicted masks are derived from a spatially point-wise product between the upscale classwise image embedding and MLP outputs whose size are 4 x (1,32).

(3)	Weighted Losses: When the data distribution in landcover classification dataset of SAR images is unbalance, some categories that are easy to be recognized or have sufficient samples will dominate the model iteration and convergence during the training process. Therefore, the weighted cross entropy losses are employed in our experiments to pay more attention to these sparse and difficult categories. For five types of landcover in FUSAR-Map1.0 dataset, the weights of building, vegetation, water, road, background are 1.5, 1, 0.5, 1.9 and 0.1 respectively, and the background is excluded during evaluation. For FUSAR-Map2.0 dataset, the weights of water, bare soil, road, industry, residence, vegetation, paddy field, planting area, human-built and woodland are 0.4, 1.6, 0.4, 1.6, 1.0, 0.5, 1.5, 1.7, 1.4 and 0.9, respectively. All of these ten landcover categories are evaluated in experiments, and the data distributions of the training and test scenes are relatively similar to each other for reasonable analysis.

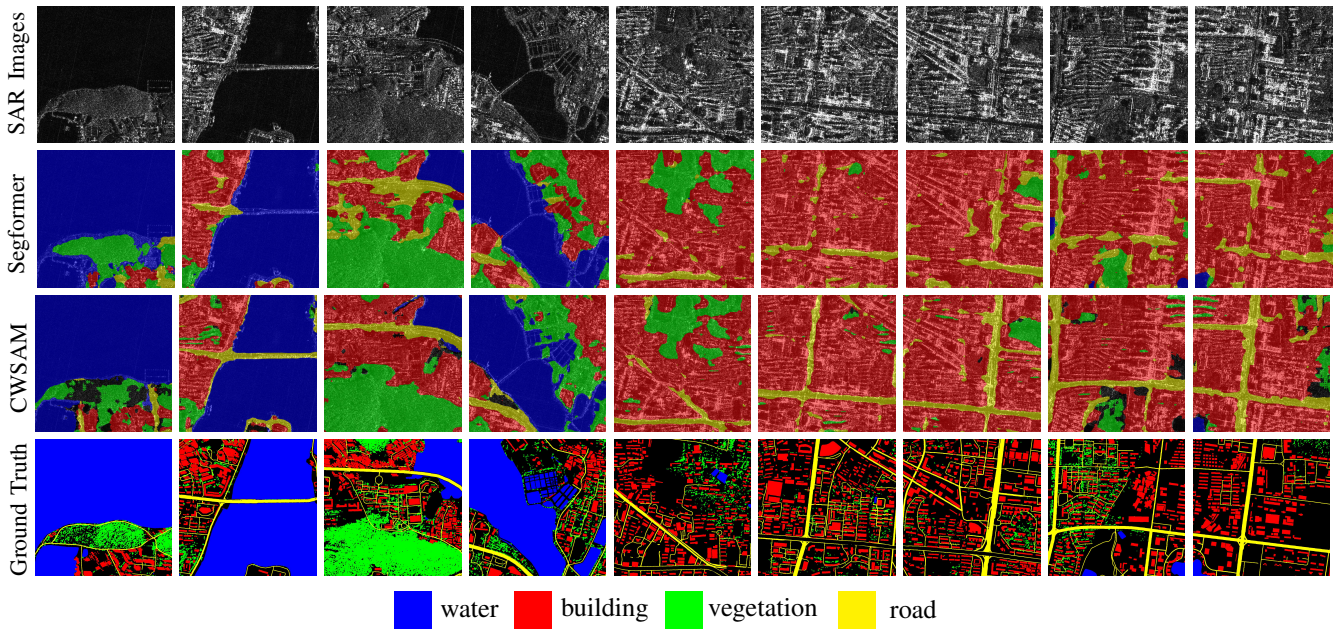
\begin{figure*}[tbhp]
\centering
    \begin{minipage}[t]{\linewidth}
	   \centering
            \small\rotatebox{90}{\hspace{3pt}SAR Images}
	       \includegraphics[width=.1\linewidth]{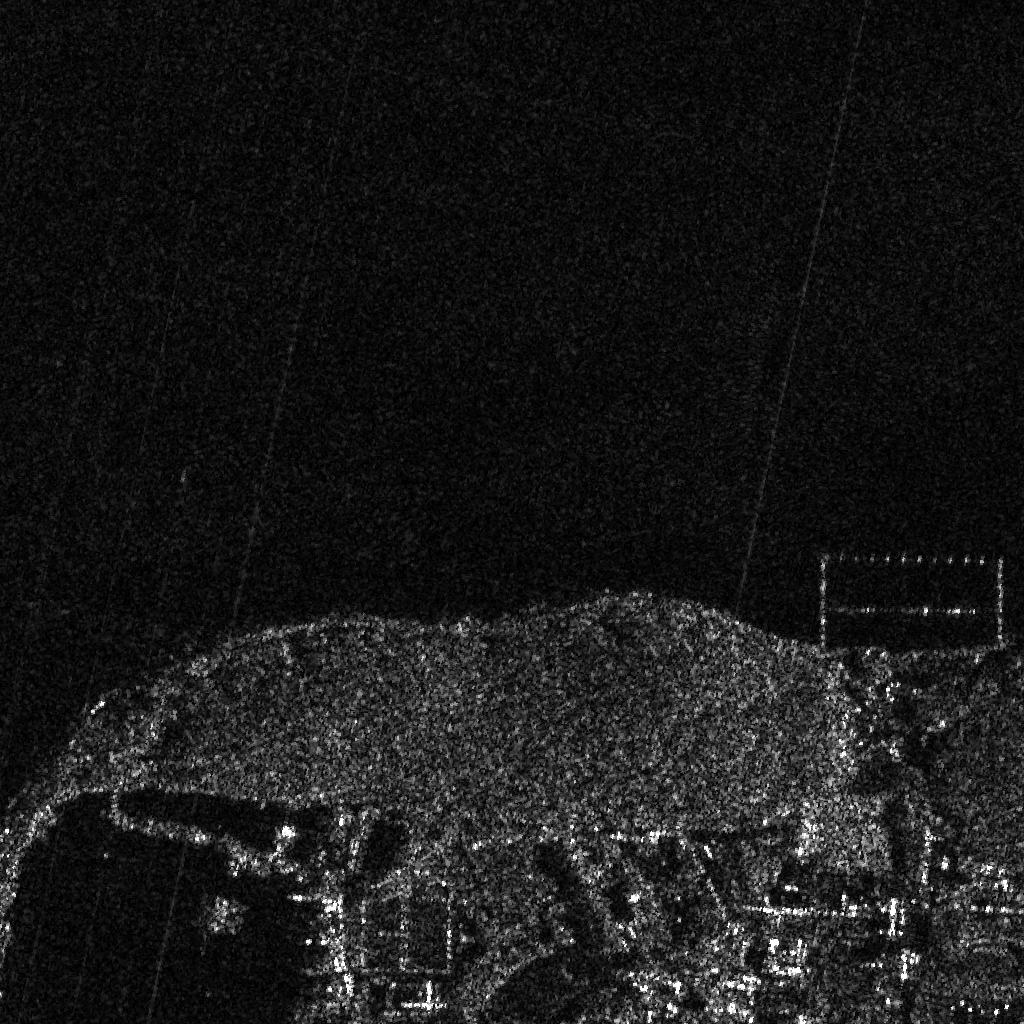}
	       \includegraphics[width=.1\linewidth]{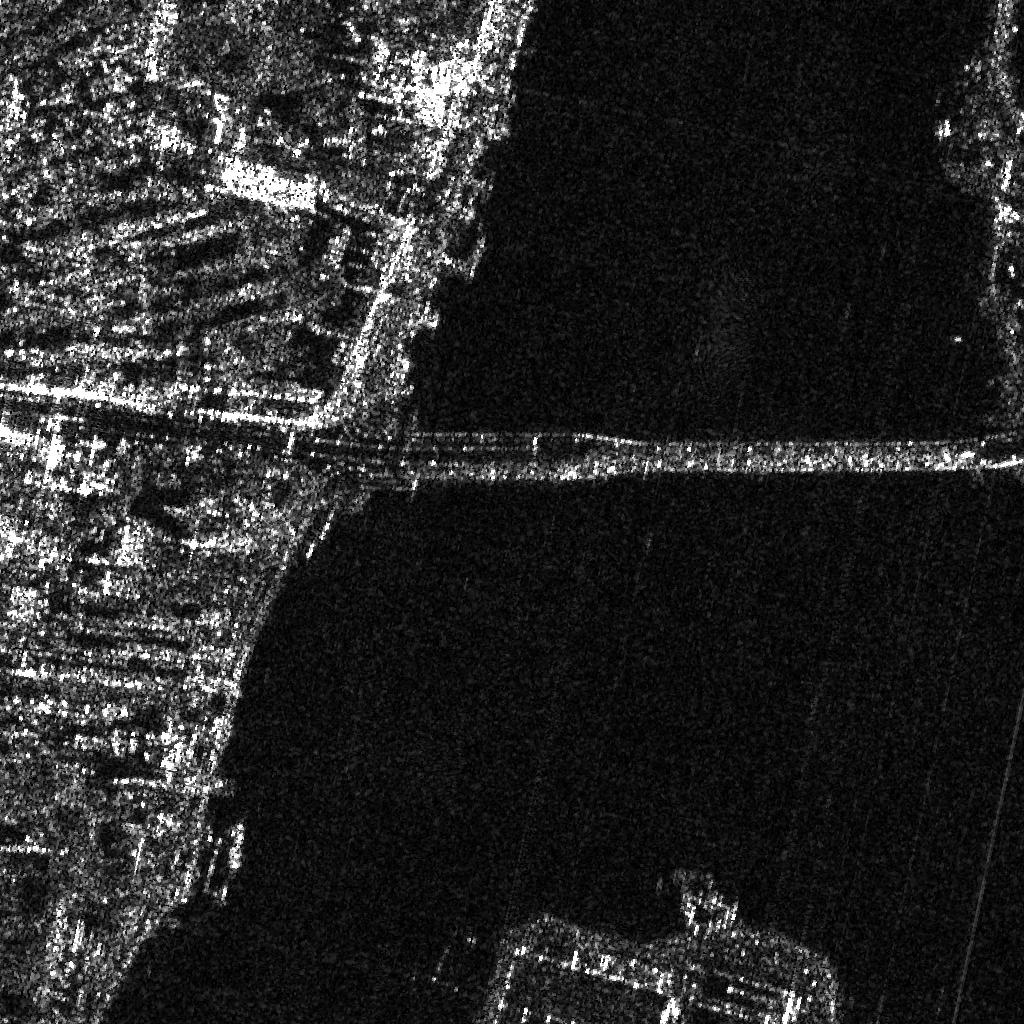}
              \includegraphics[width=.1\linewidth]{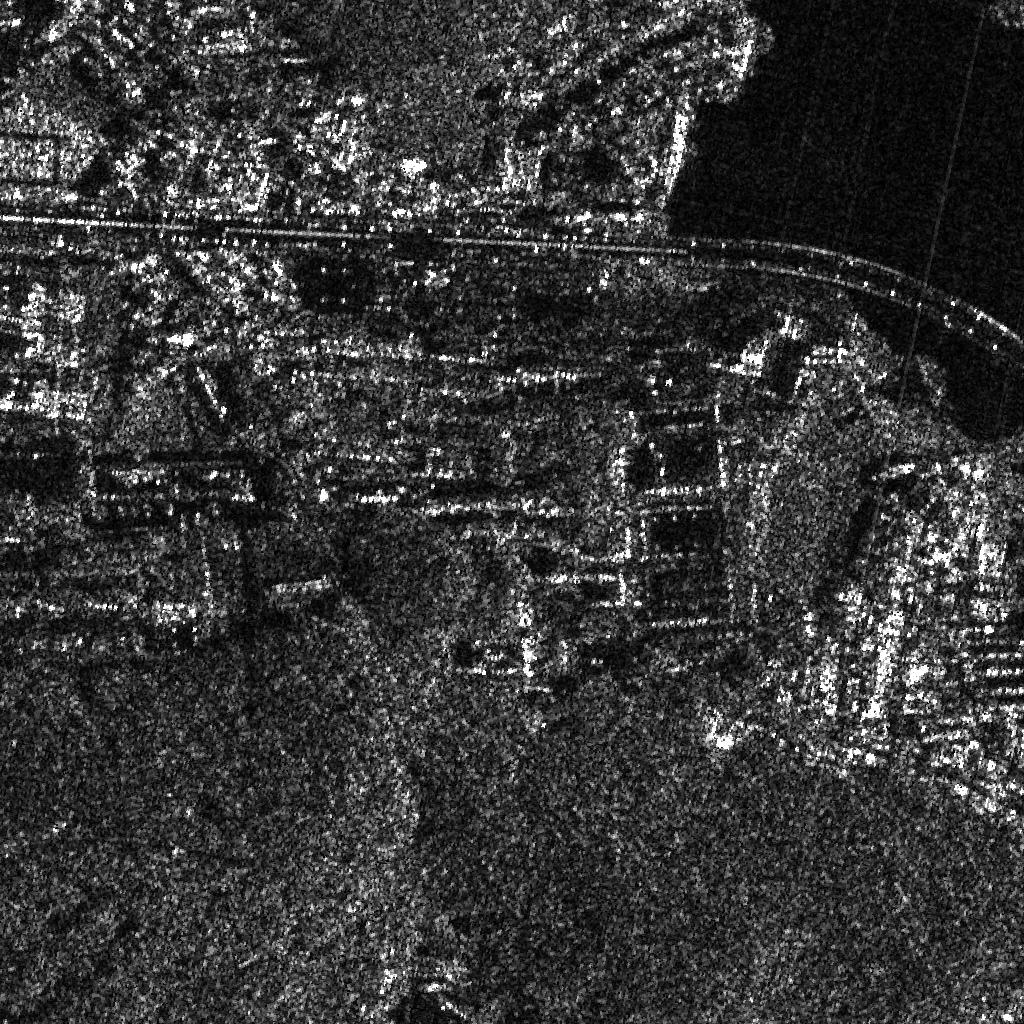}
              \includegraphics[width=.1\linewidth]{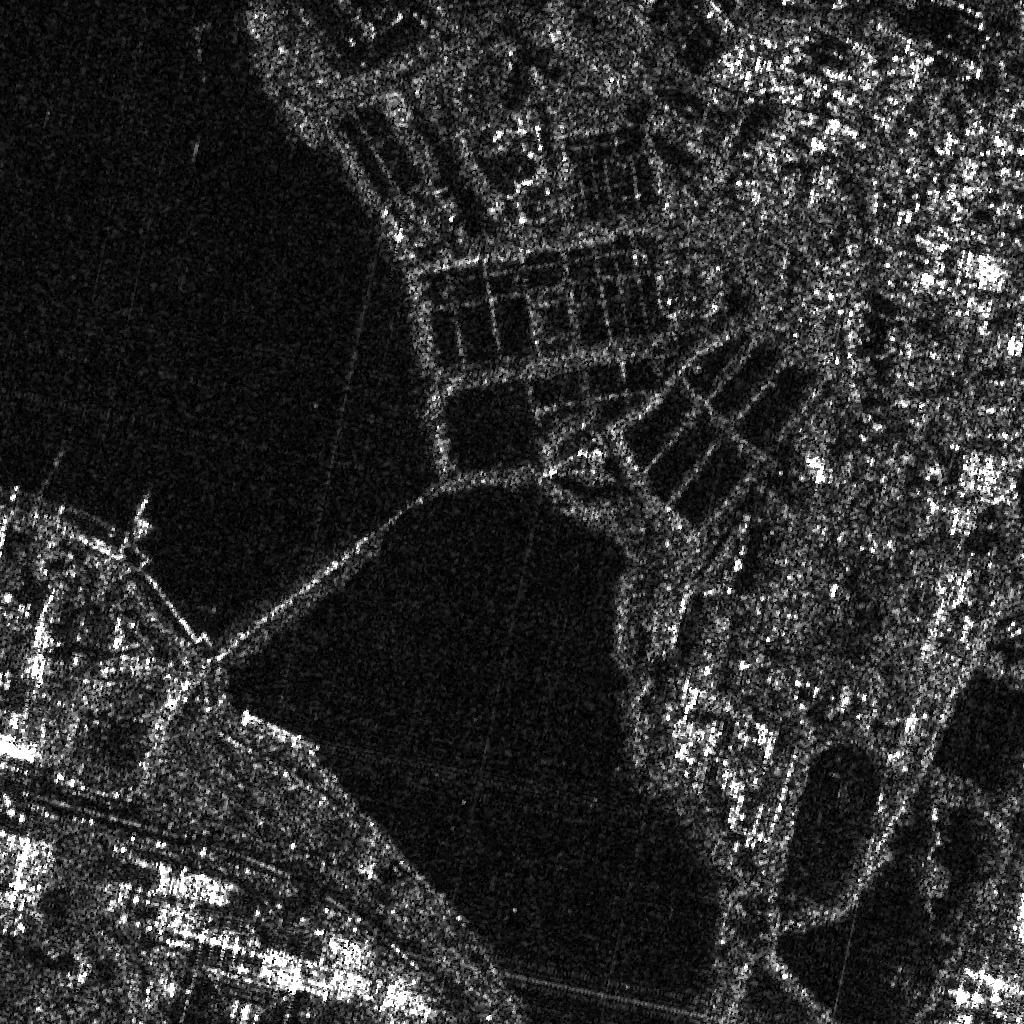}
              \includegraphics[width=.1\linewidth]{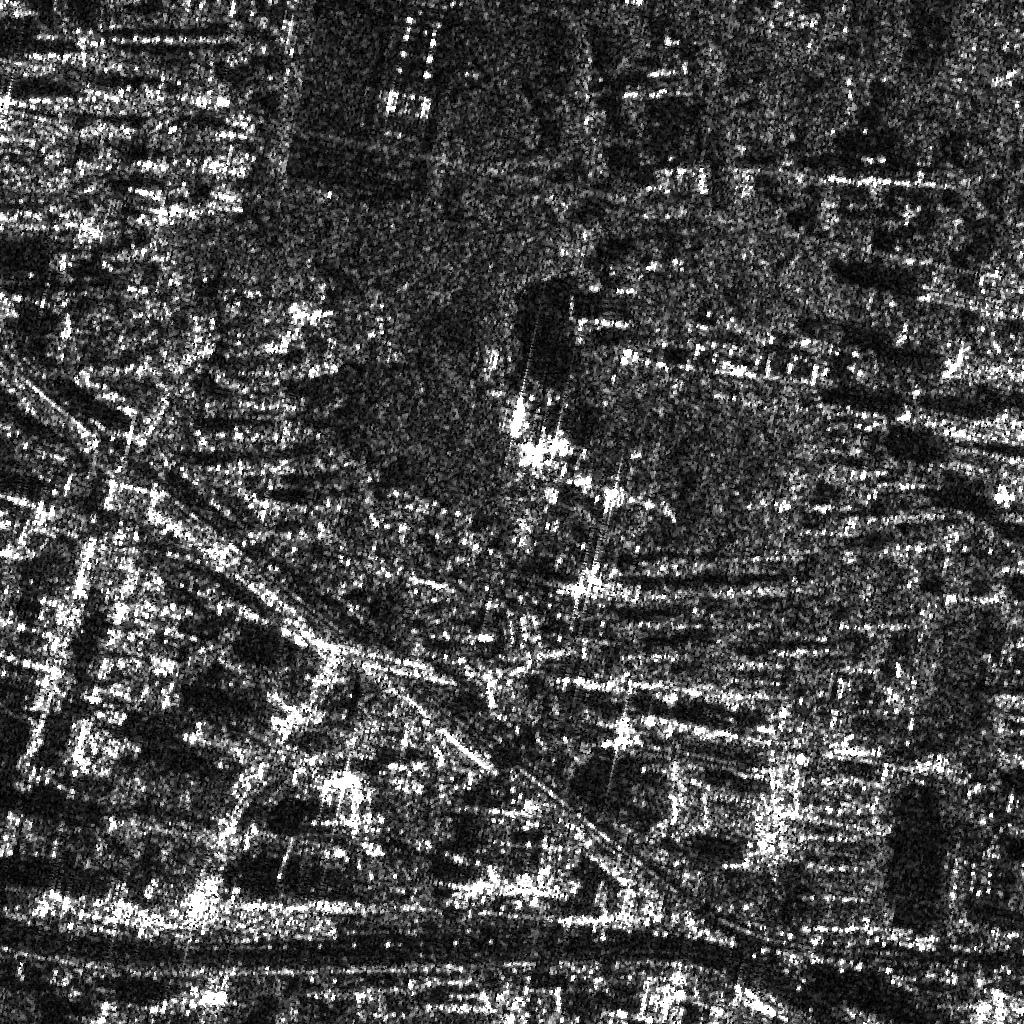}
              \includegraphics[width=.1\linewidth]{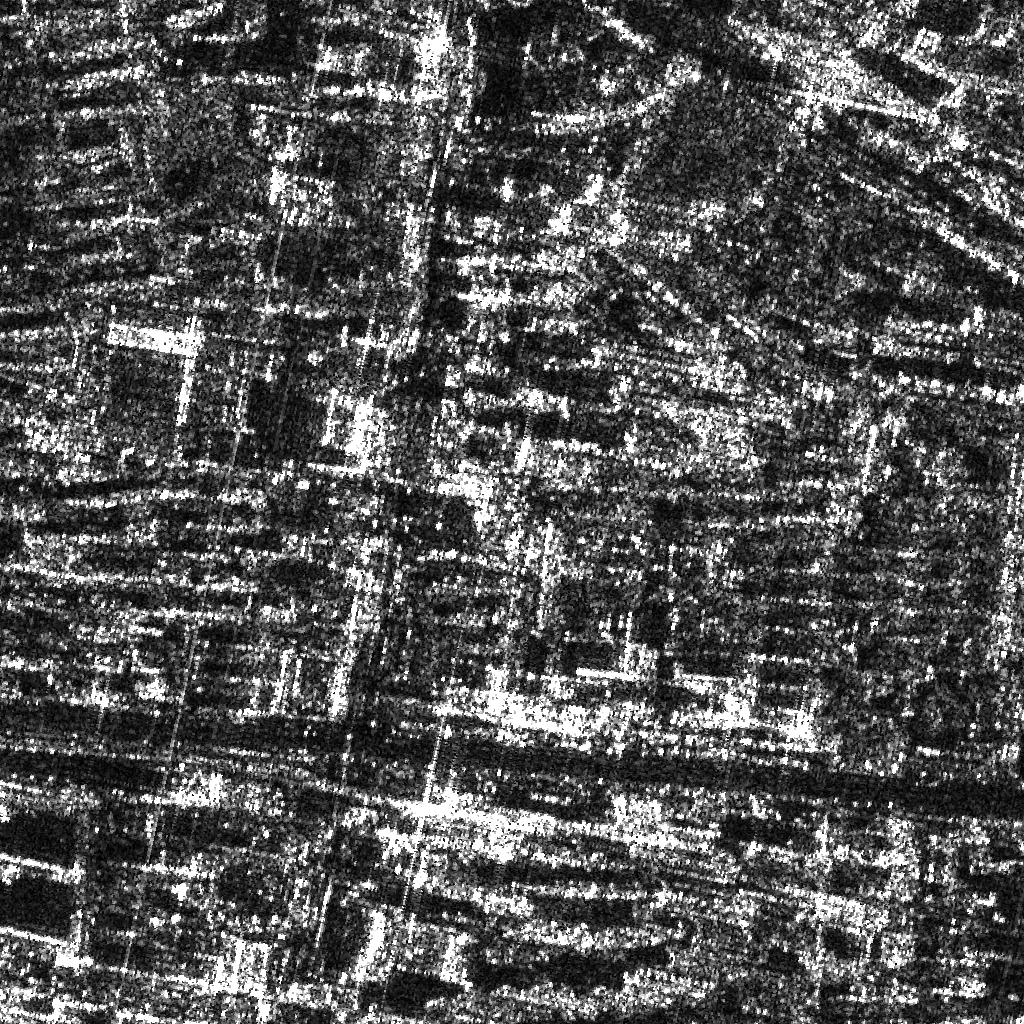}
              \includegraphics[width=.1\linewidth]{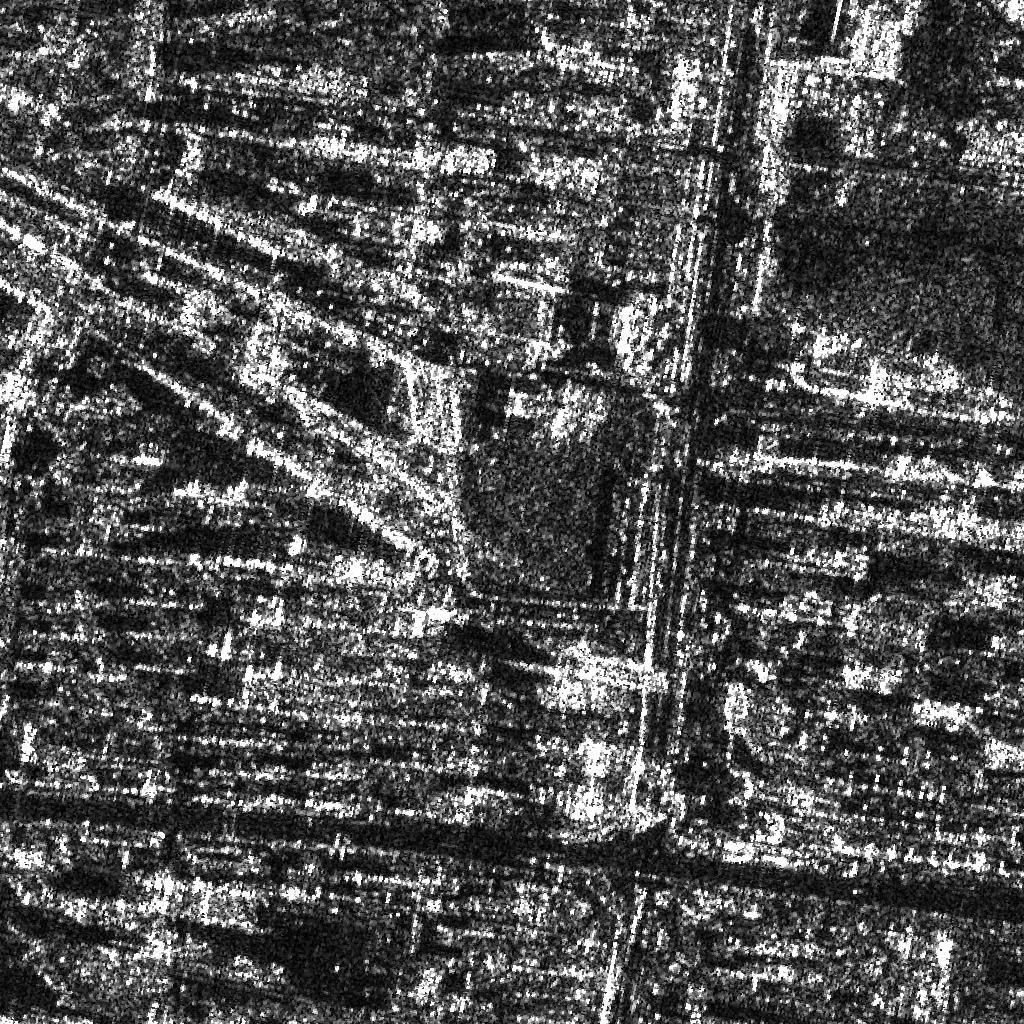}
              \includegraphics[width=.1\linewidth]{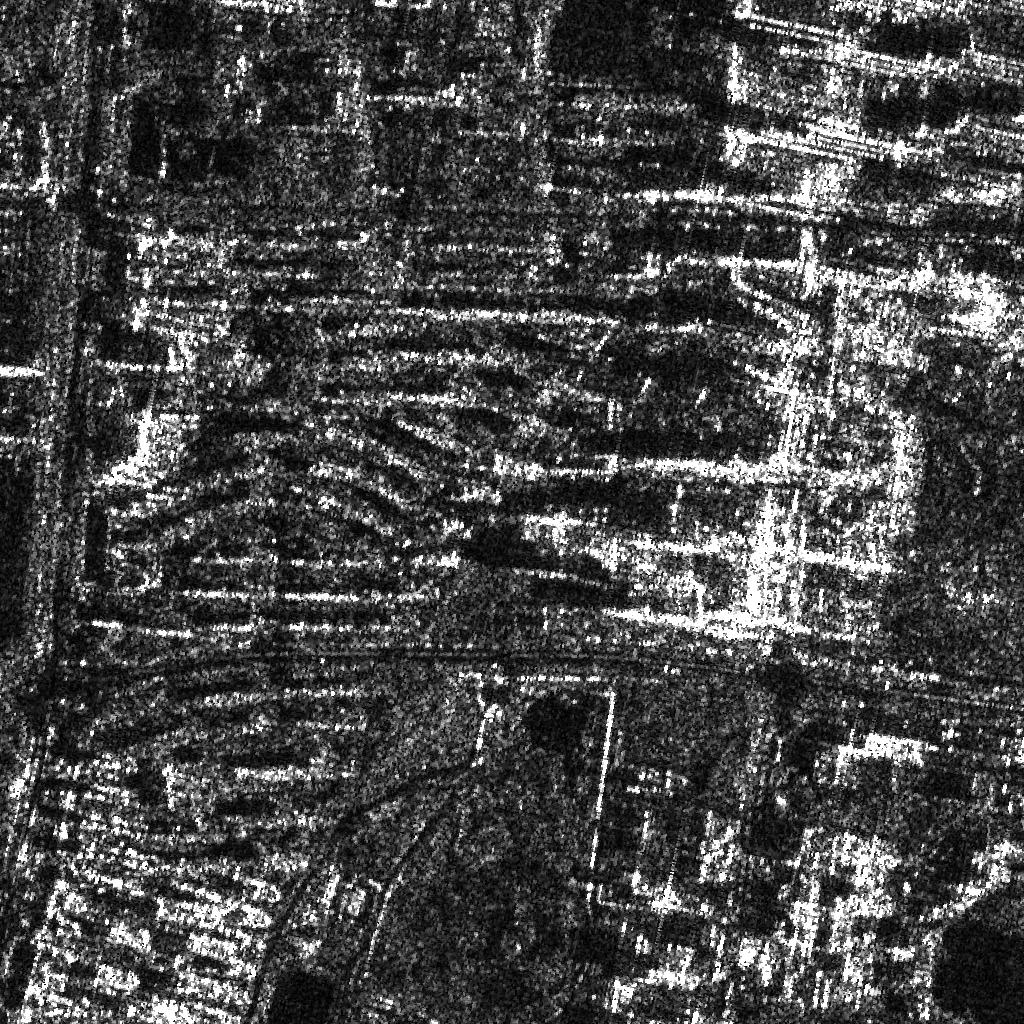}
              \includegraphics[width=.1\linewidth]{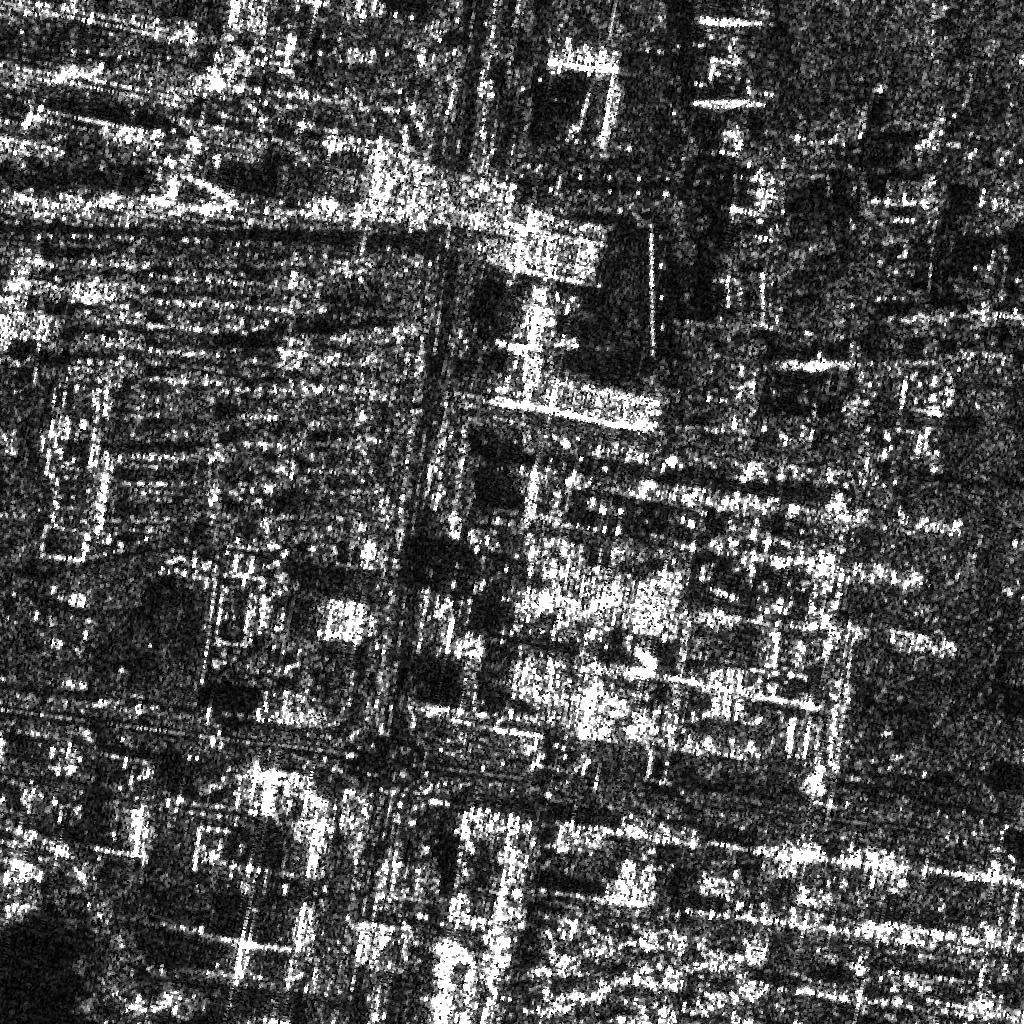}
	\end{minipage}\vspace{3pt}

    \begin{minipage}[t]{\linewidth}
	   \centering
            \small\rotatebox{90}{\hspace{5pt}Segformer}
	       \includegraphics[width=.1\linewidth]{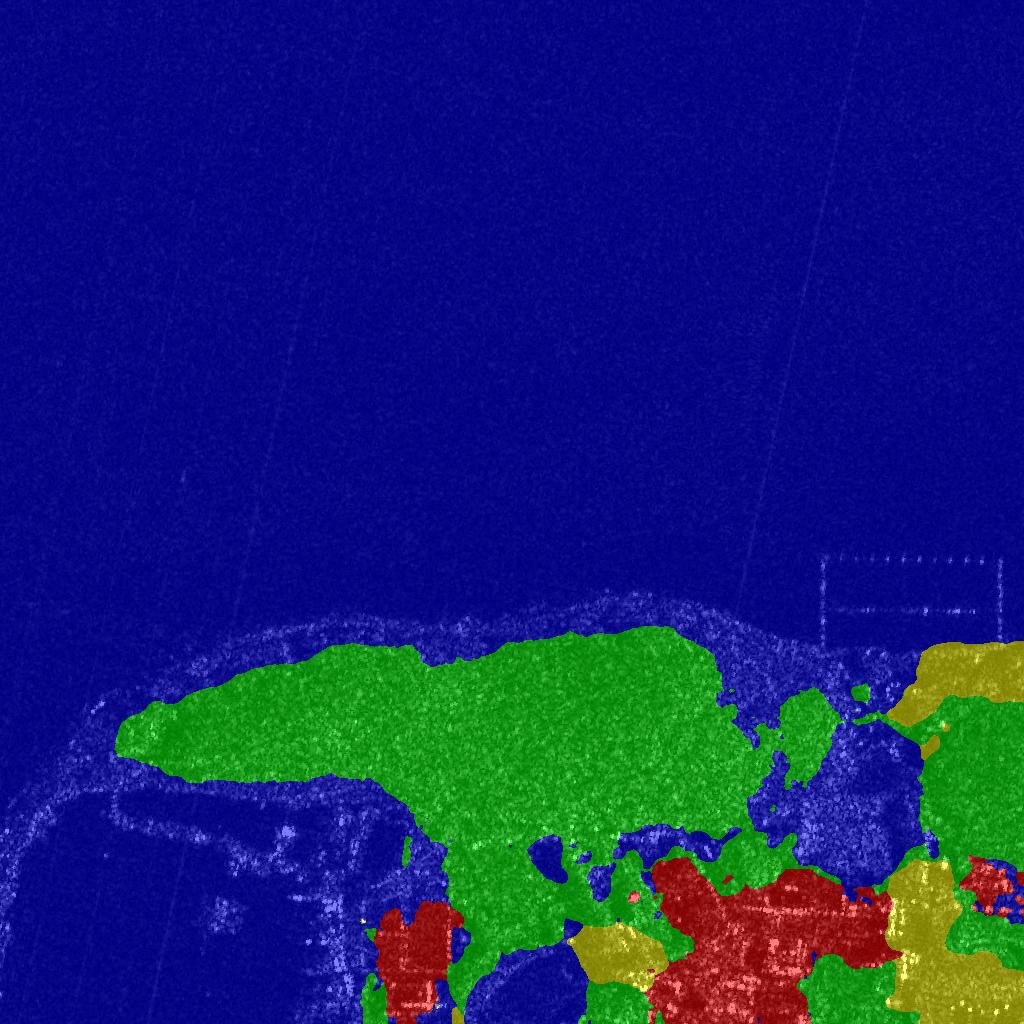}
	       \includegraphics[width=.1\linewidth]{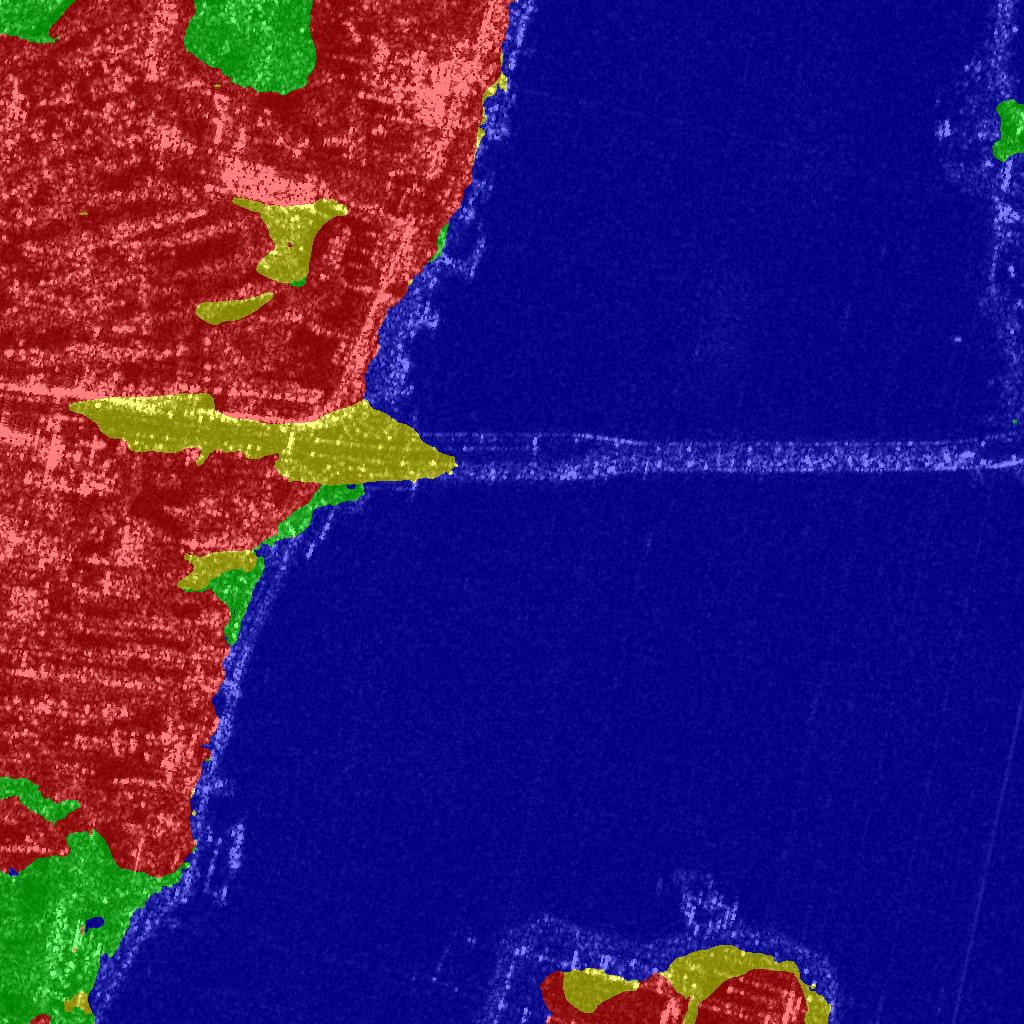}
              \includegraphics[width=.1\linewidth]{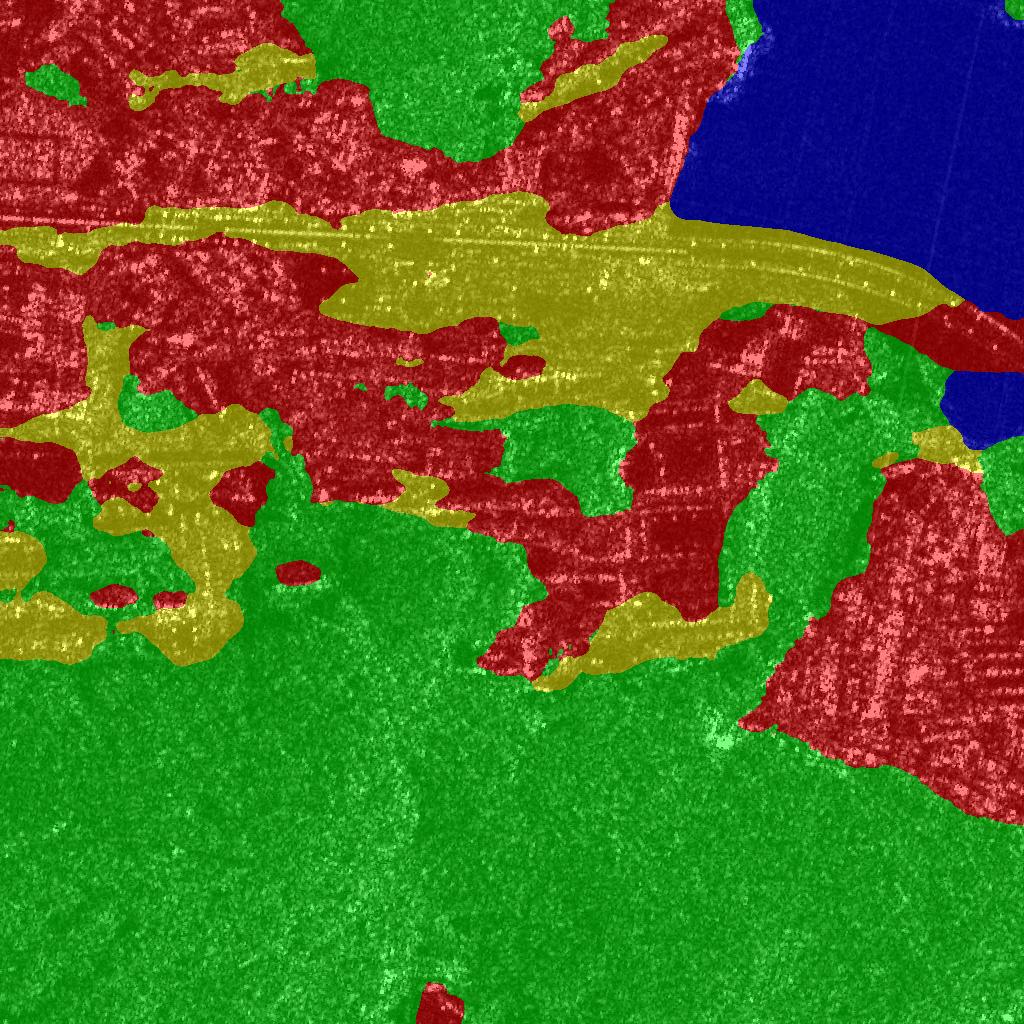}
              \includegraphics[width=.1\linewidth]{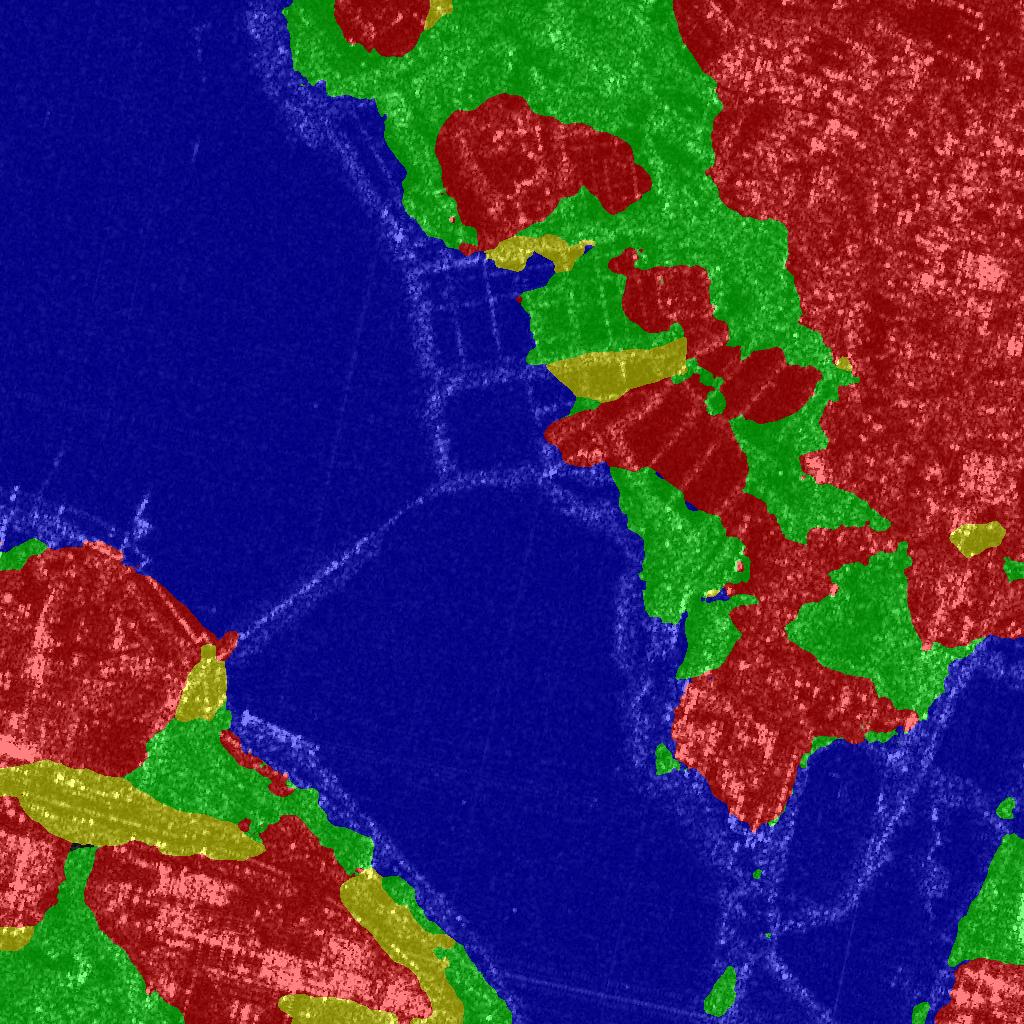}
              \includegraphics[width=.1\linewidth]{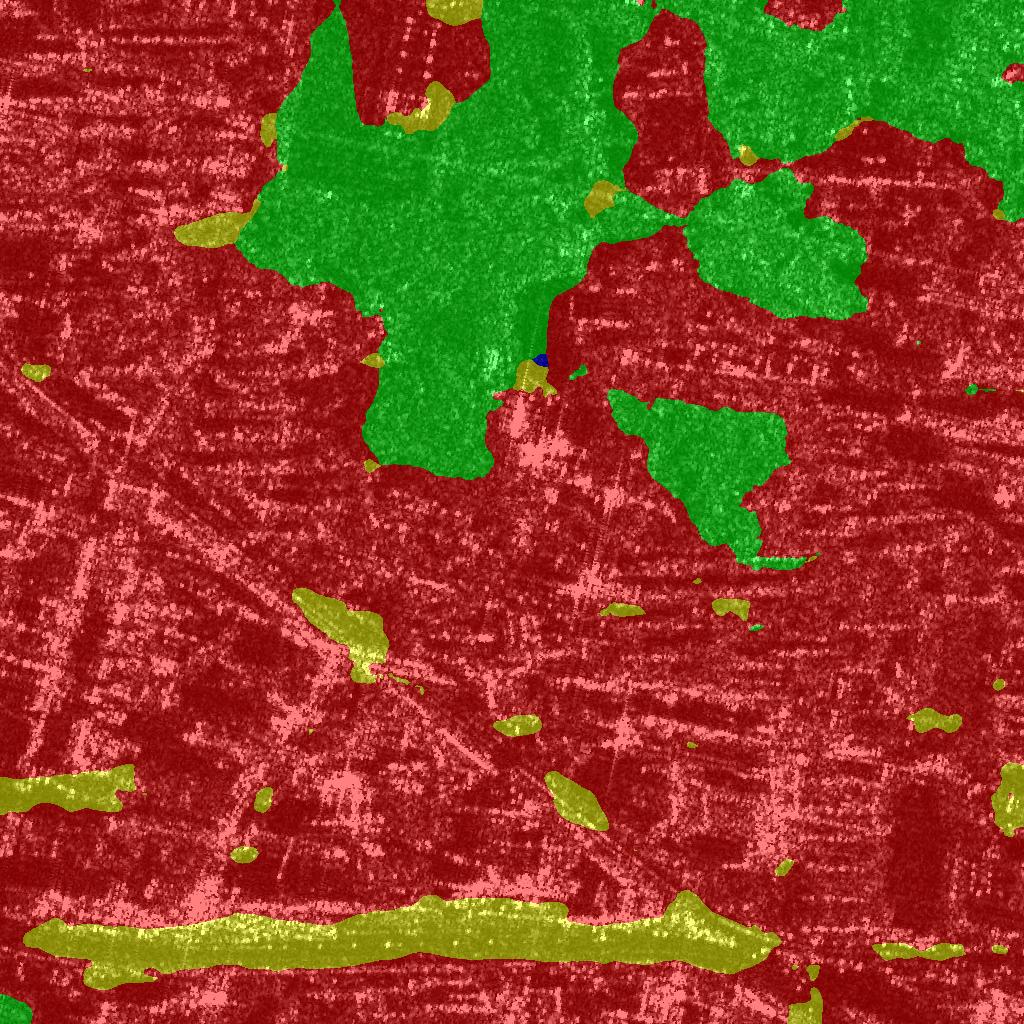}
              \includegraphics[width=.1\linewidth]{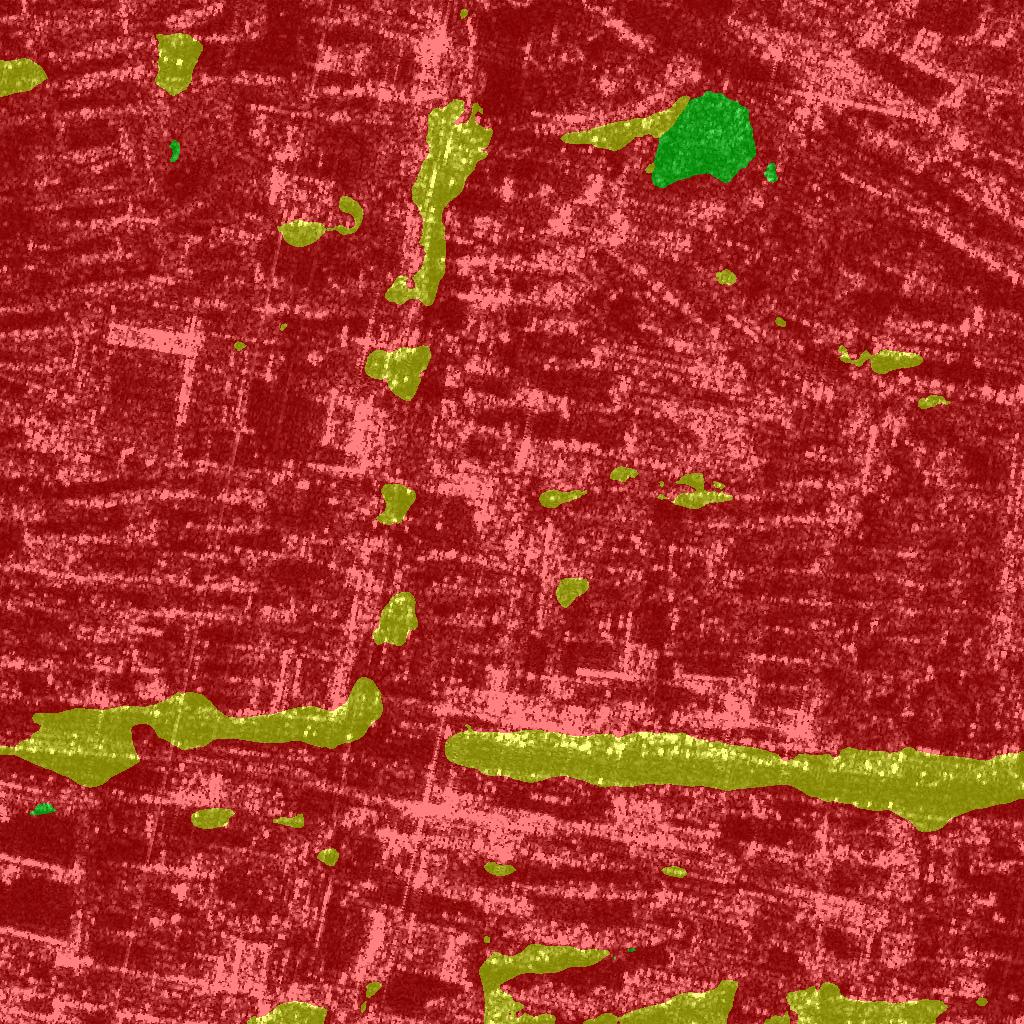}
              \includegraphics[width=.1\linewidth]{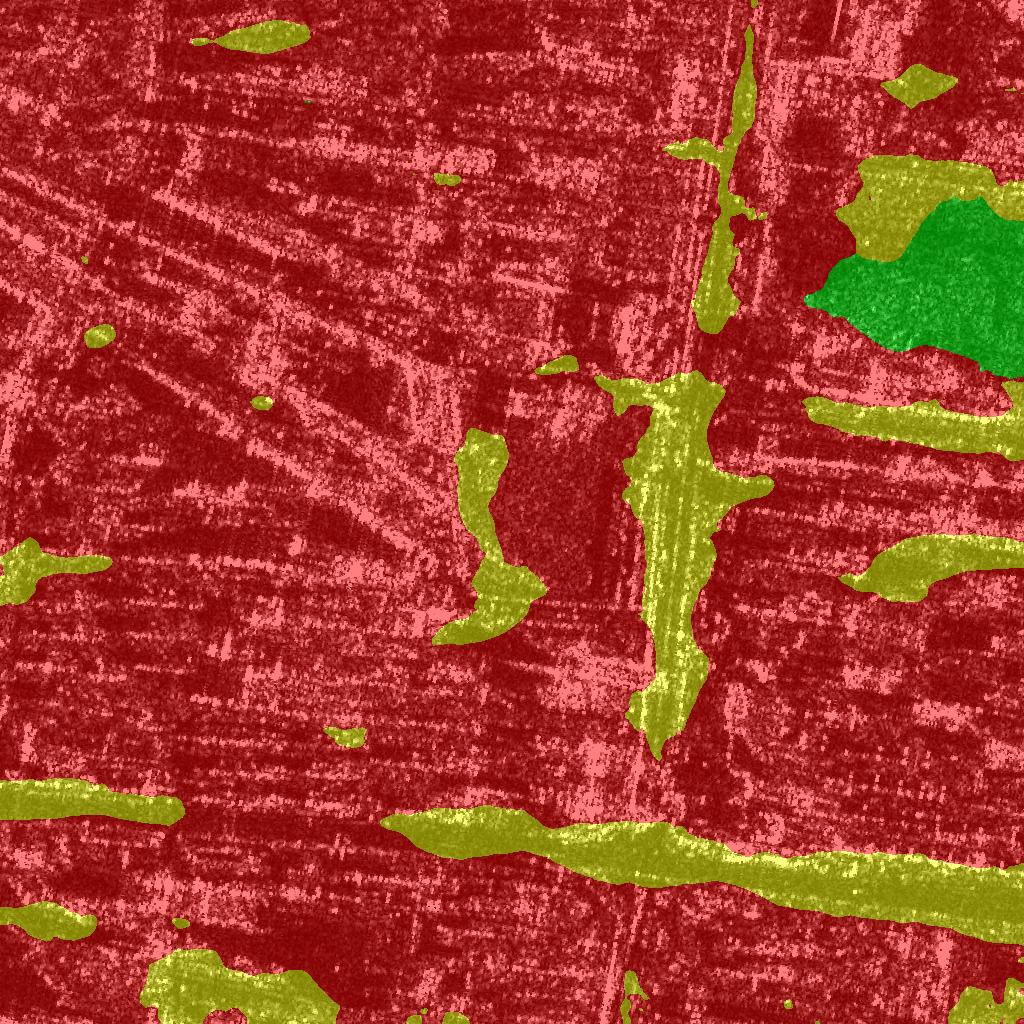}
              \includegraphics[width=.1\linewidth]{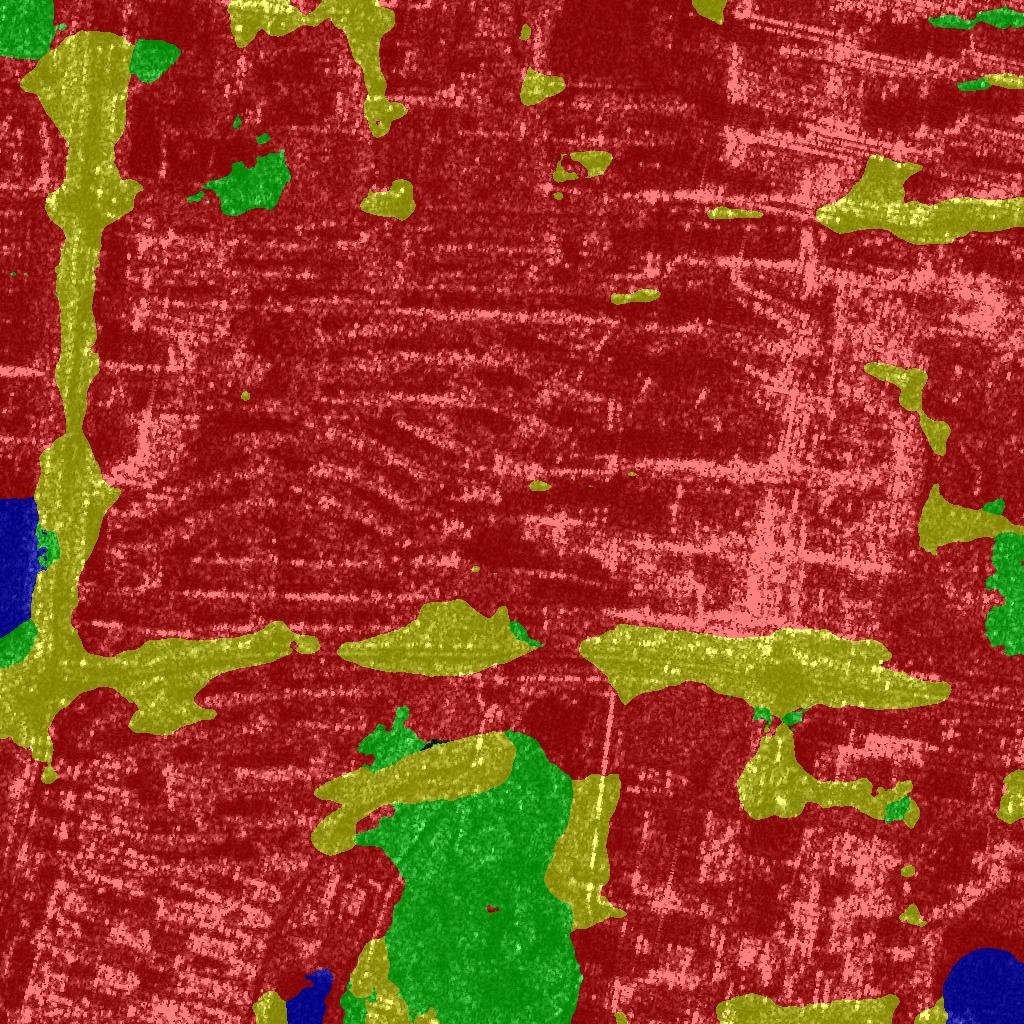}
              \includegraphics[width=.1\linewidth]{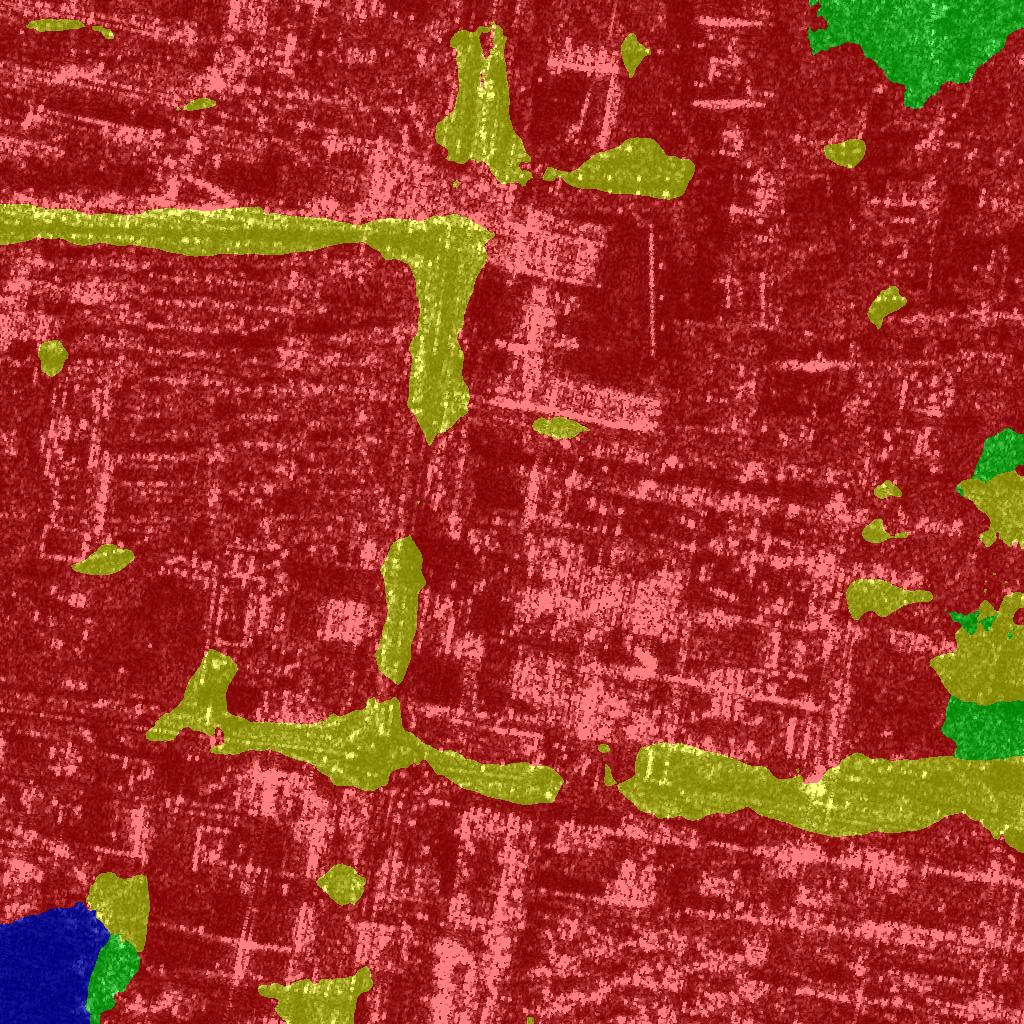}
	\end{minipage}\vspace{3pt}

    \begin{minipage}[c]{\linewidth}
	   \centering
            \small\rotatebox{90}{\hspace{7pt}CWSAM}
	       \includegraphics[width=.1\linewidth]{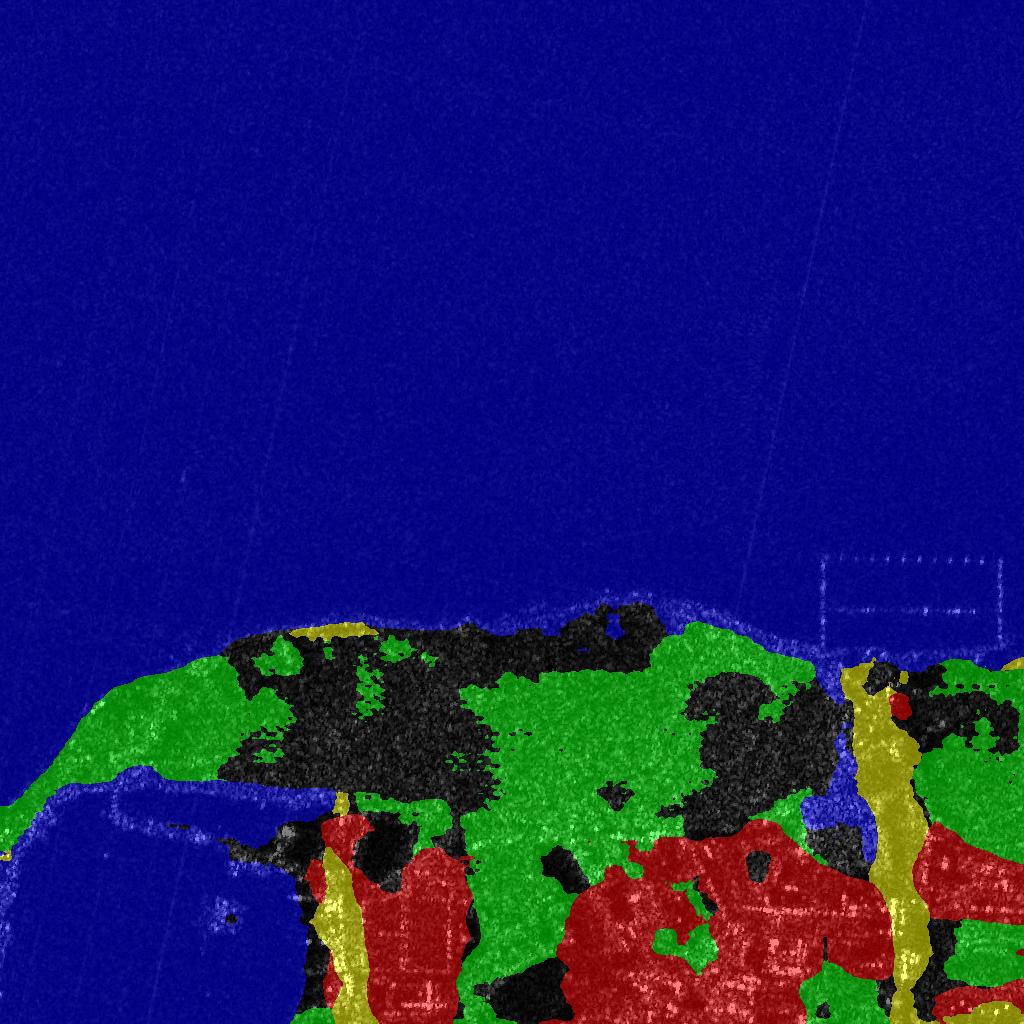}
	       \includegraphics[width=.1\linewidth]{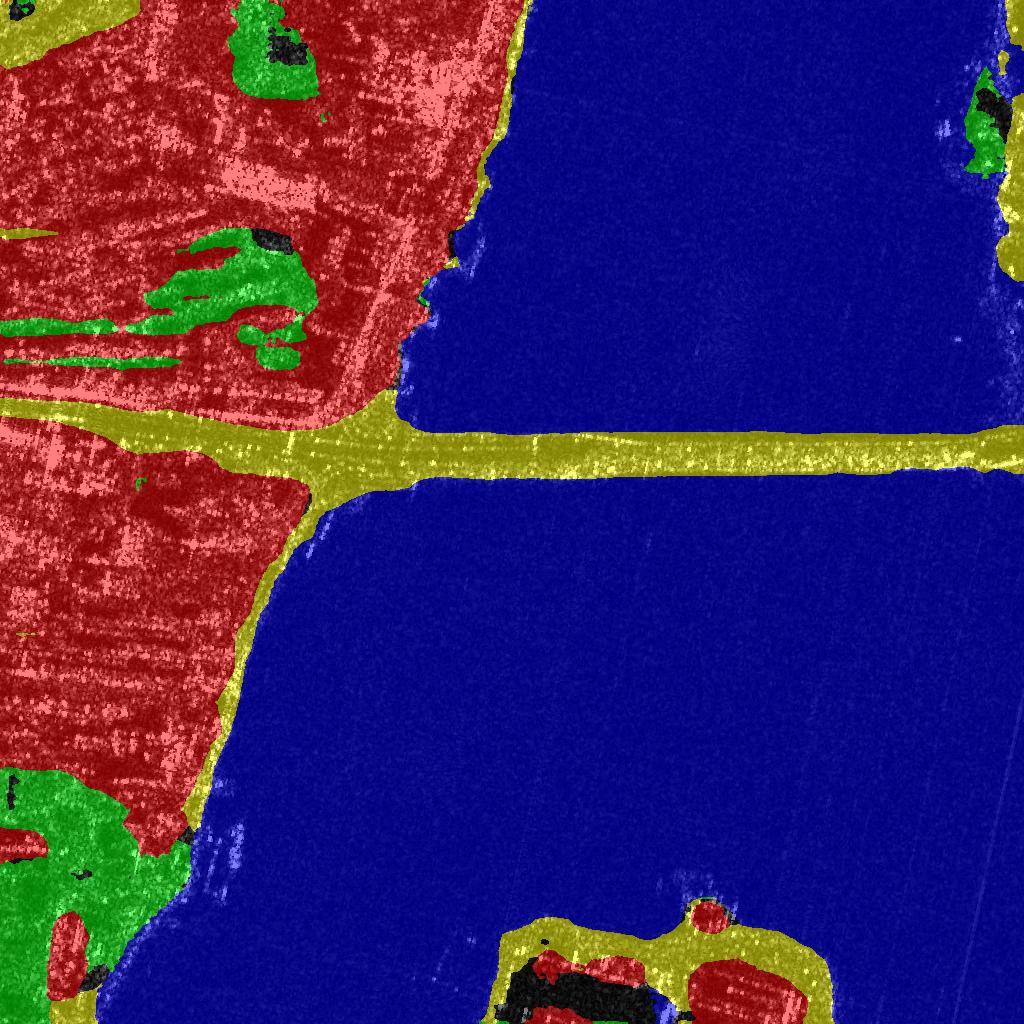}
              \includegraphics[width=.1\linewidth]{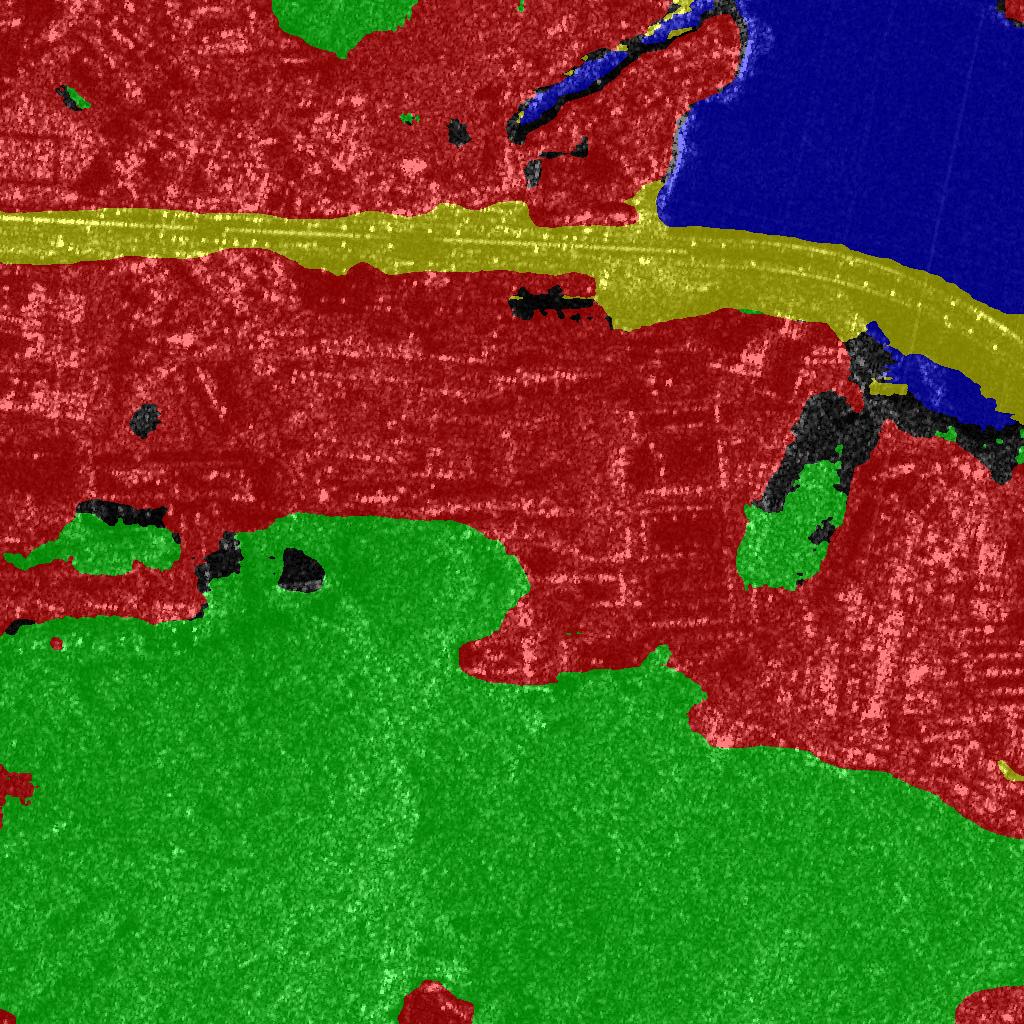}
              \includegraphics[width=.1\linewidth]{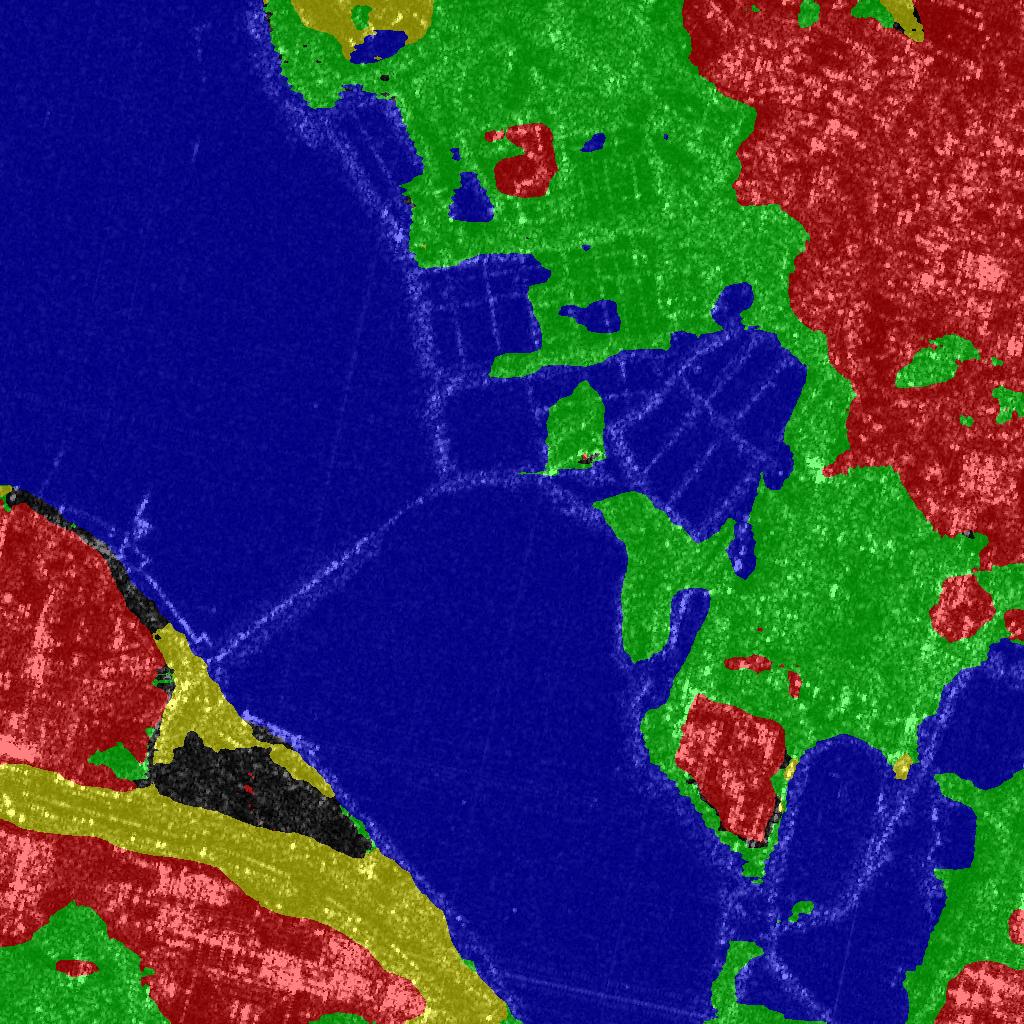}
              \includegraphics[width=.1\linewidth]{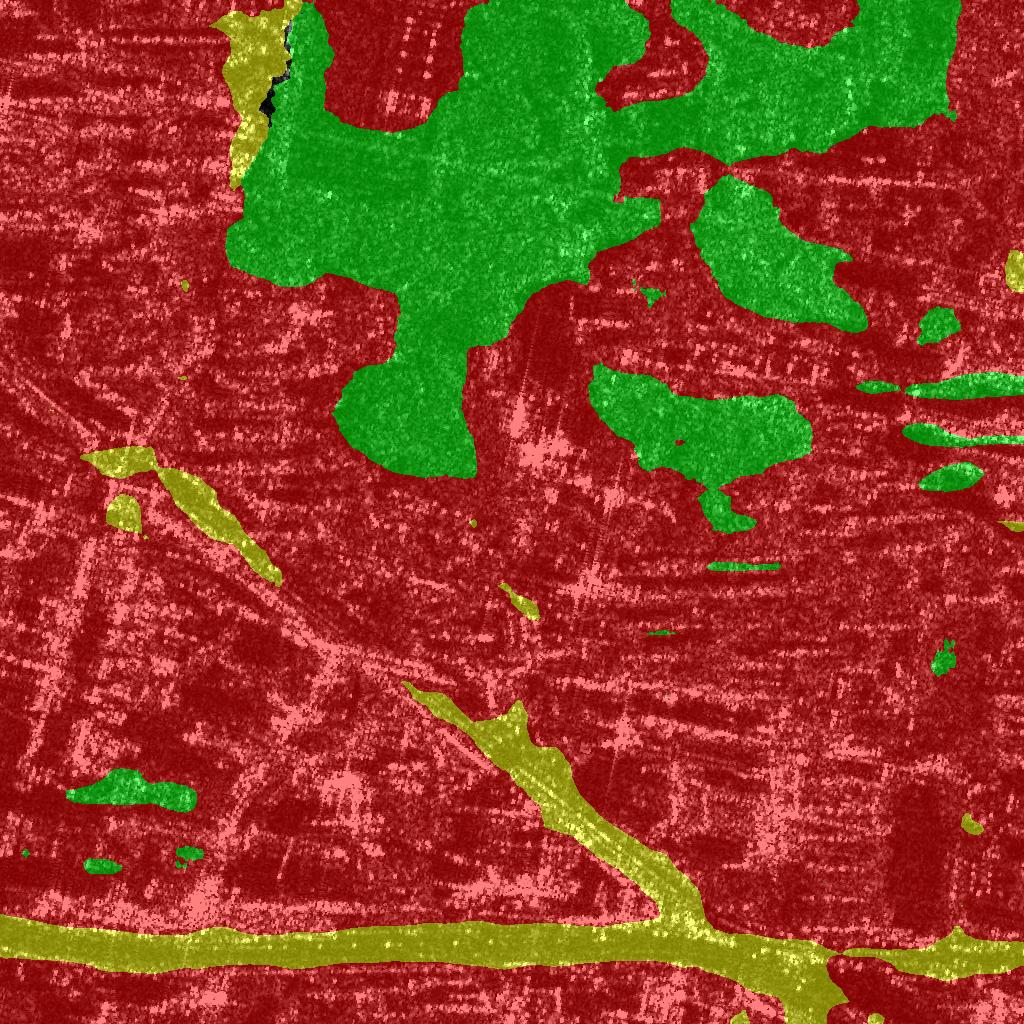}
              \includegraphics[width=.1\linewidth]{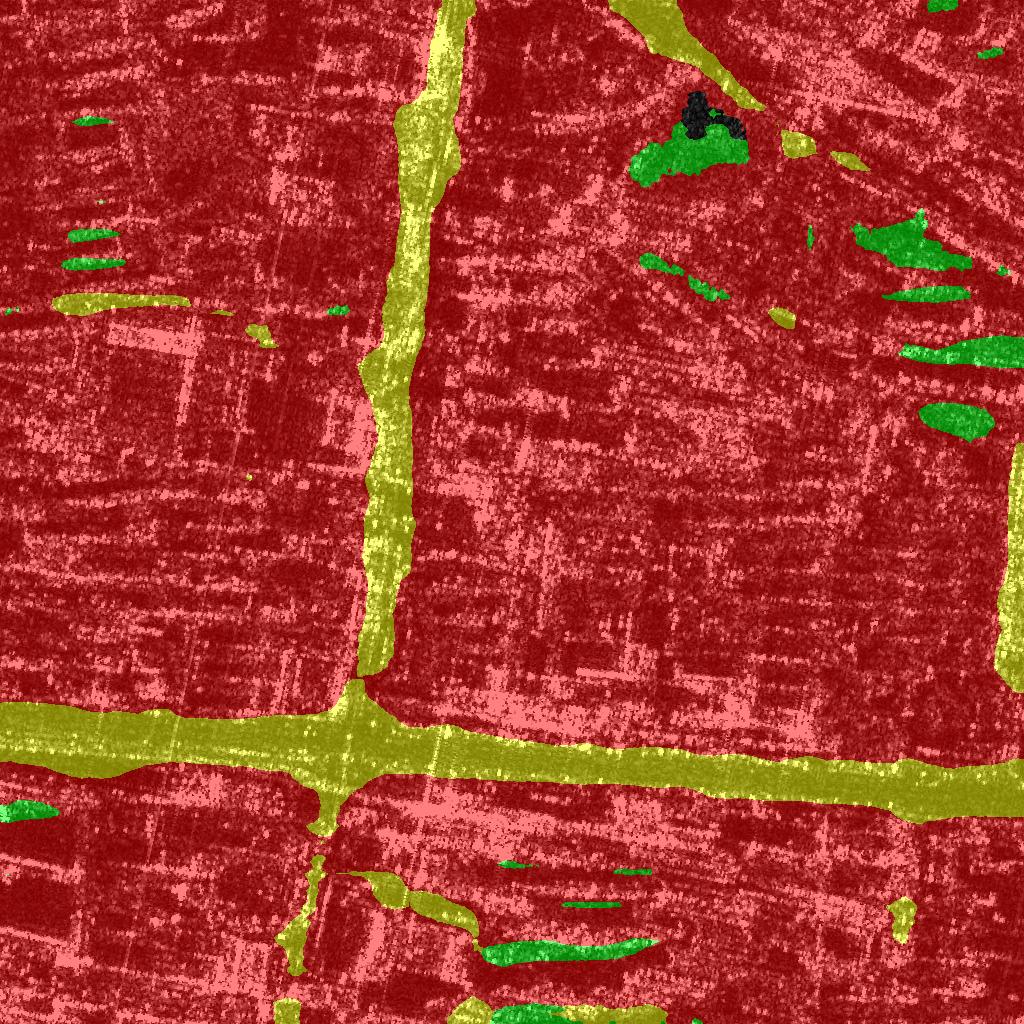}
              \includegraphics[width=.1\linewidth]{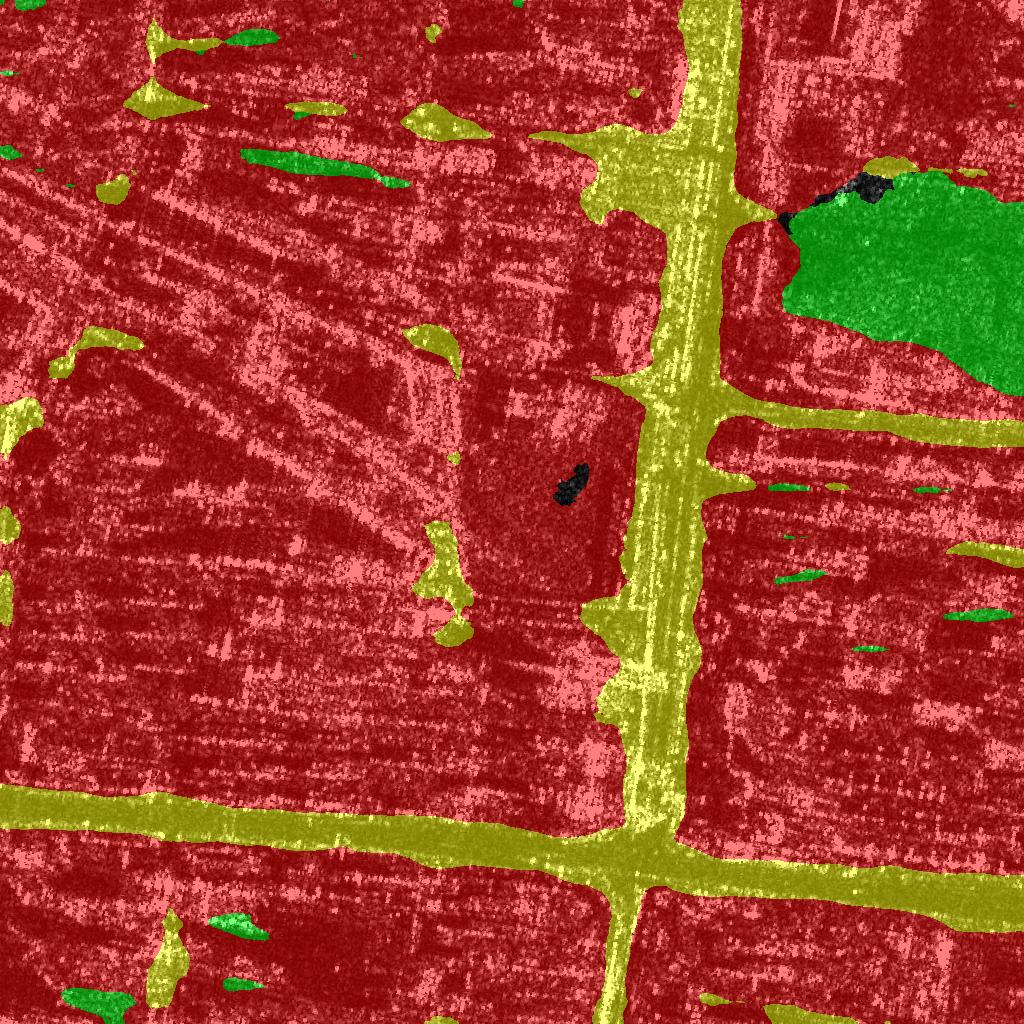}
              \includegraphics[width=.1\linewidth]{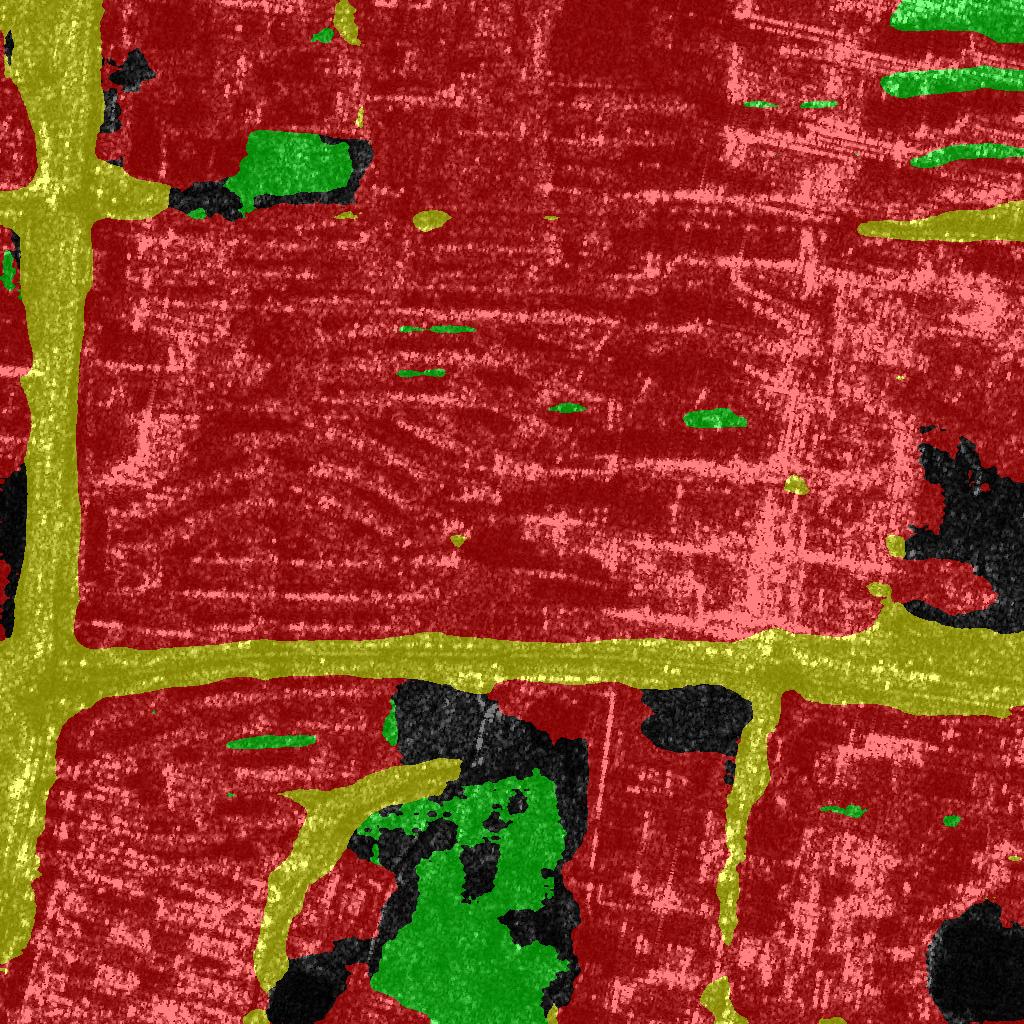}
              \includegraphics[width=.1\linewidth]{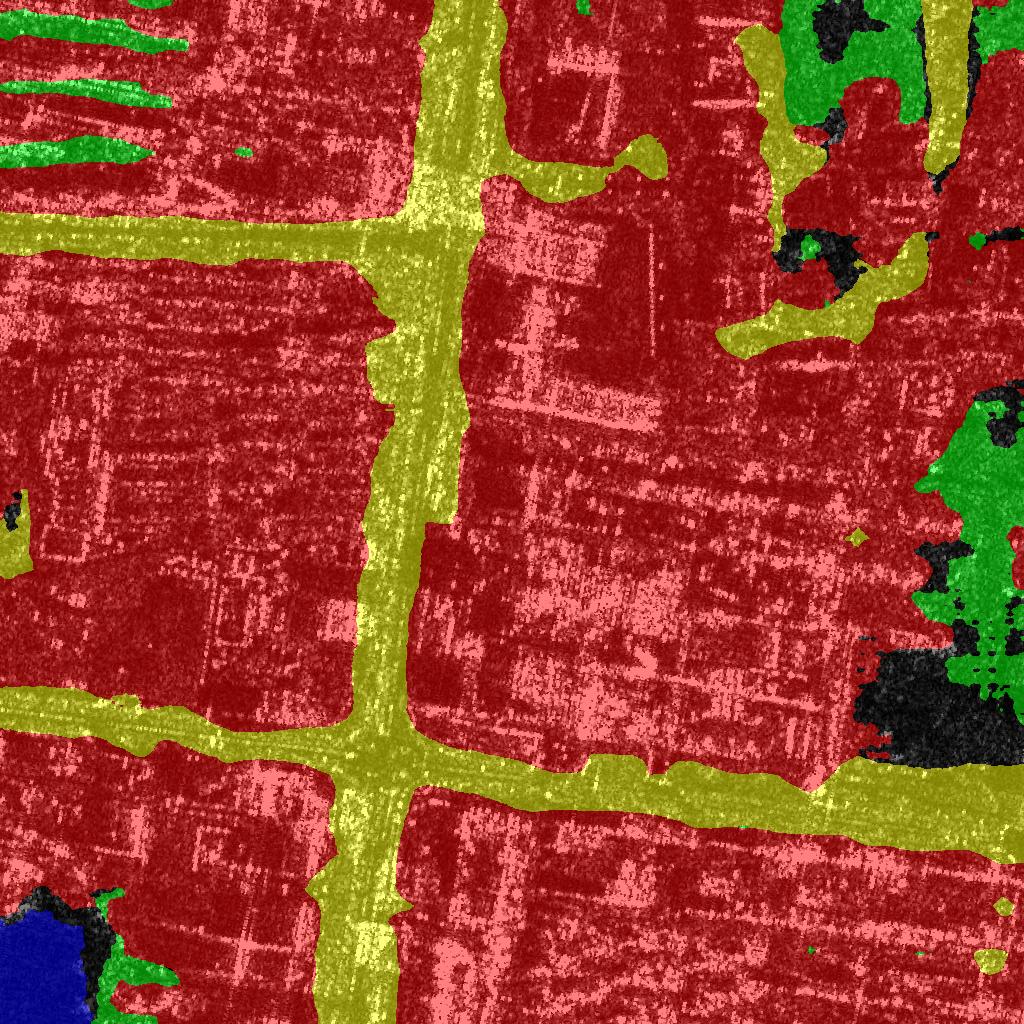}
	\end{minipage}\vspace{3pt}

    \begin{minipage}[t]{\linewidth}
	   \centering
            \small\rotatebox{90}{Ground Truth}
	       \includegraphics[width=.1\linewidth]{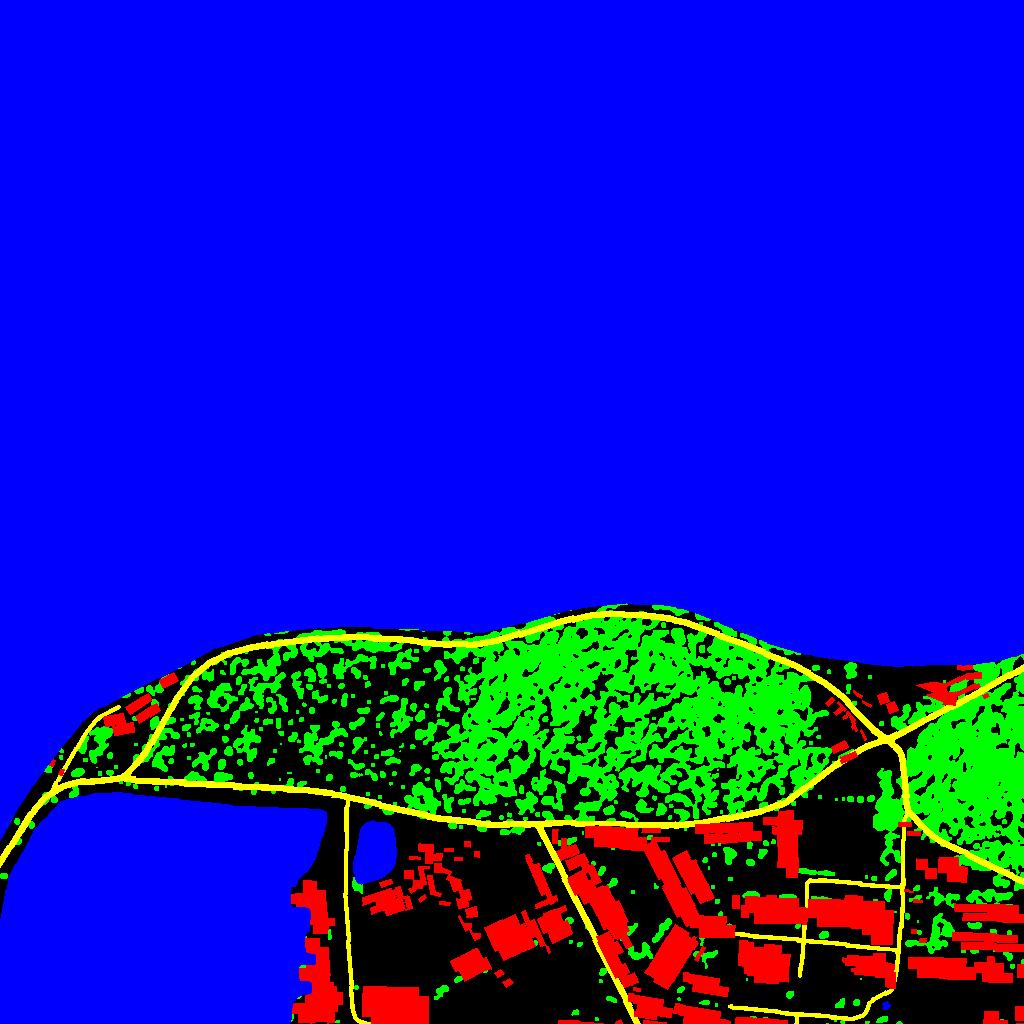}
	       \includegraphics[width=.1\linewidth]{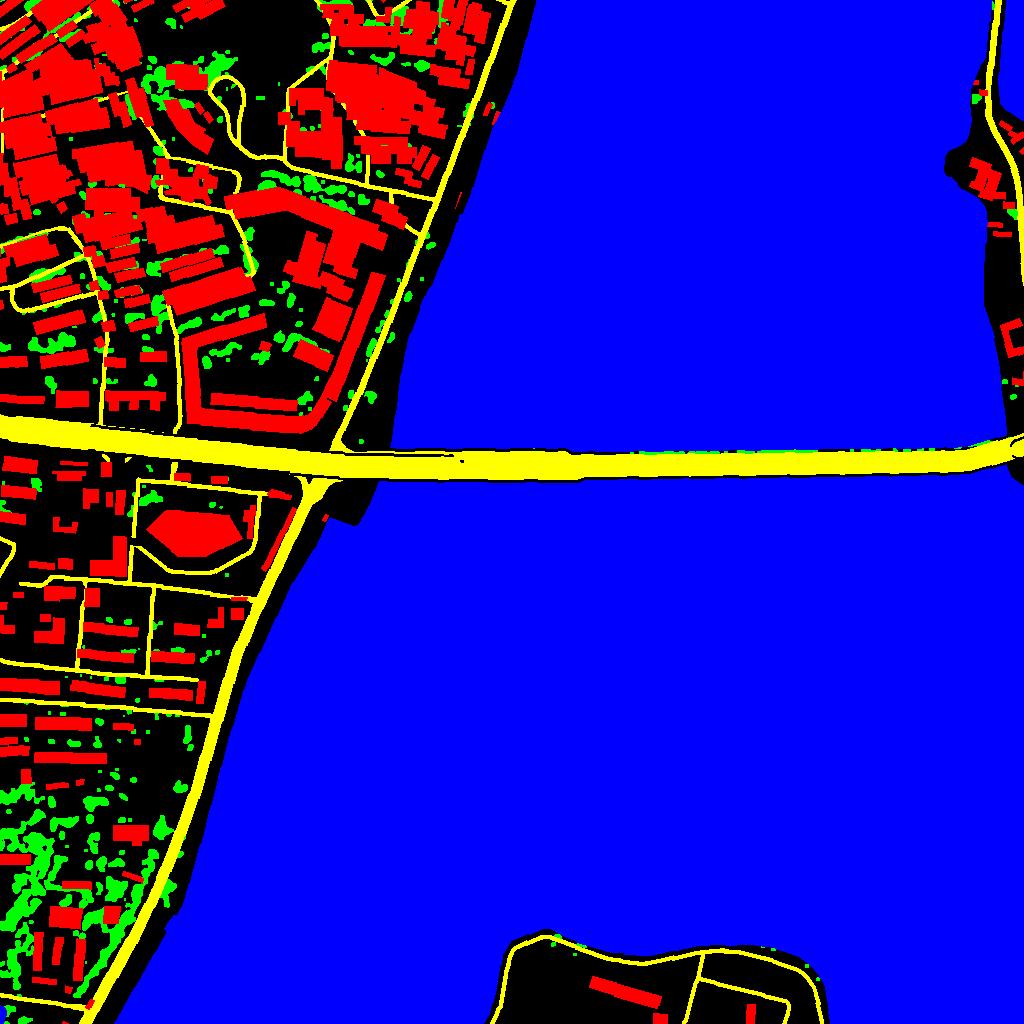}
              \includegraphics[width=.1\linewidth]{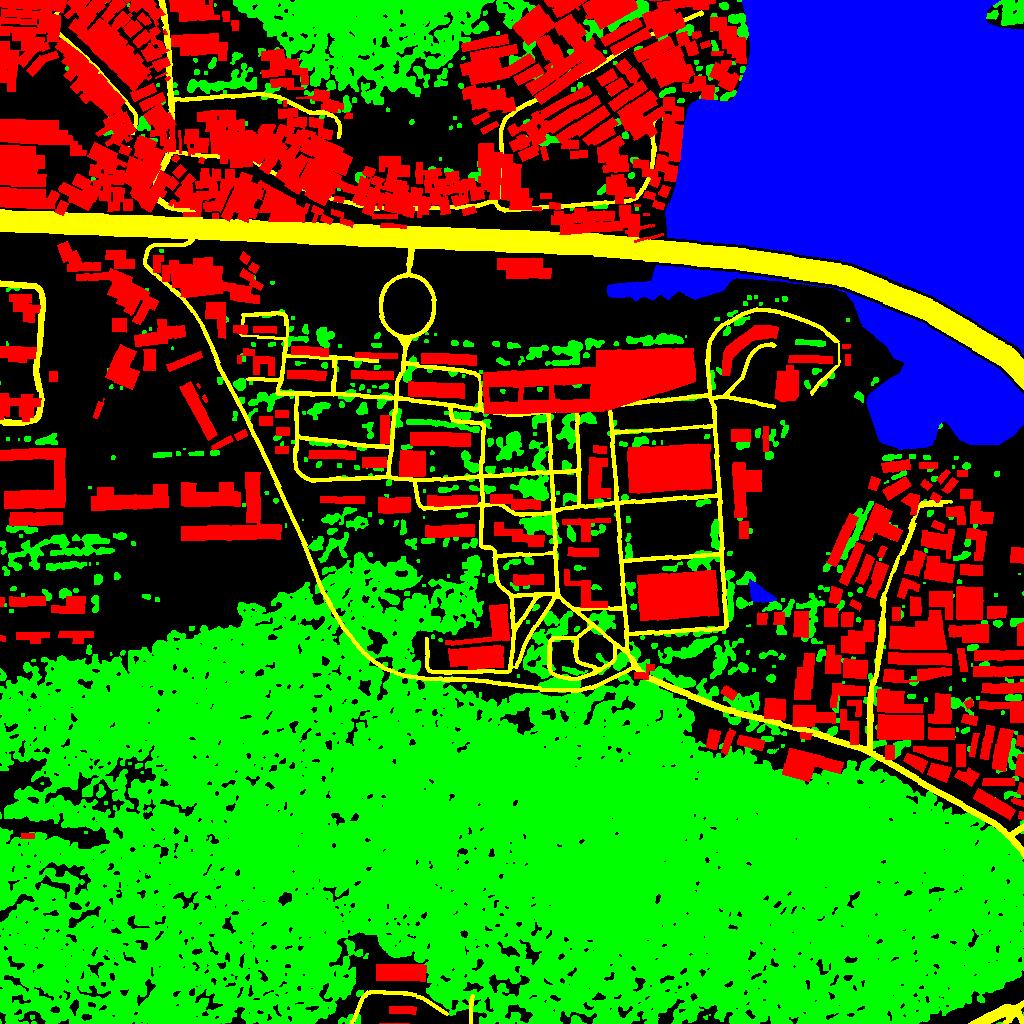}
              \includegraphics[width=.1\linewidth]{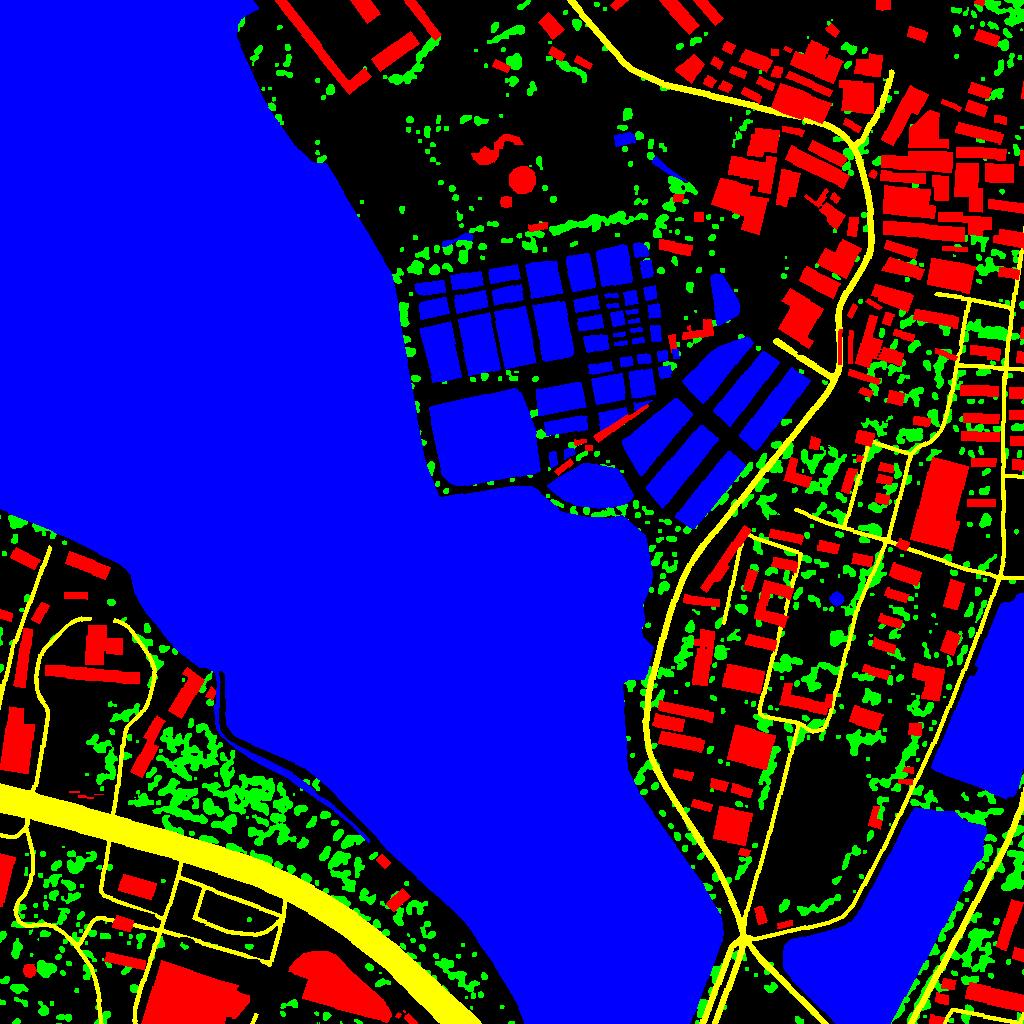}
              \includegraphics[width=.1\linewidth]{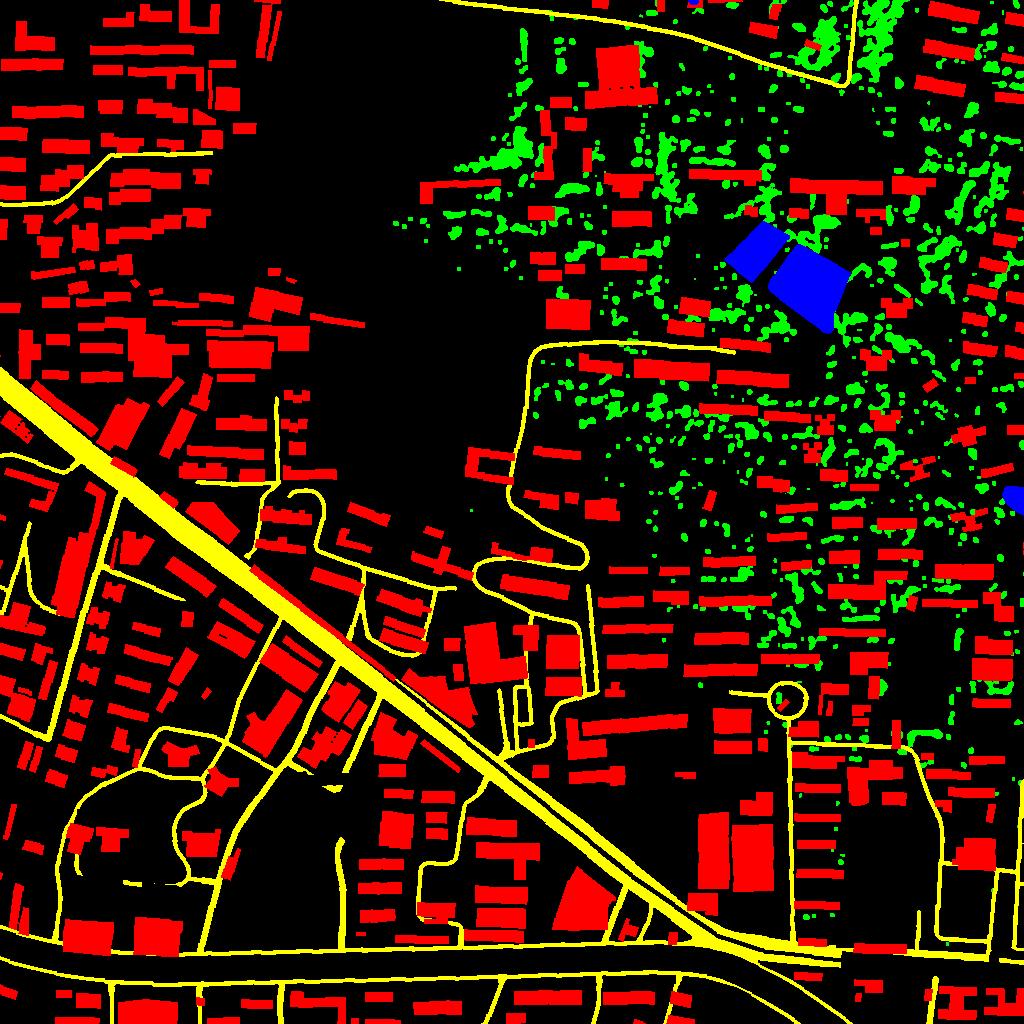}
              \includegraphics[width=.1\linewidth]{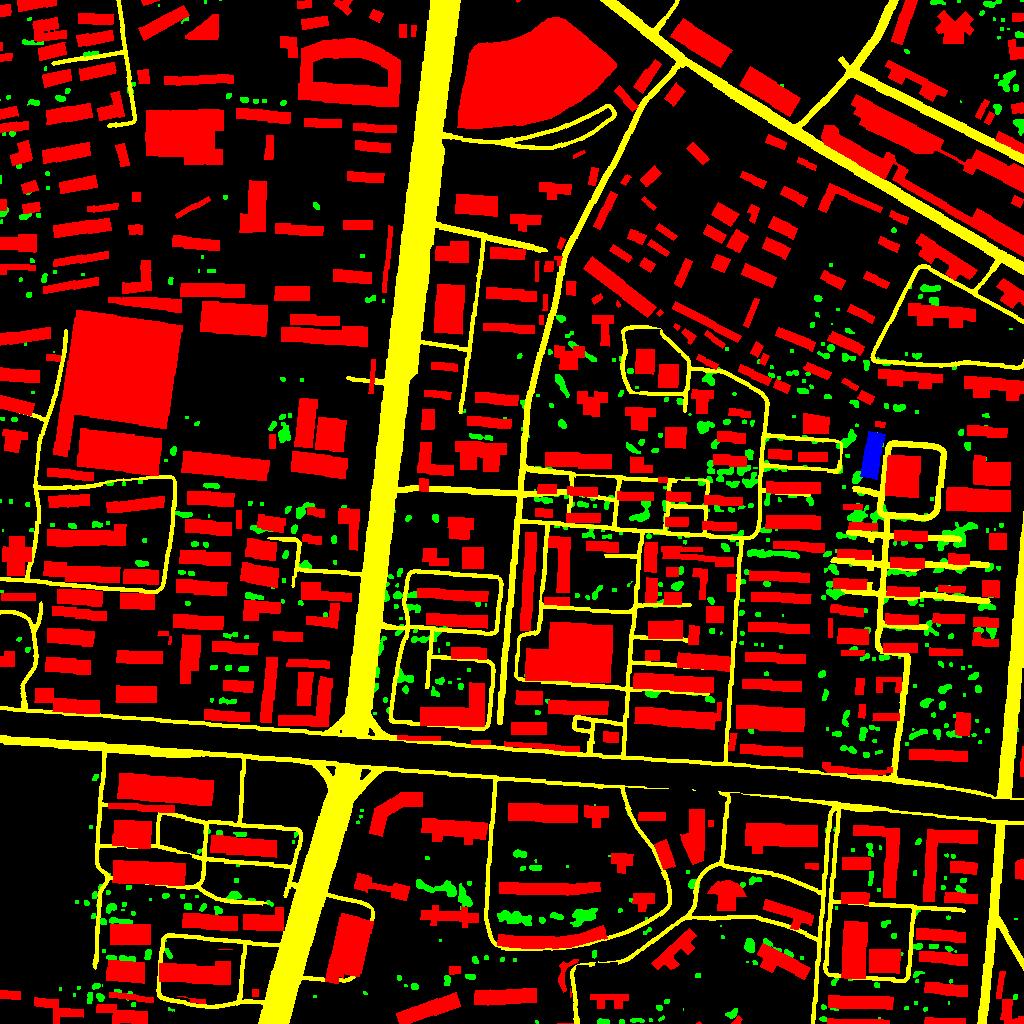}
              \includegraphics[width=.1\linewidth]{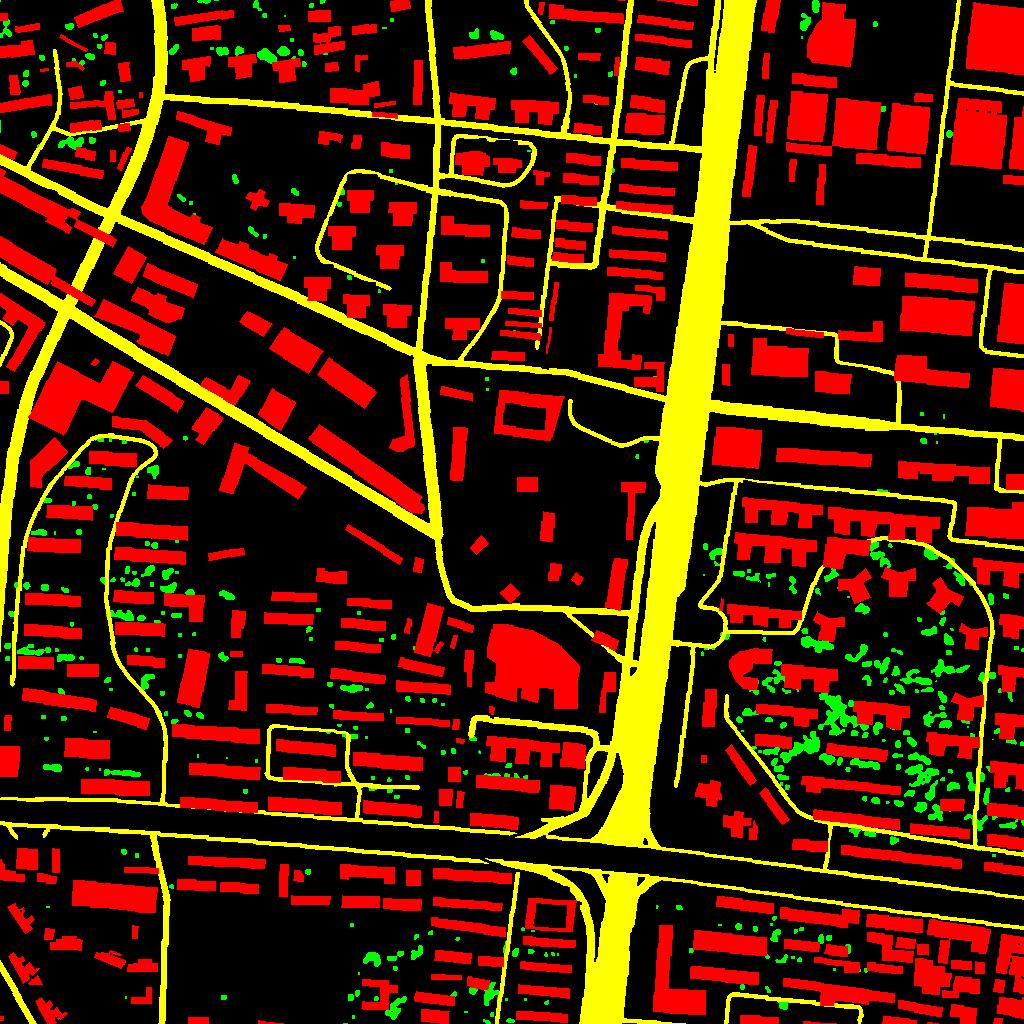}
              \includegraphics[width=.1\linewidth]{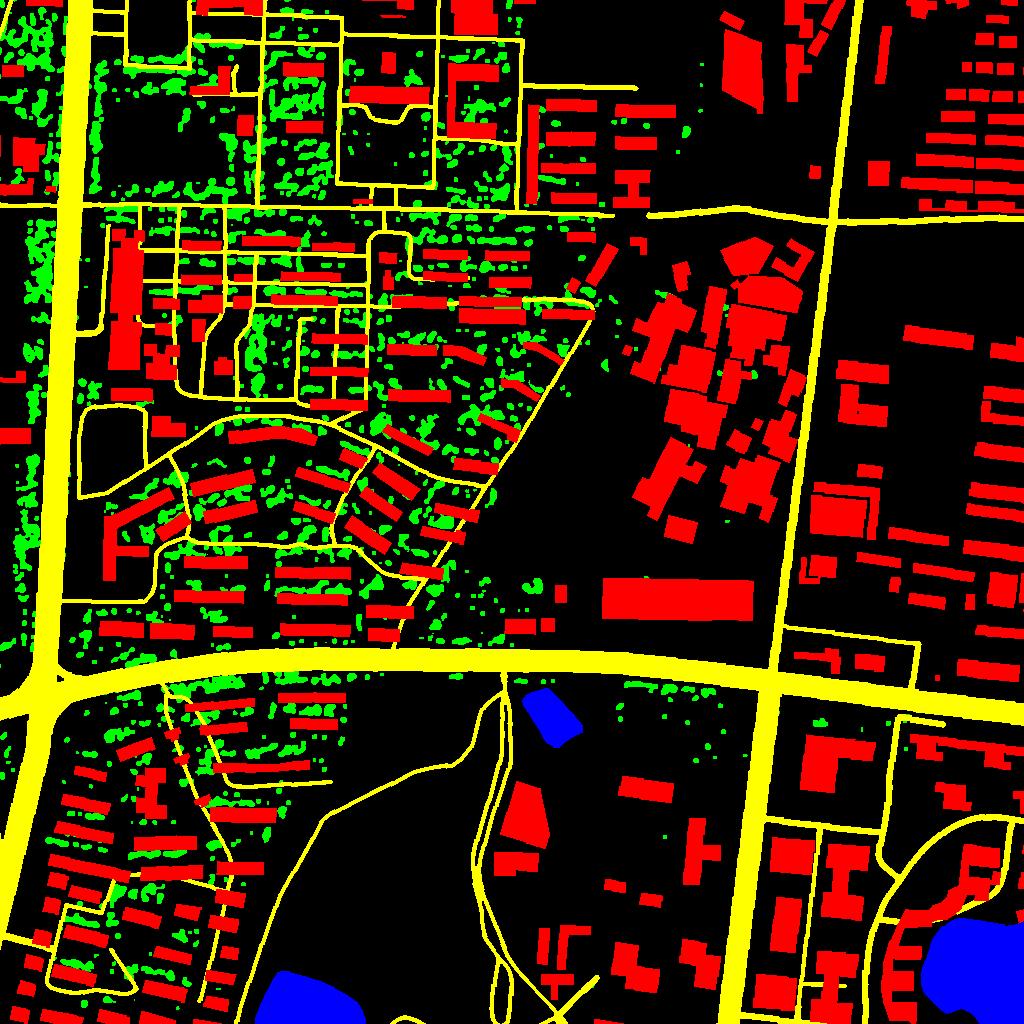}
              \includegraphics[width=.1\linewidth]{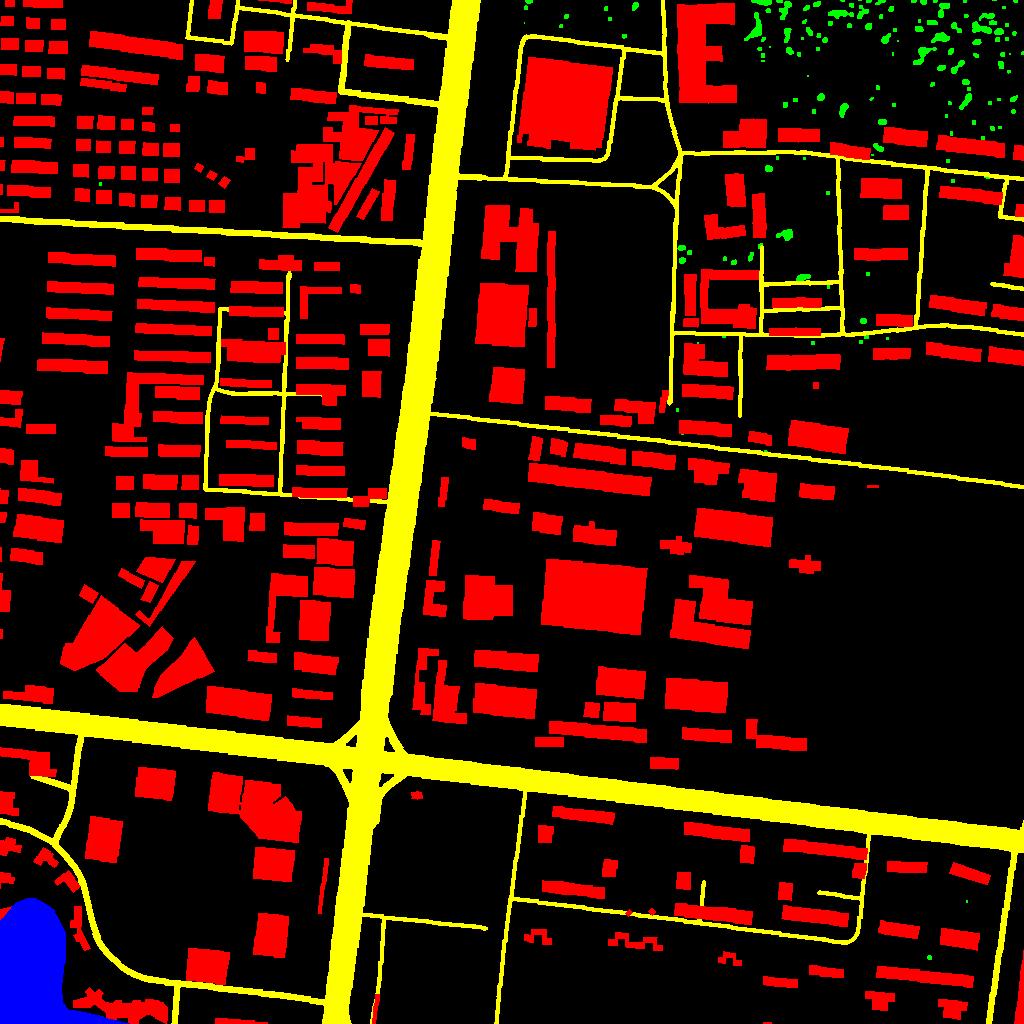}
	\end{minipage}\vspace{5pt}
 
 \begin{tikzpicture}
    
    \fill[blue] (-10,5) rectangle (-9.5,5.5);
    \node at (-9,5.25) {water};
    
    \fill[red] (-8.4,5) rectangle (-7.9,5.5);
    \node at (-7.2,5.25) {building};
    
    \fill[green] (-6.4,5) rectangle (-5.9,5.5);
    \node at (-5.1,5.25) {vegetation};
    
    \fill[yellow] (-4.1,5) rectangle (-3.6,5.5);
    \node at (-3.1,5.25) {road};

\end{tikzpicture}

\caption{Qualitative results of landcover classification masks. The compared visualizations between the proposed method and the best state-of-the-art algorithm Segformer on FUSAR-Map1.0 dataset. The predicted masks in the second and third rows are overlapped with the original SAR image for visualization by using a transparency of 0.5.}
\end{figure*}

\subsection{Comparison with the State-of-the-Art}
In this section, the proposed method is compared with multiple state-of-the-art semantic segmentation algorithms to evaluate its efficiency. Diverse approaches of different backbones and architectures are involved to build a comprehensive investigation including Deeplabv3 \cite{AtrousConvolution} series with ResNet \cite{ResNet} convolution-based modules, HRNet \cite{HRNet} series and SegFormer \cite{SegFormer} networks based on Vision Transformer encoder. 

In our experiments, Deeplabv3, feature enhanced deeplabv3 \cite{ShiFCWX21} and deeplabv3+ \cite{EncoderDecoderAtrous} employ encoder-decoder structures and convolution module for leveraging heavy image features to improve model performance. HRNet networks maintain high resolution feature of multi-scale semantic information and SegFormer utilizes the powerful Transformer backbone for image encoder and designs a light-weighted decoder to obtain outstanding segmentation performance. 

Even though various other algorithms have their own well-designed architectures to enhance model performance, the proposed method outperforms other state-of-the-art methods on multiple standard evaluation metrics for both FUSAR-Map1.0 and FUSAR-Map2.0 datasets. mIOU is regarded as main metric in our experiments to evaluate the average segmentation performance of all categories and also to concern about these difficult categories. 

(1) Results on FUSAR-Map1.0: The general evaluation metrics are superior to other algorithms including mIoU, OA, Accuracy, Precision and mDice as shown in Table I. The proposed algorithm achieves the highest scores on all evaluation metrics, indicating its optimal performance from all assessment perspectives. Additionally, the most referential evaluation metric, mIOU, surpasses the second highest algorithm SegFormer with MiT-B3 backbone \cite{SegFormer} by 3.67$\%$.

\begin{figure*}[!b]
\centering
    \begin{minipage}[t]{\linewidth}
	   \centering
            \small\rotatebox{90}{\hspace{3pt}SAR Images}
	       \includegraphics[width=.1\linewidth]{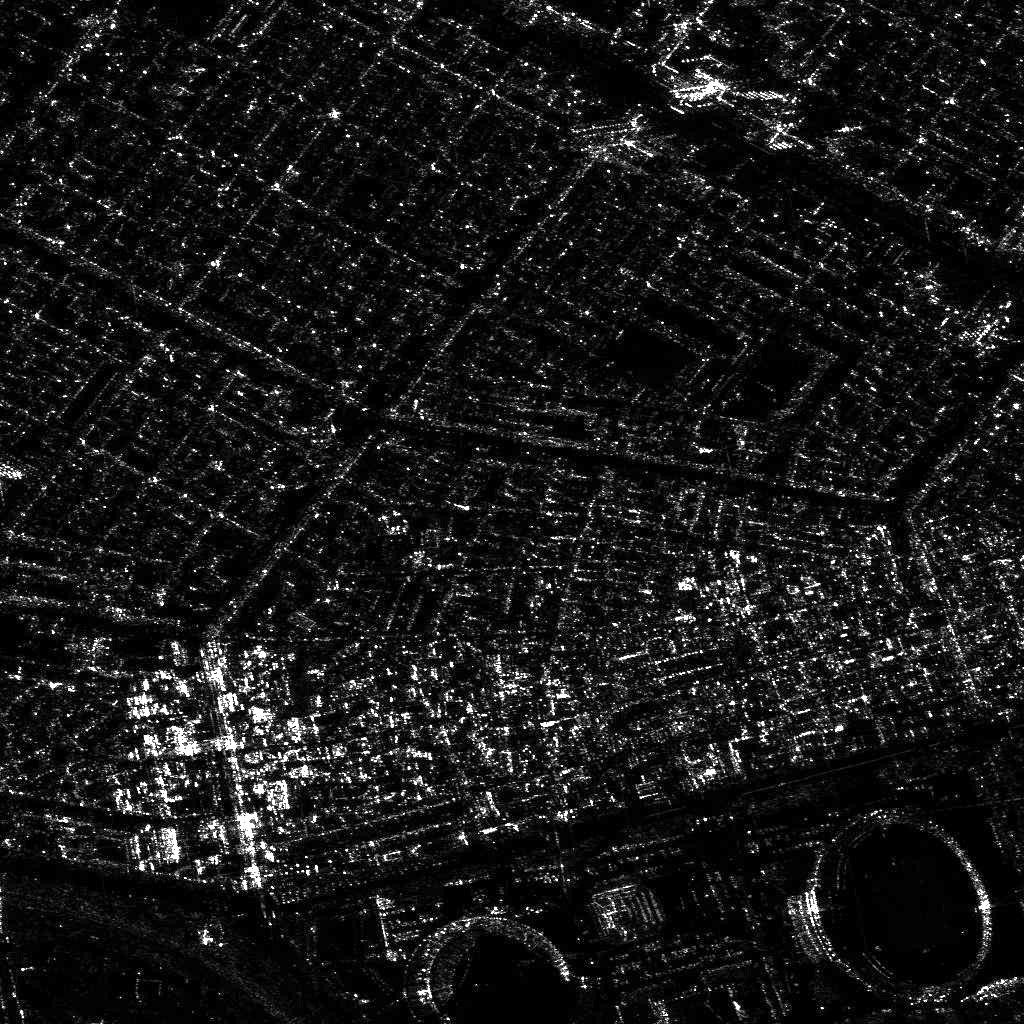}
	       \includegraphics[width=.1\linewidth]{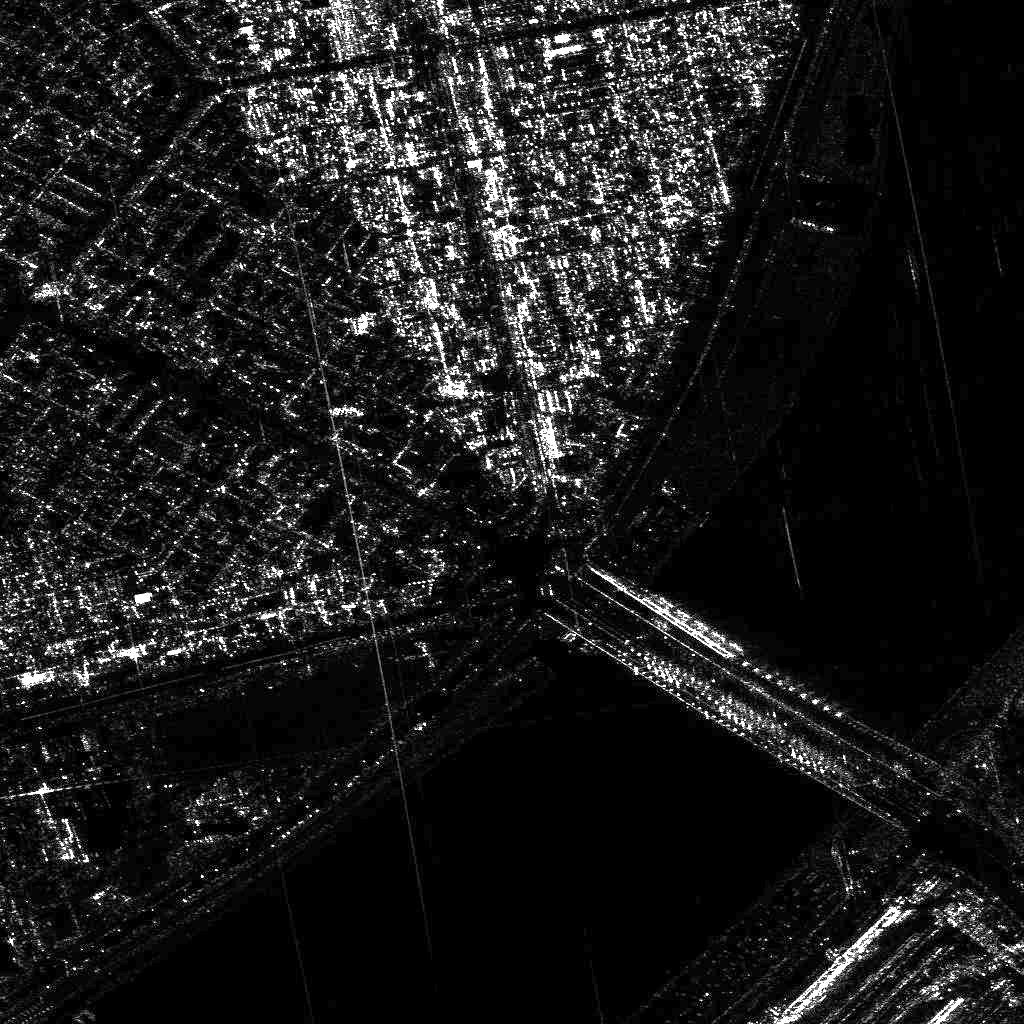}
              \includegraphics[width=.1\linewidth]{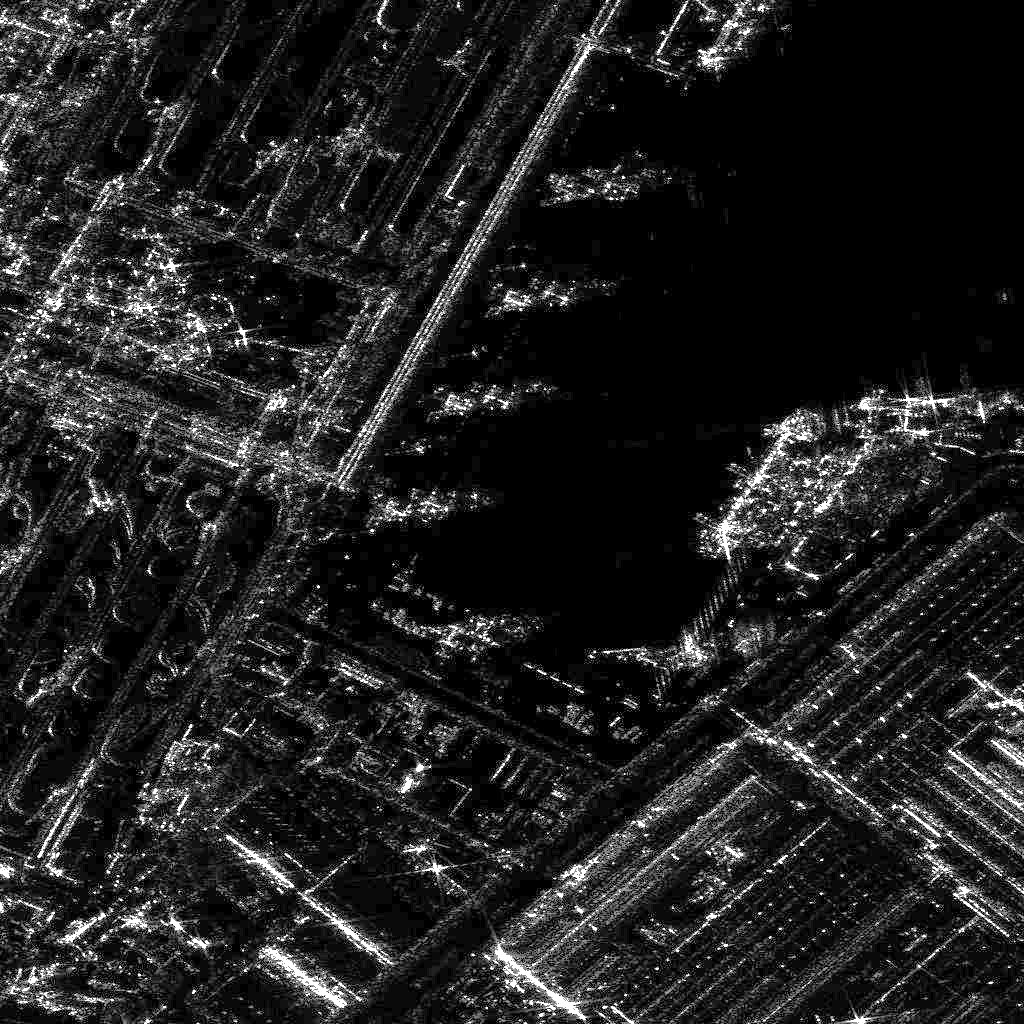}
              \includegraphics[width=.1\linewidth]{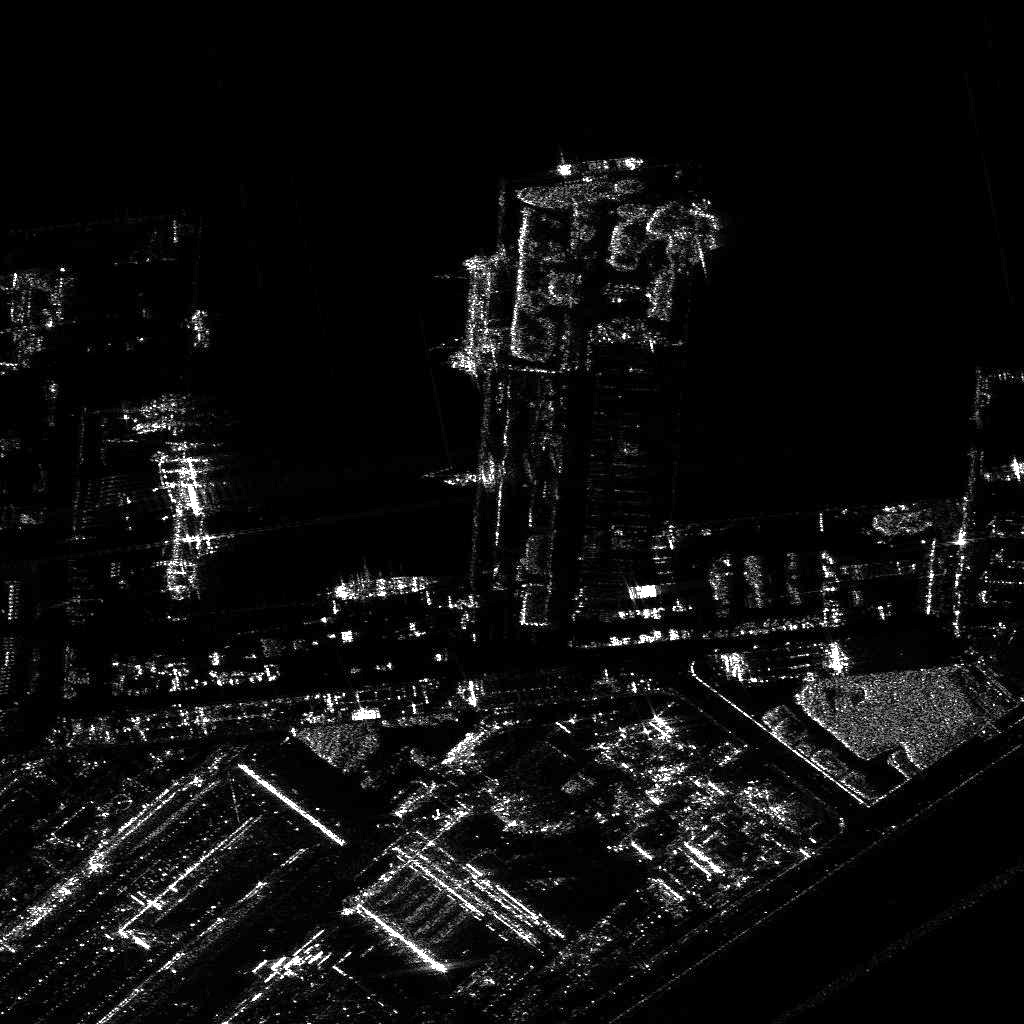}
              \includegraphics[width=.1\linewidth]{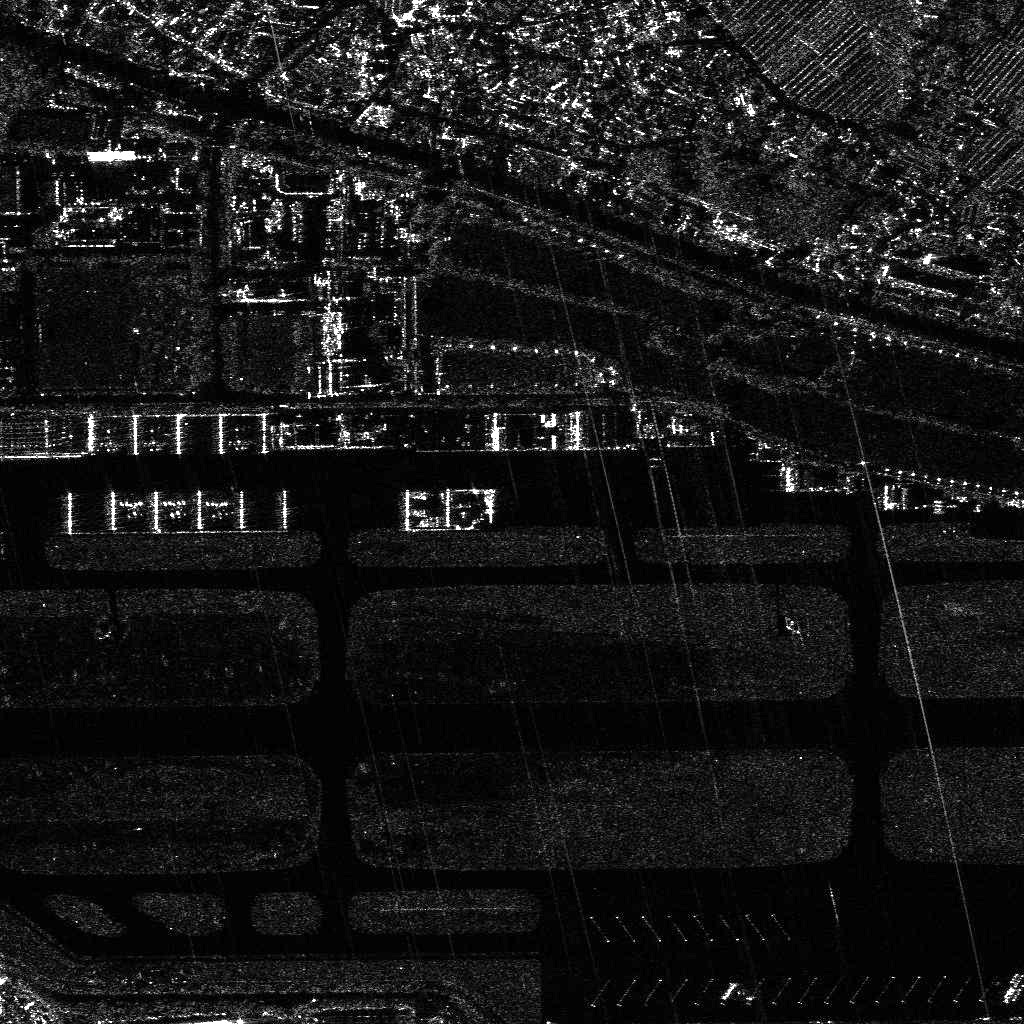}
              \includegraphics[width=.1\linewidth]{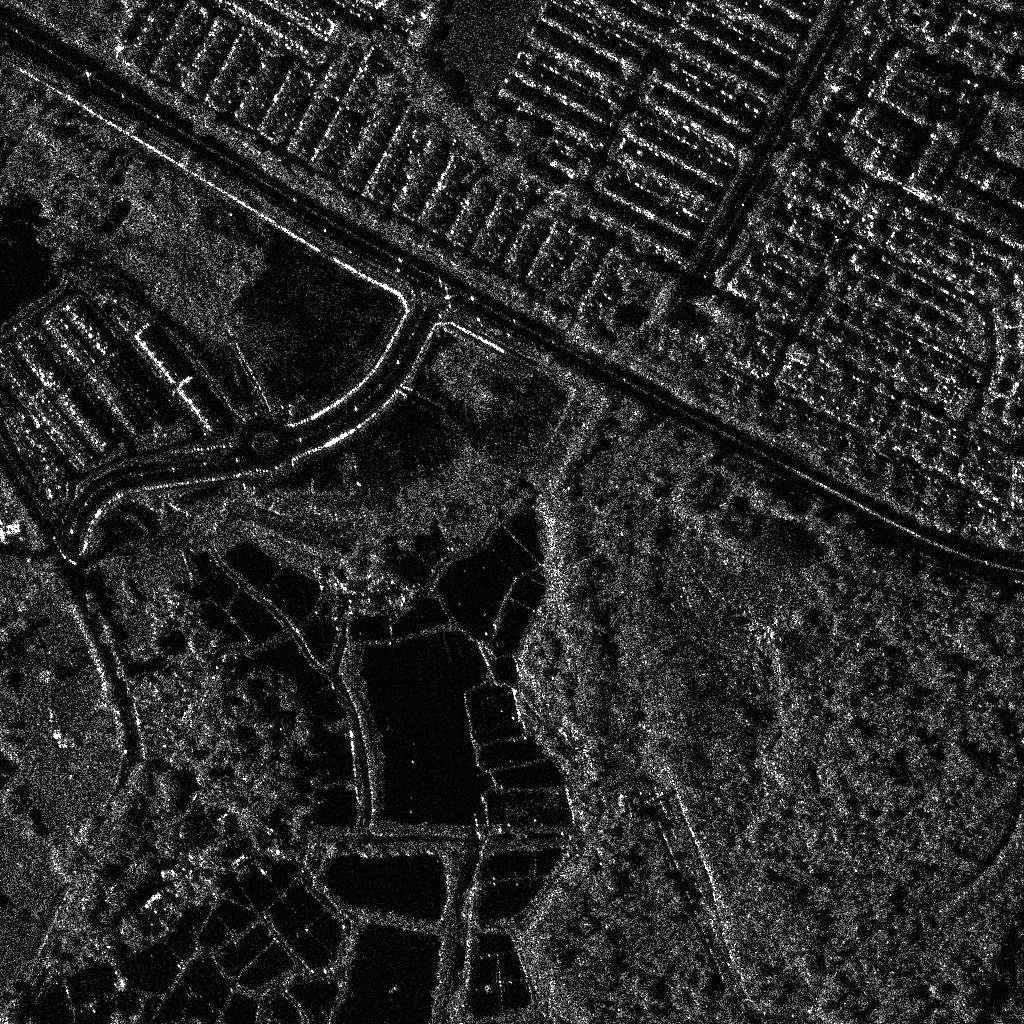}
              \includegraphics[width=.1\linewidth]{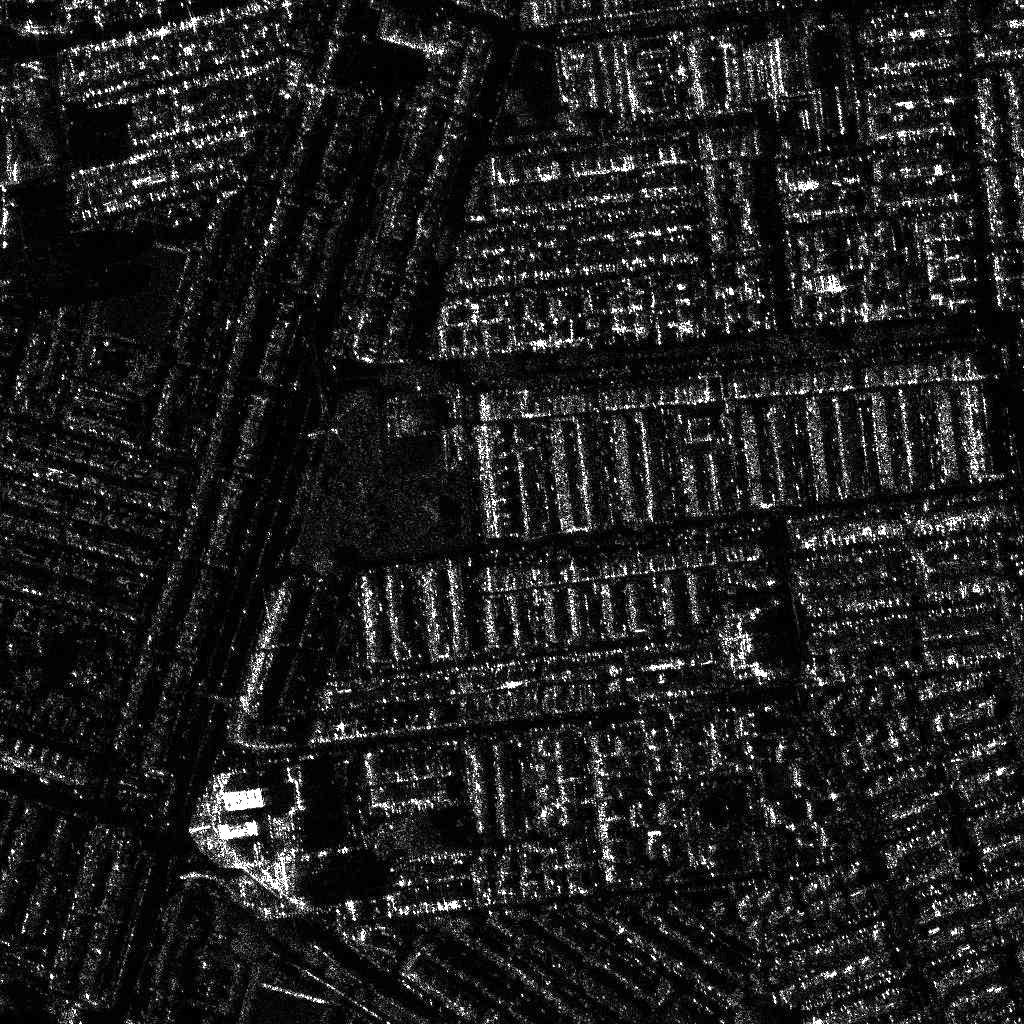}
              \includegraphics[width=.1\linewidth]{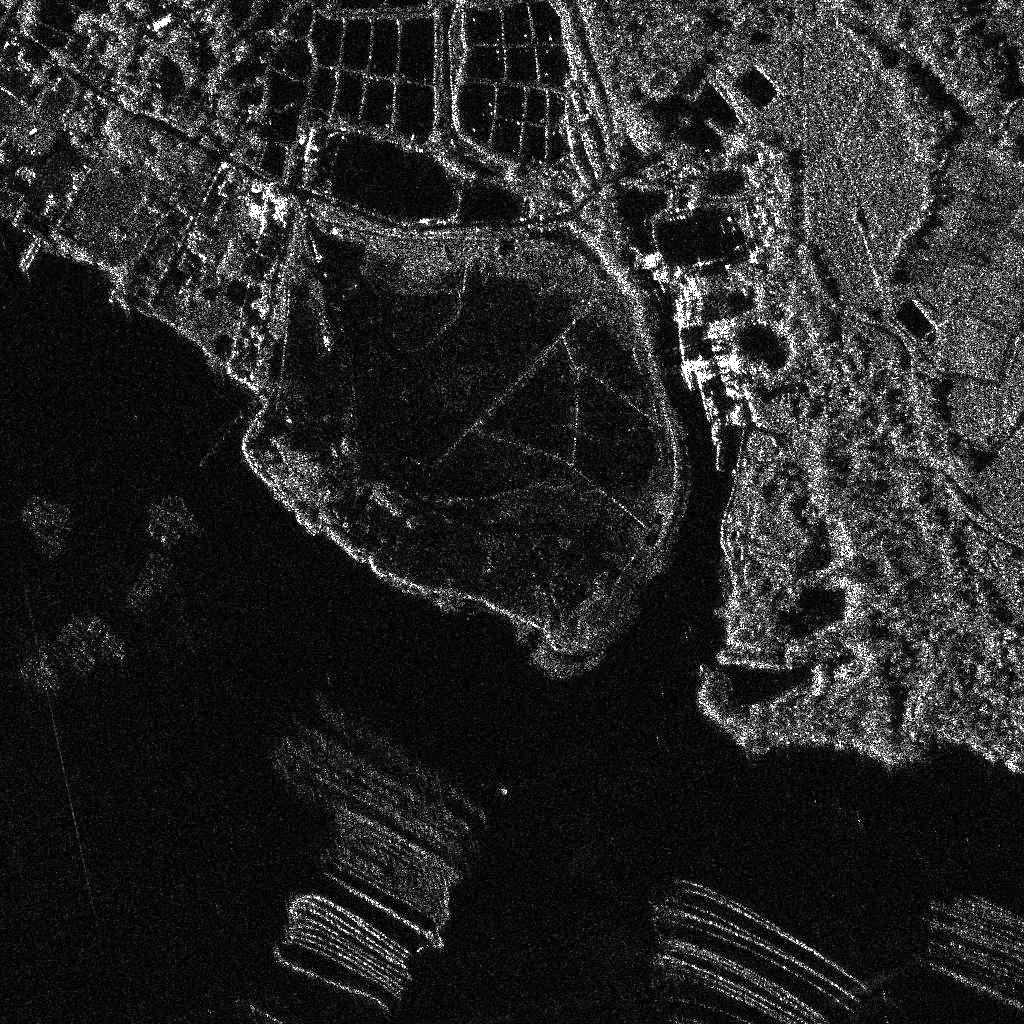}
              \includegraphics[width=.1\linewidth]{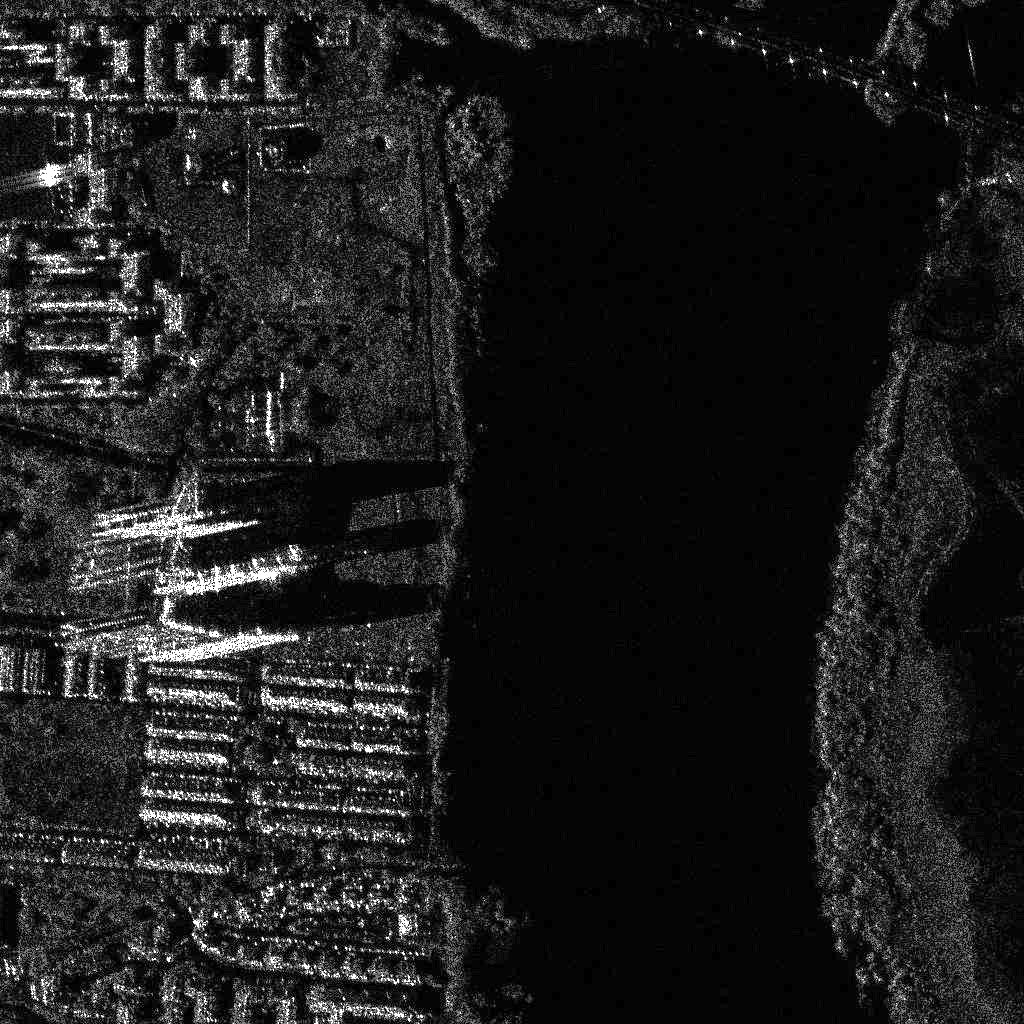}
	\end{minipage}\vspace{3pt}

    \begin{minipage}[t]{\linewidth}
	   \centering
                \small\rotatebox{90}{\hspace{3pt}HRNet-48}
	       \includegraphics[width=.1\linewidth]{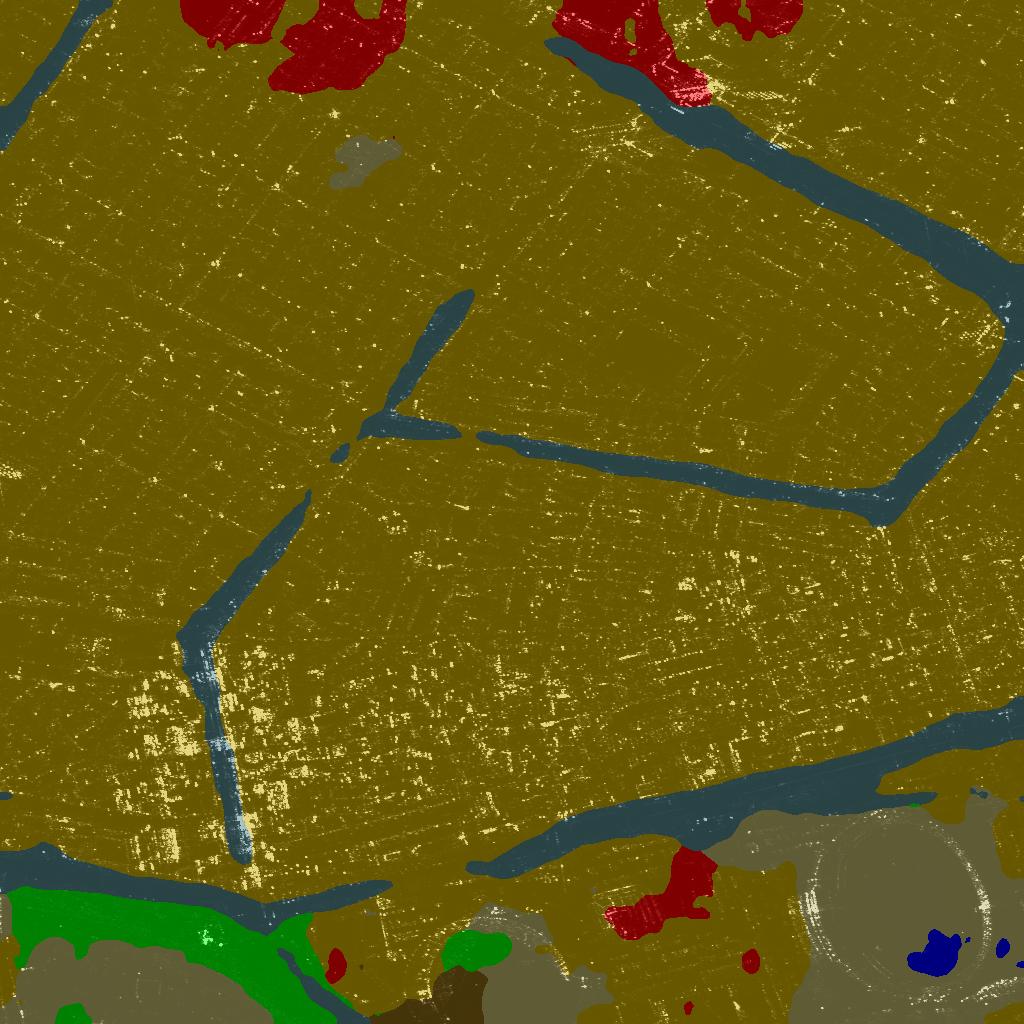}
	       \includegraphics[width=.1\linewidth]{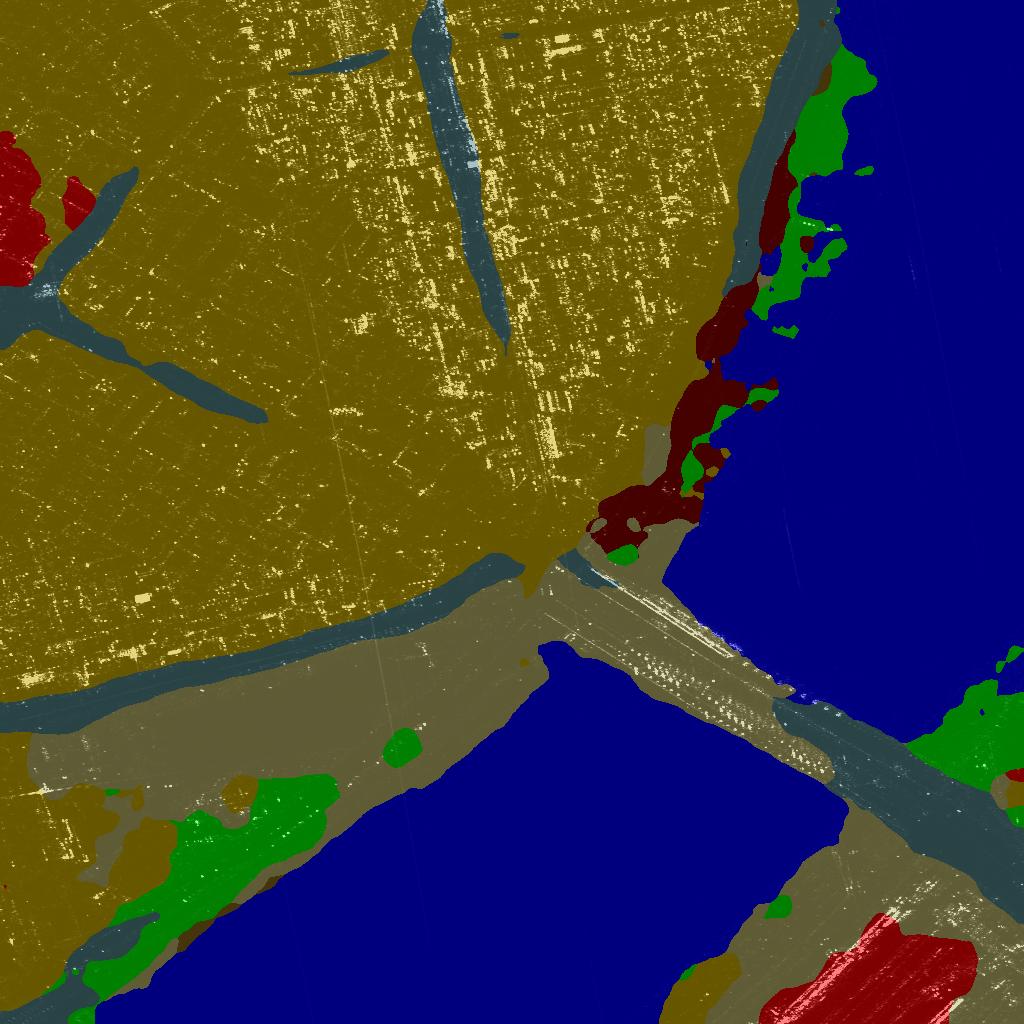}
              \includegraphics[width=.1\linewidth]{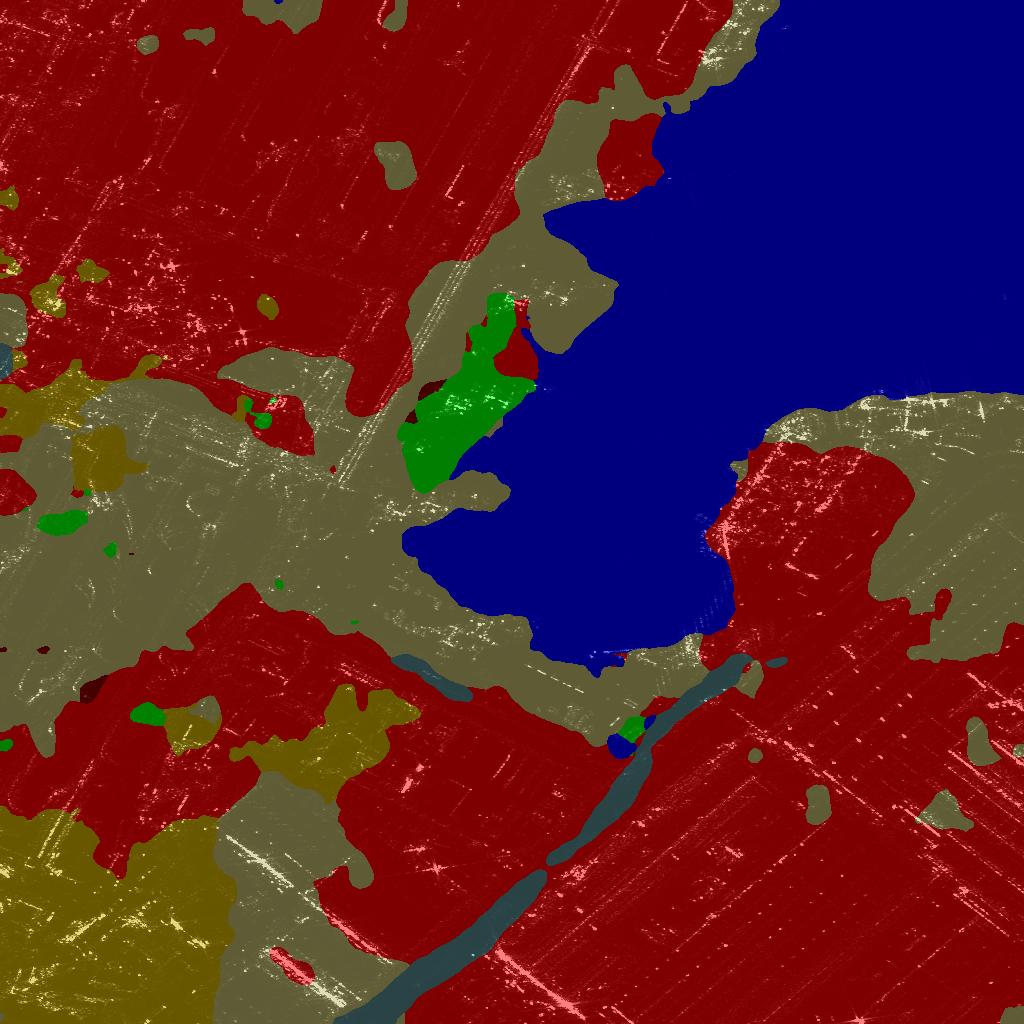}
              \includegraphics[width=.1\linewidth]{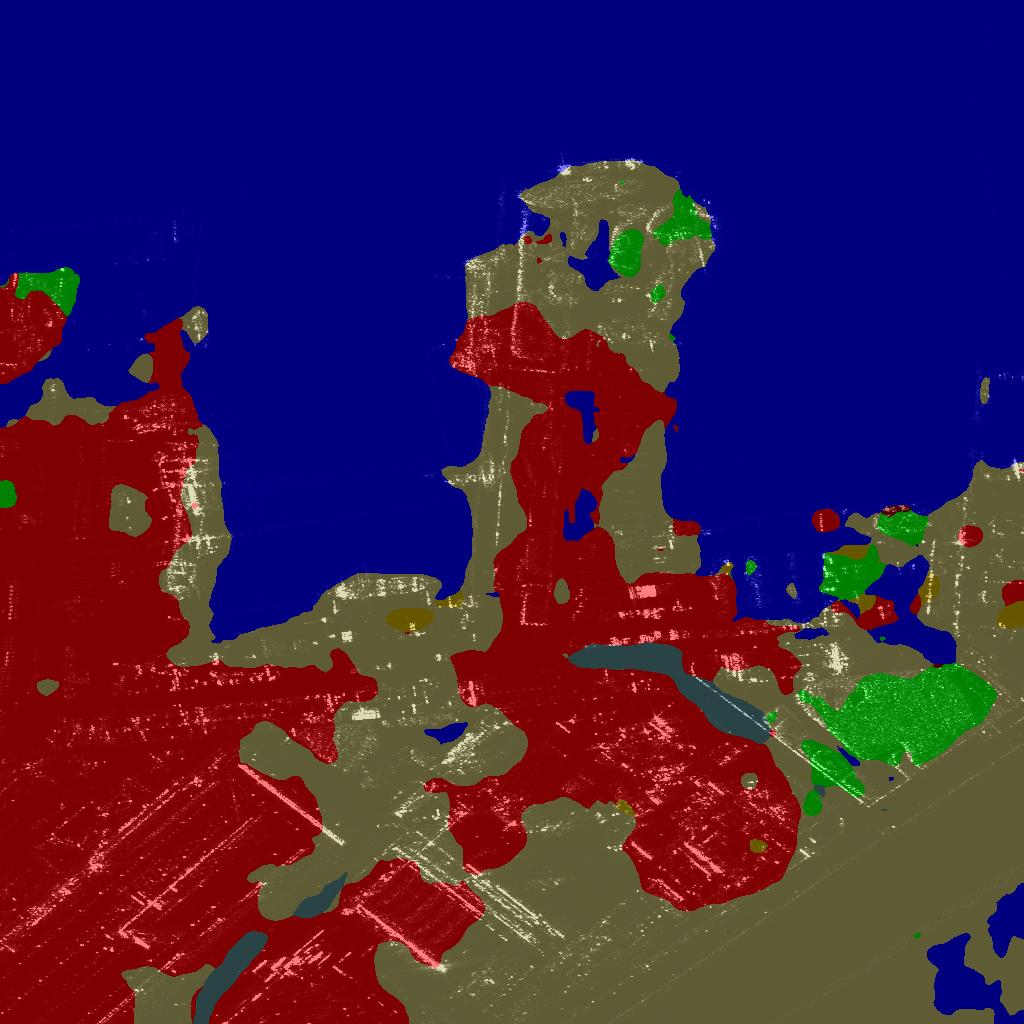}
              \includegraphics[width=.1\linewidth]{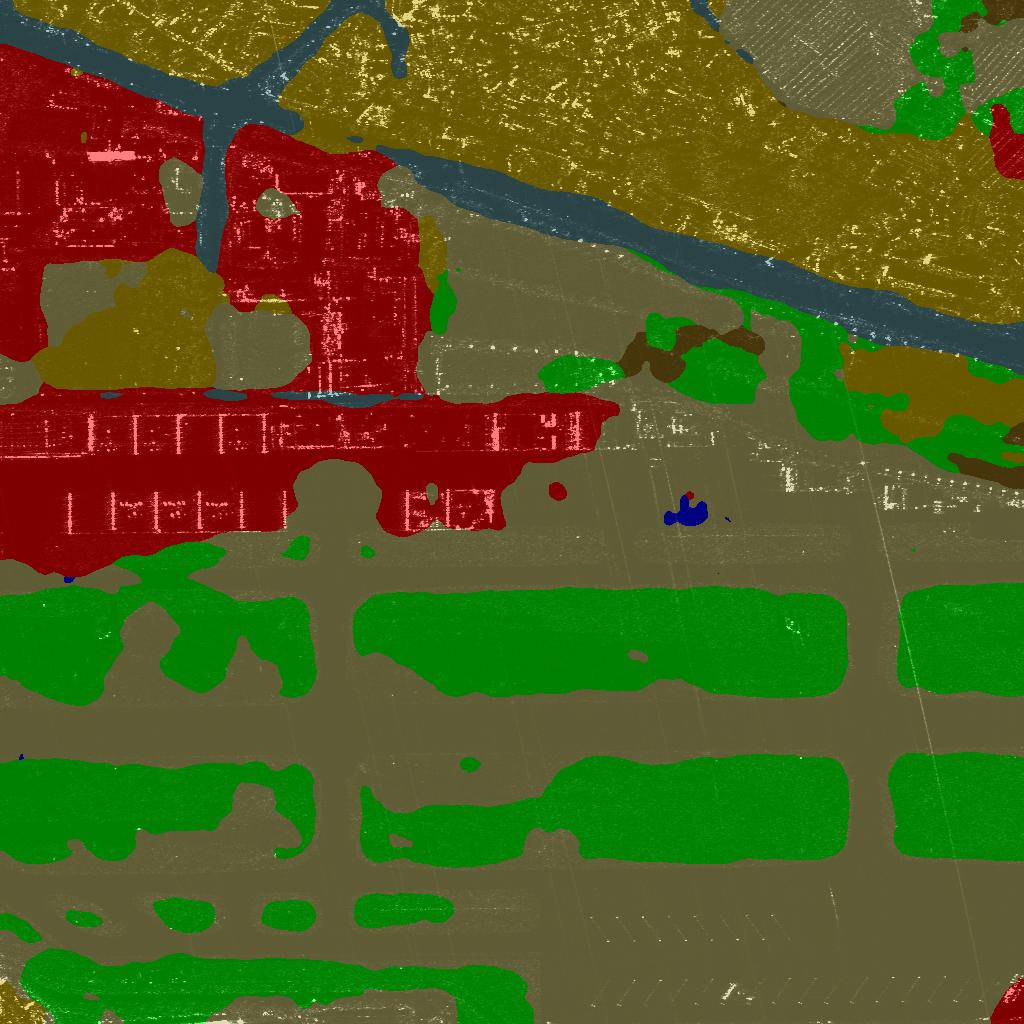}
              \includegraphics[width=.1\linewidth]{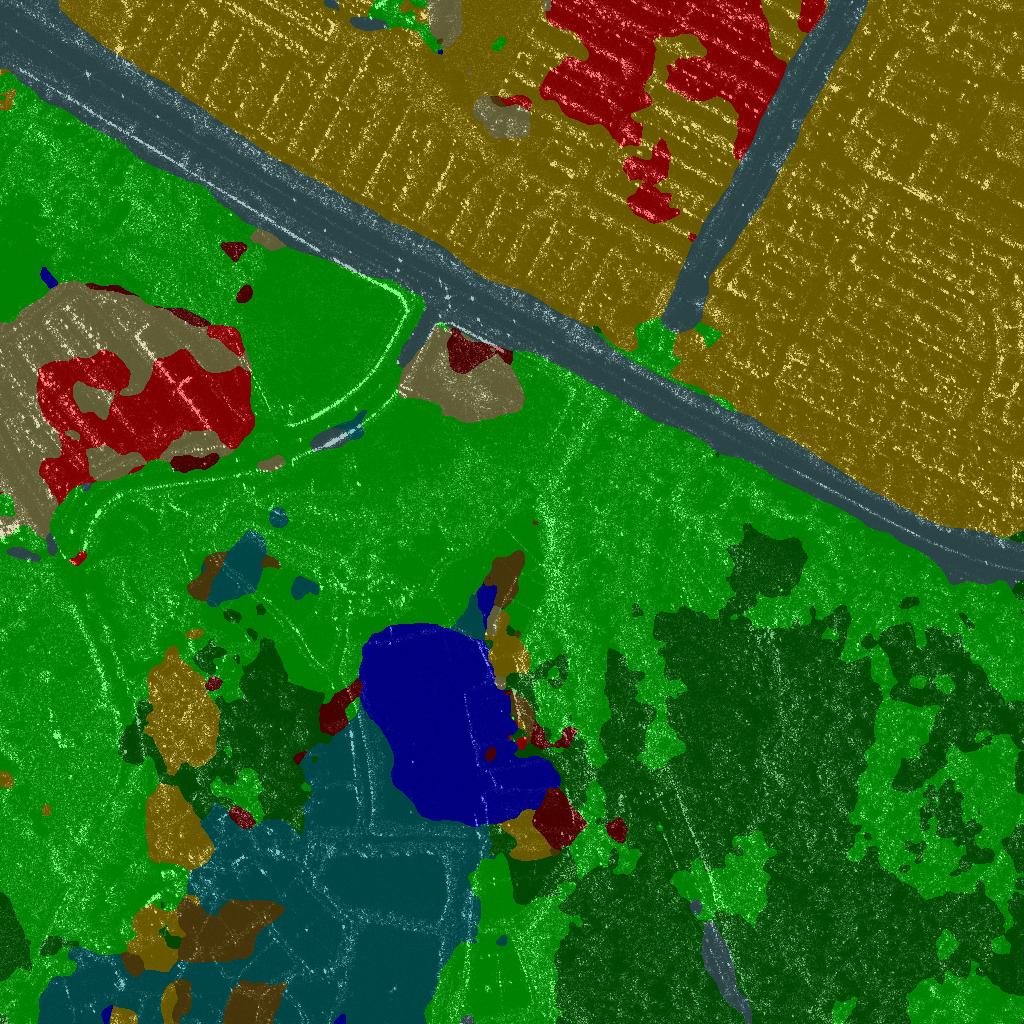}
              \includegraphics[width=.1\linewidth]{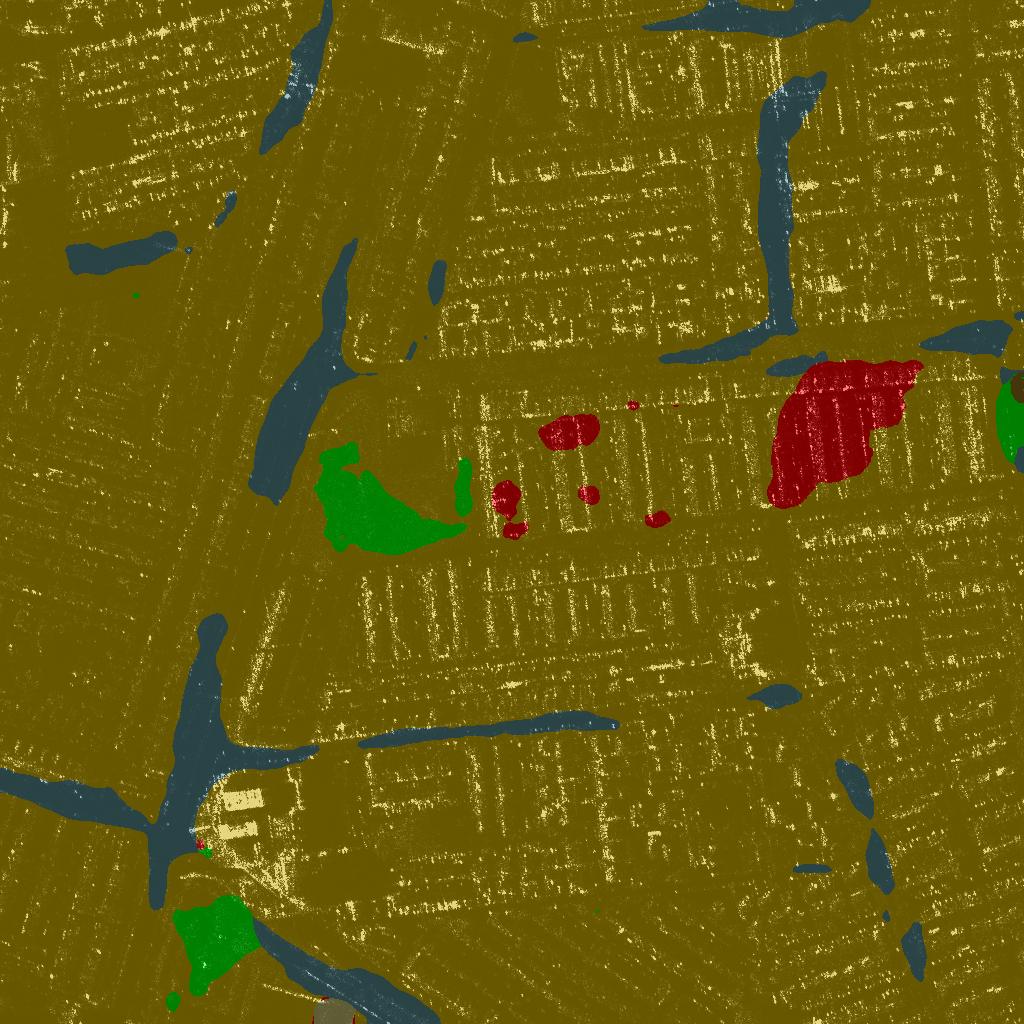}
              \includegraphics[width=.1\linewidth]{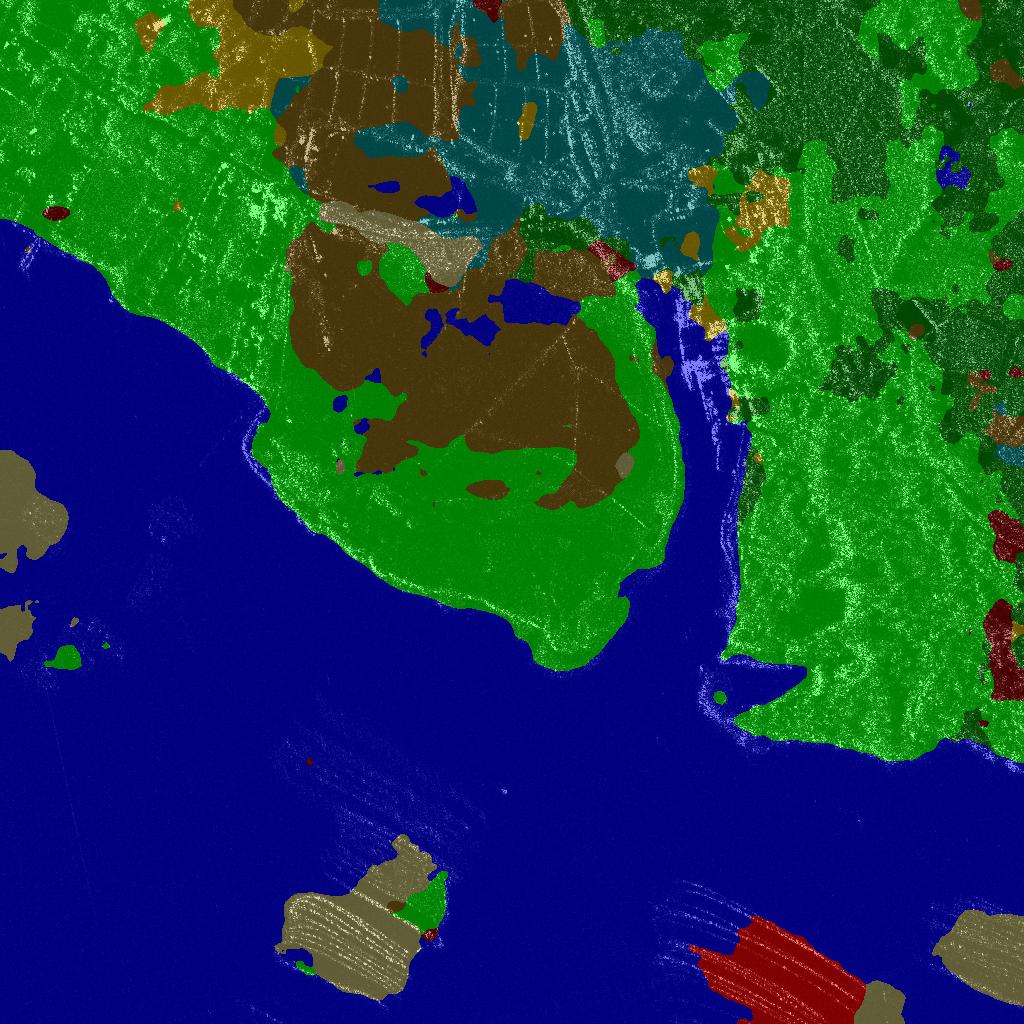}
              \includegraphics[width=.1\linewidth]{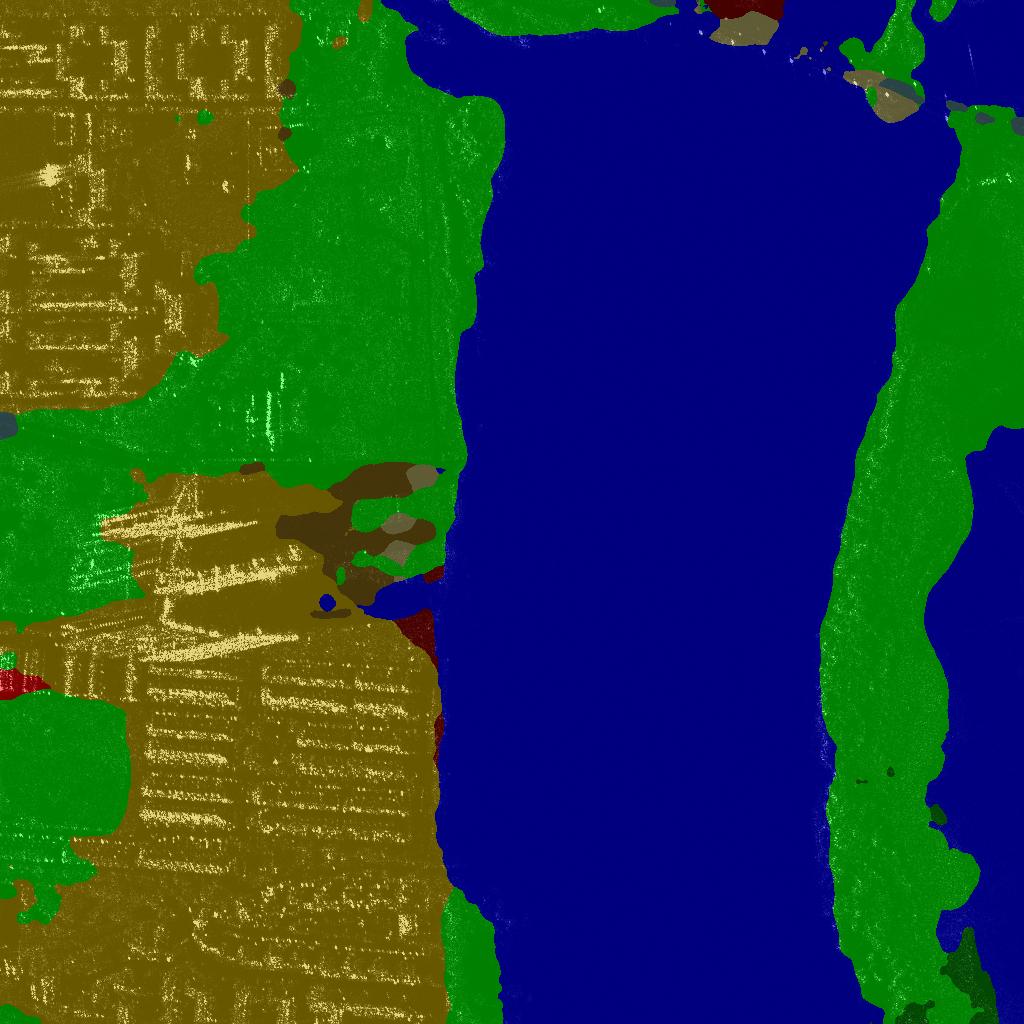}
	\end{minipage}\vspace{3pt}

    \begin{minipage}[t]{\linewidth}
	   \centering
                 \small\rotatebox{90}{\hspace{7pt}CWSAM}
    	     \includegraphics[width=.1\linewidth]{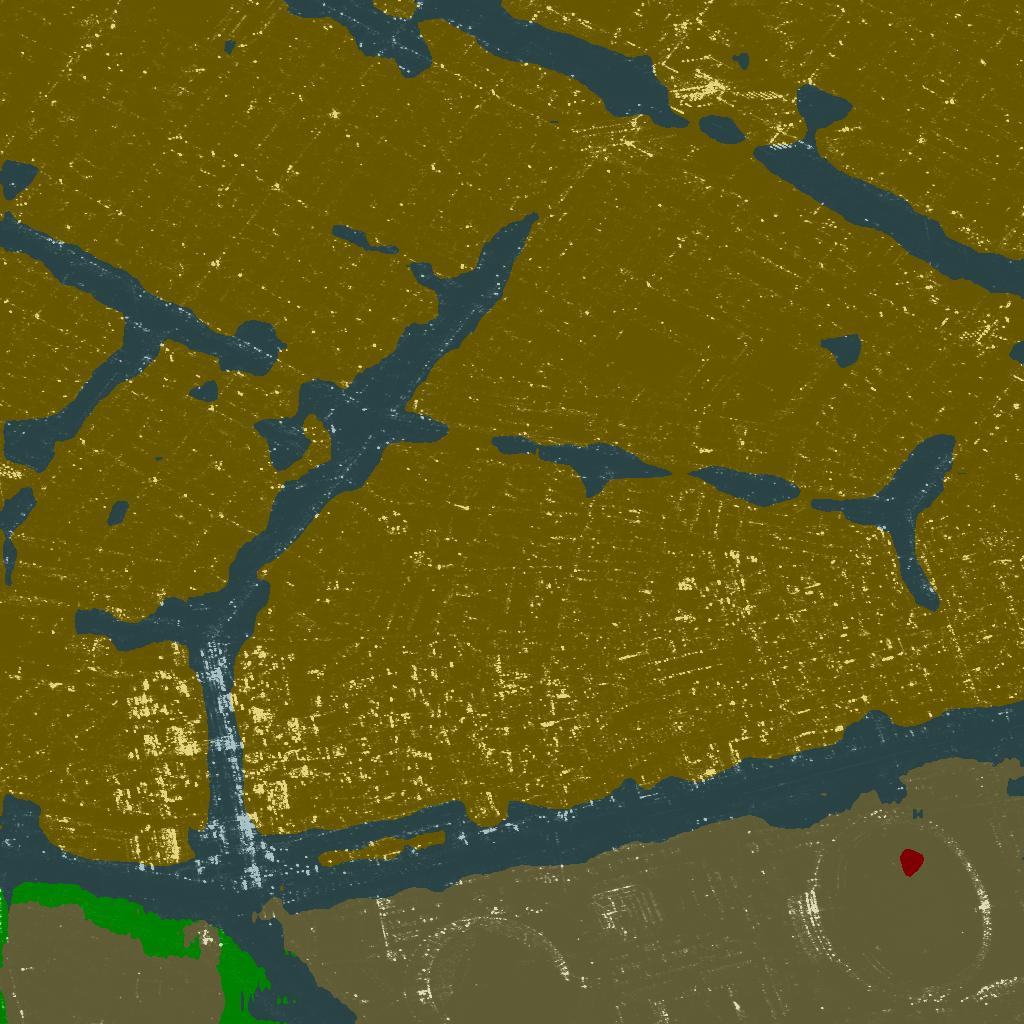}
	       \includegraphics[width=.1\linewidth]{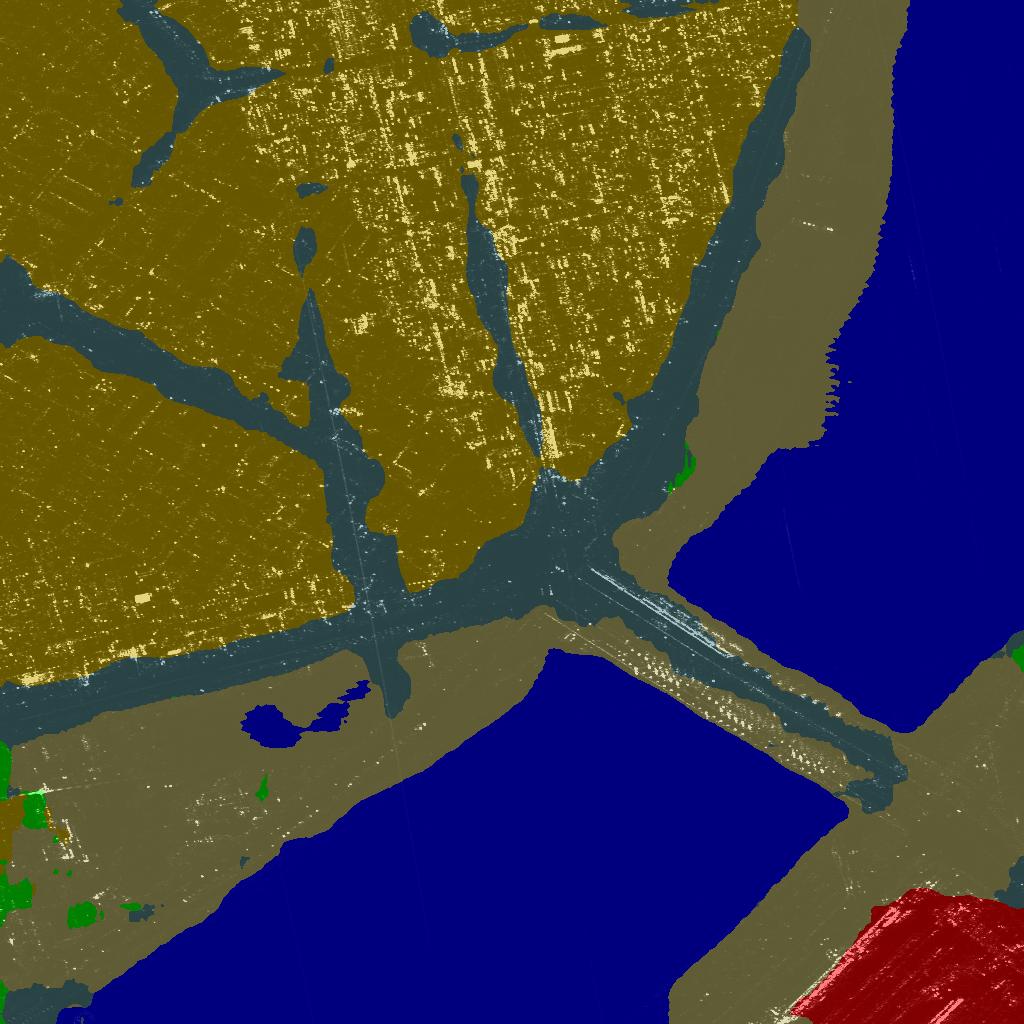}
              \includegraphics[width=.1\linewidth]{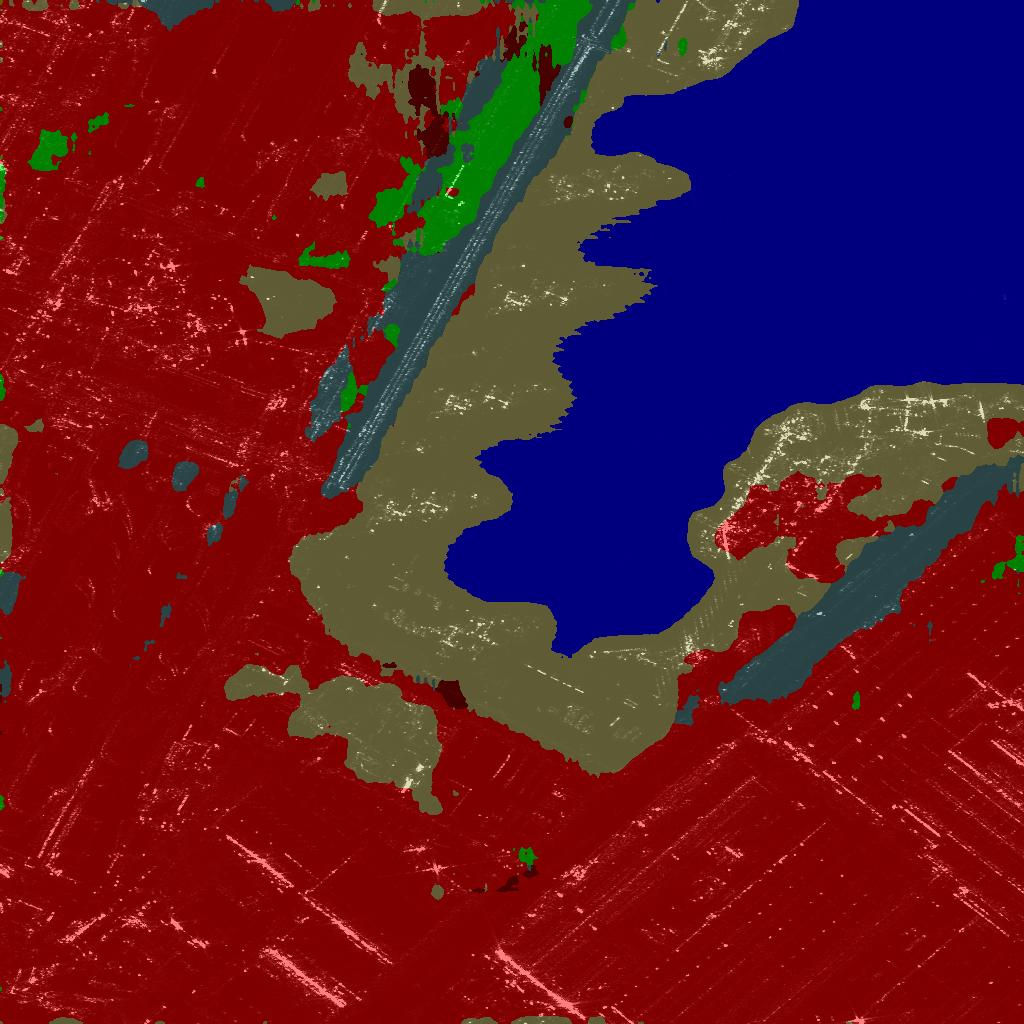}
              \includegraphics[width=.1\linewidth]{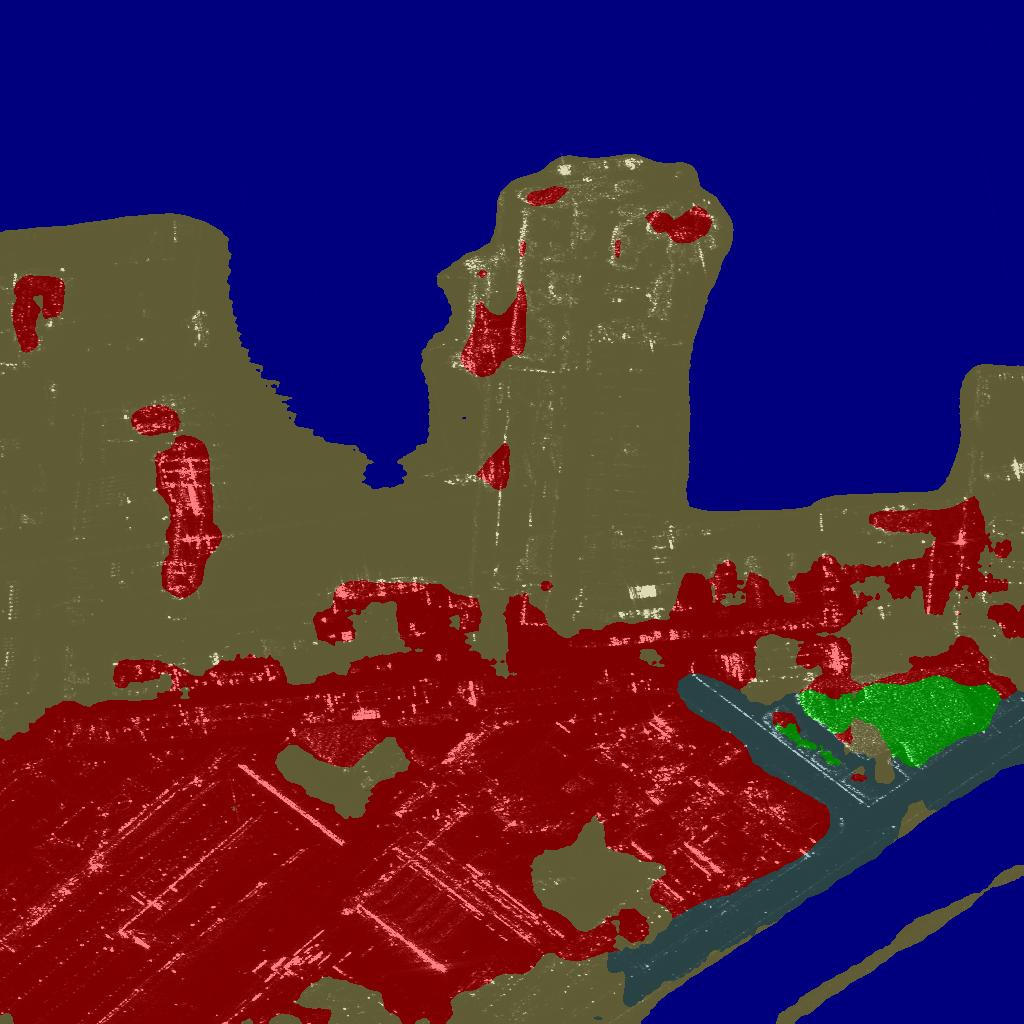}
              \includegraphics[width=.1\linewidth]{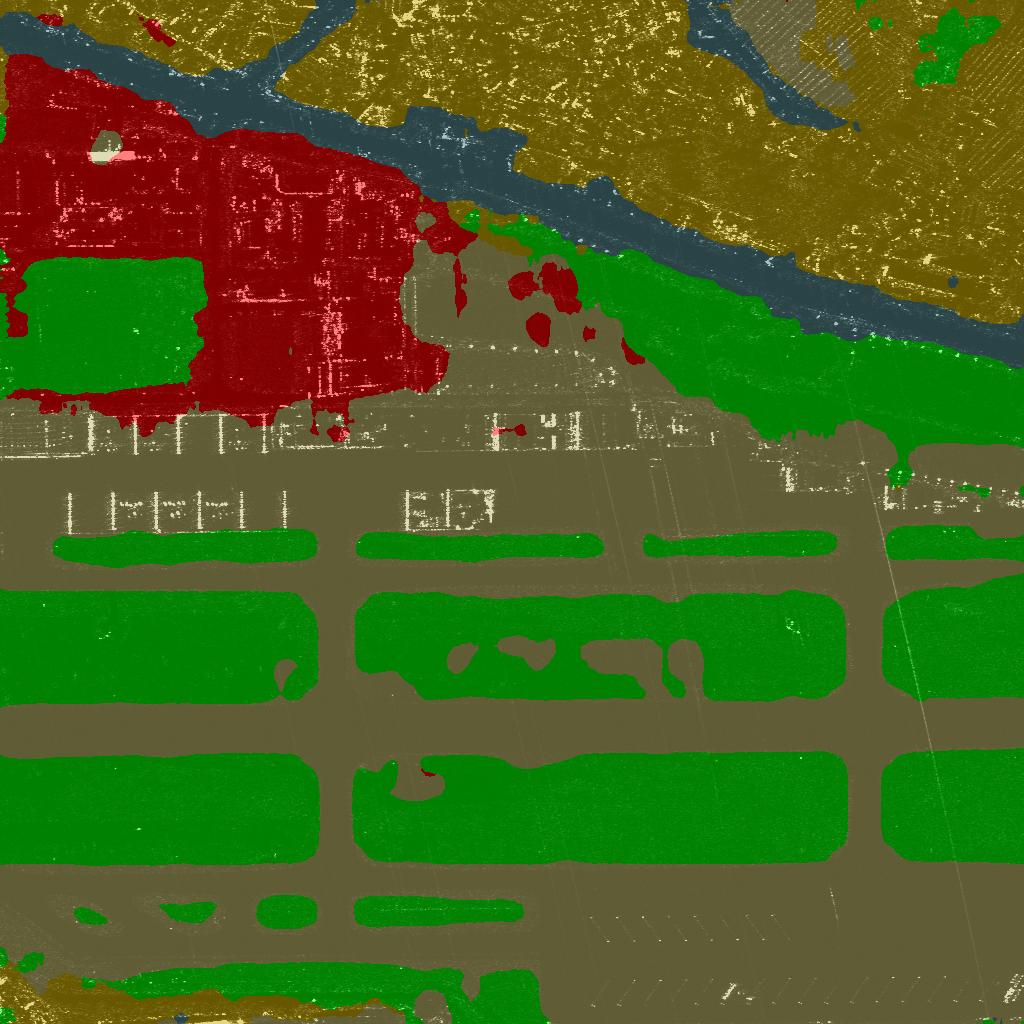}
              \includegraphics[width=.1\linewidth]{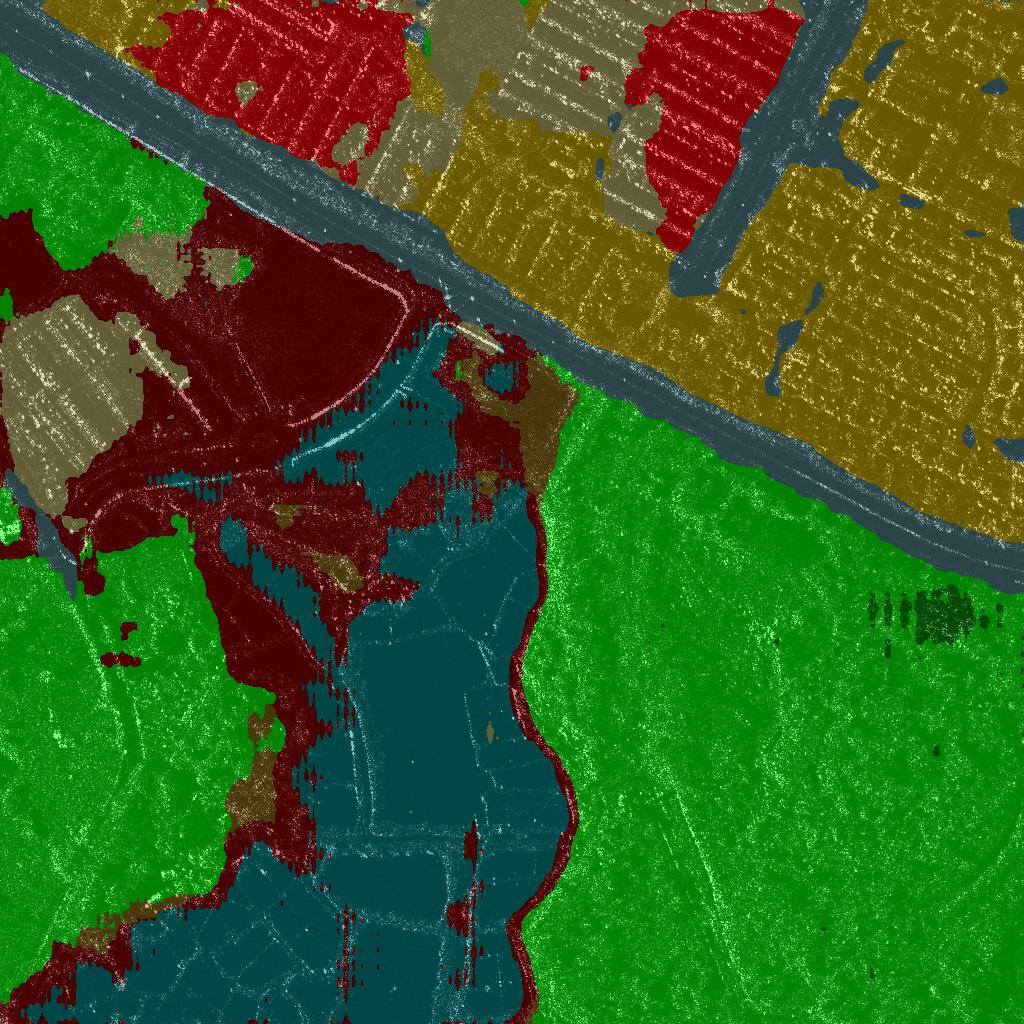}
              \includegraphics[width=.1\linewidth]{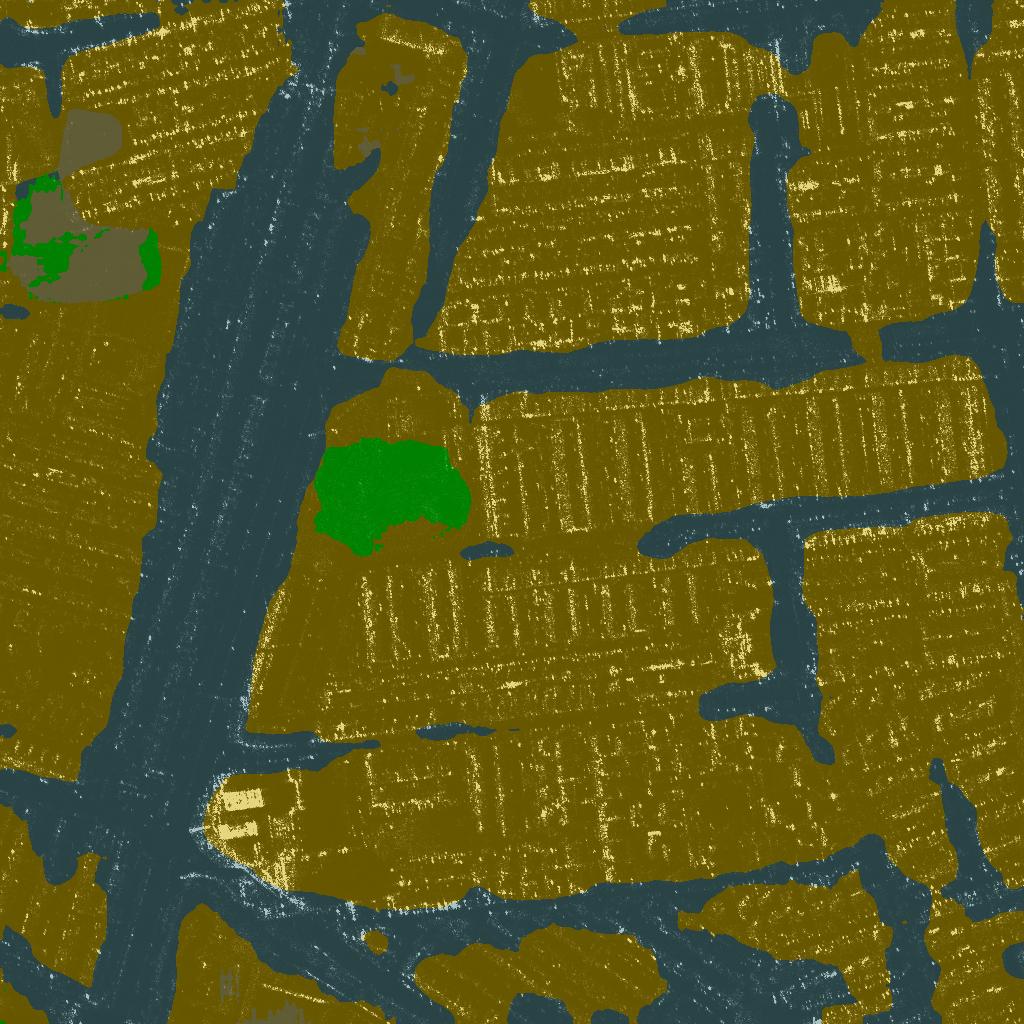}
              \includegraphics[width=.1\linewidth]{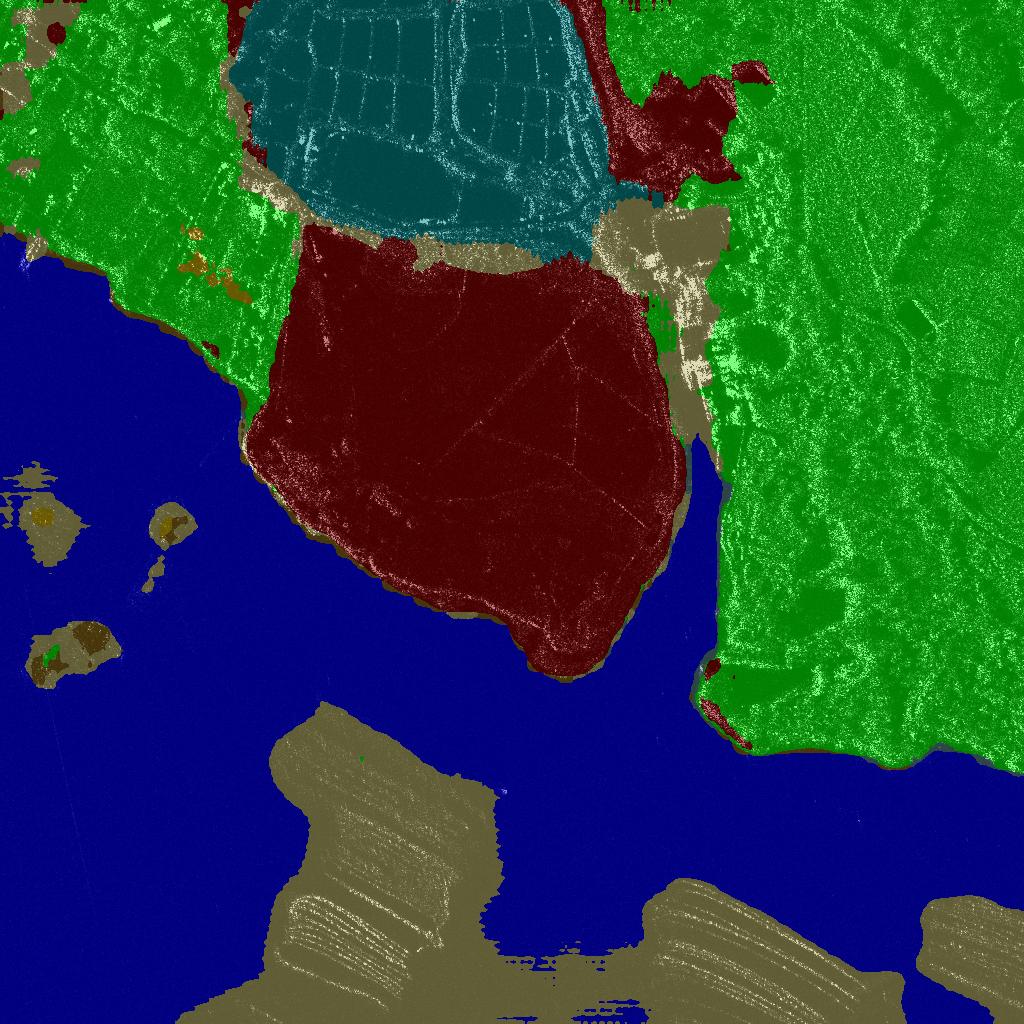}
              \includegraphics[width=.1\linewidth]{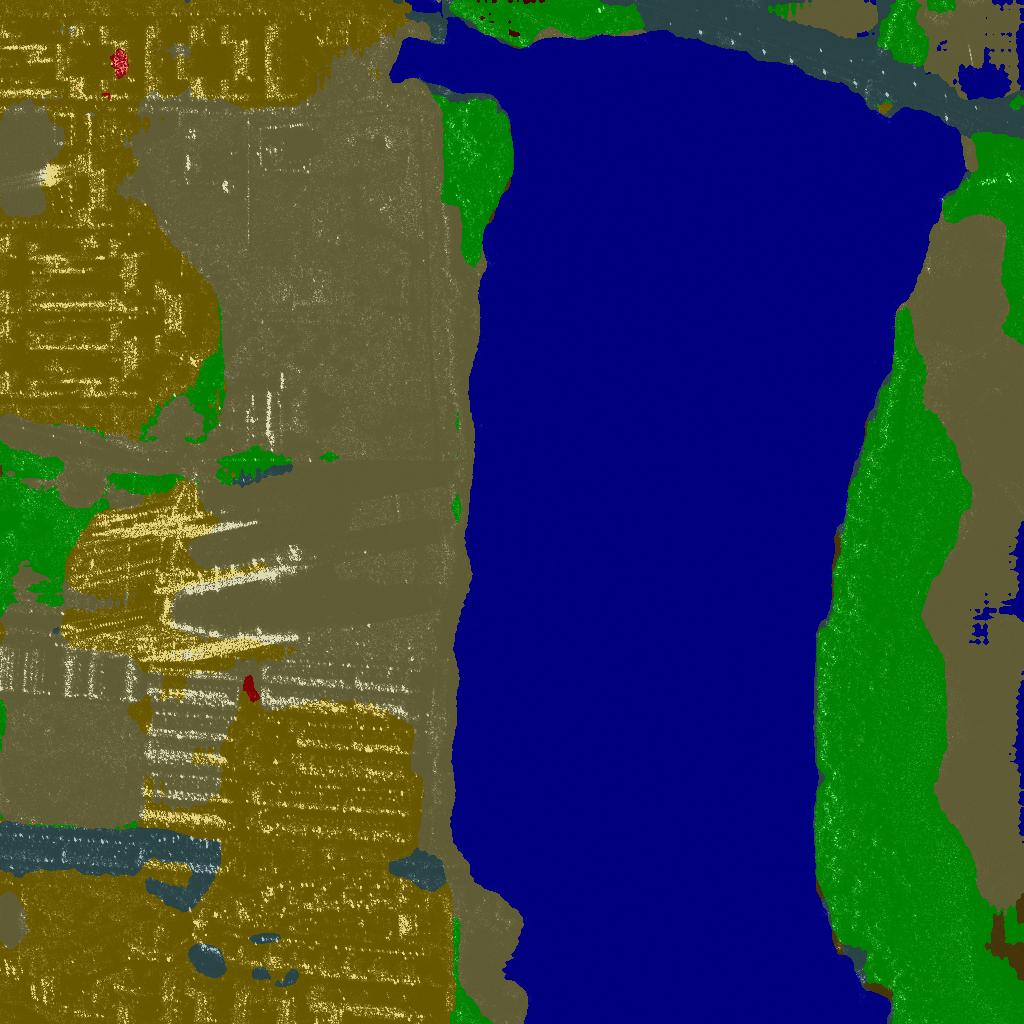}
	\end{minipage}\vspace{3pt}

    \begin{minipage}[t]{\linewidth}
	   \centering
        \small\rotatebox{90}{Ground Truth}
	       \includegraphics[width=.1\linewidth]{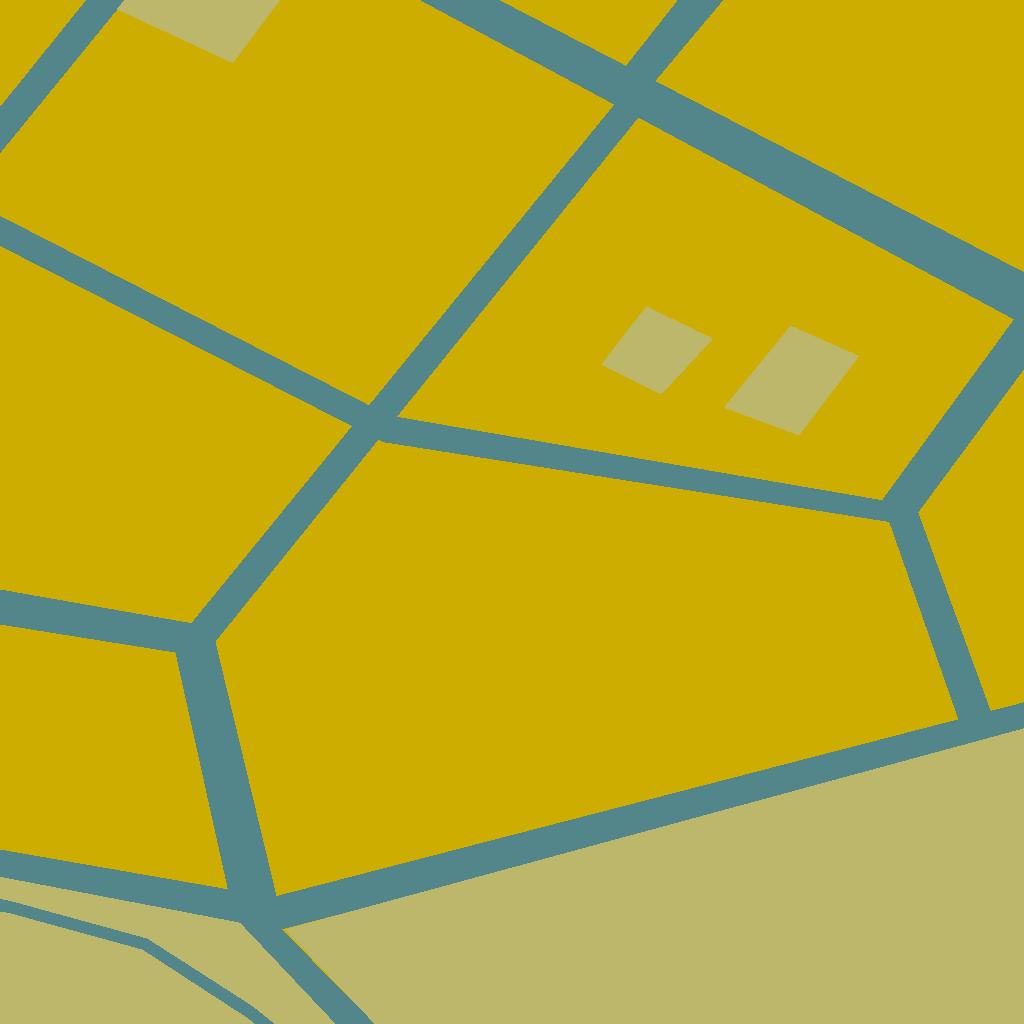}
	       \includegraphics[width=.1\linewidth]{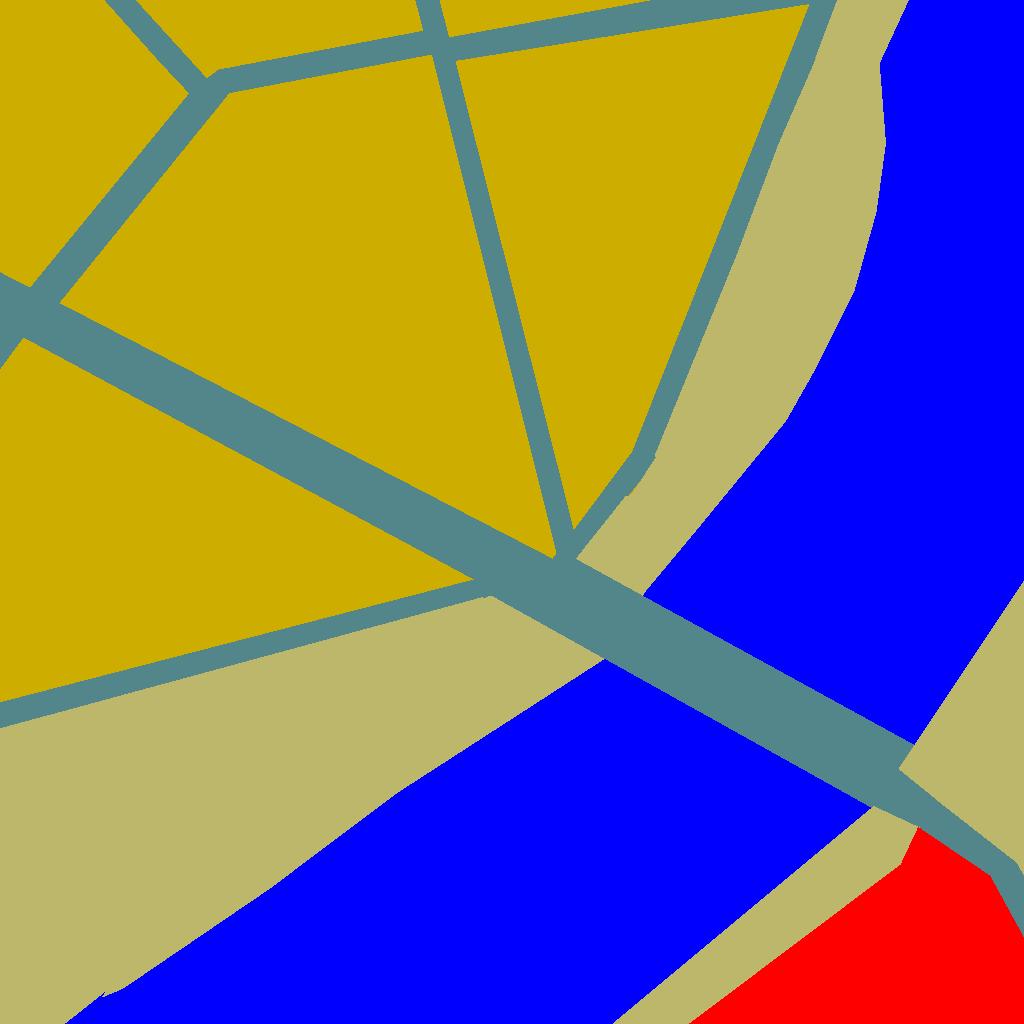}
              \includegraphics[width=.1\linewidth]{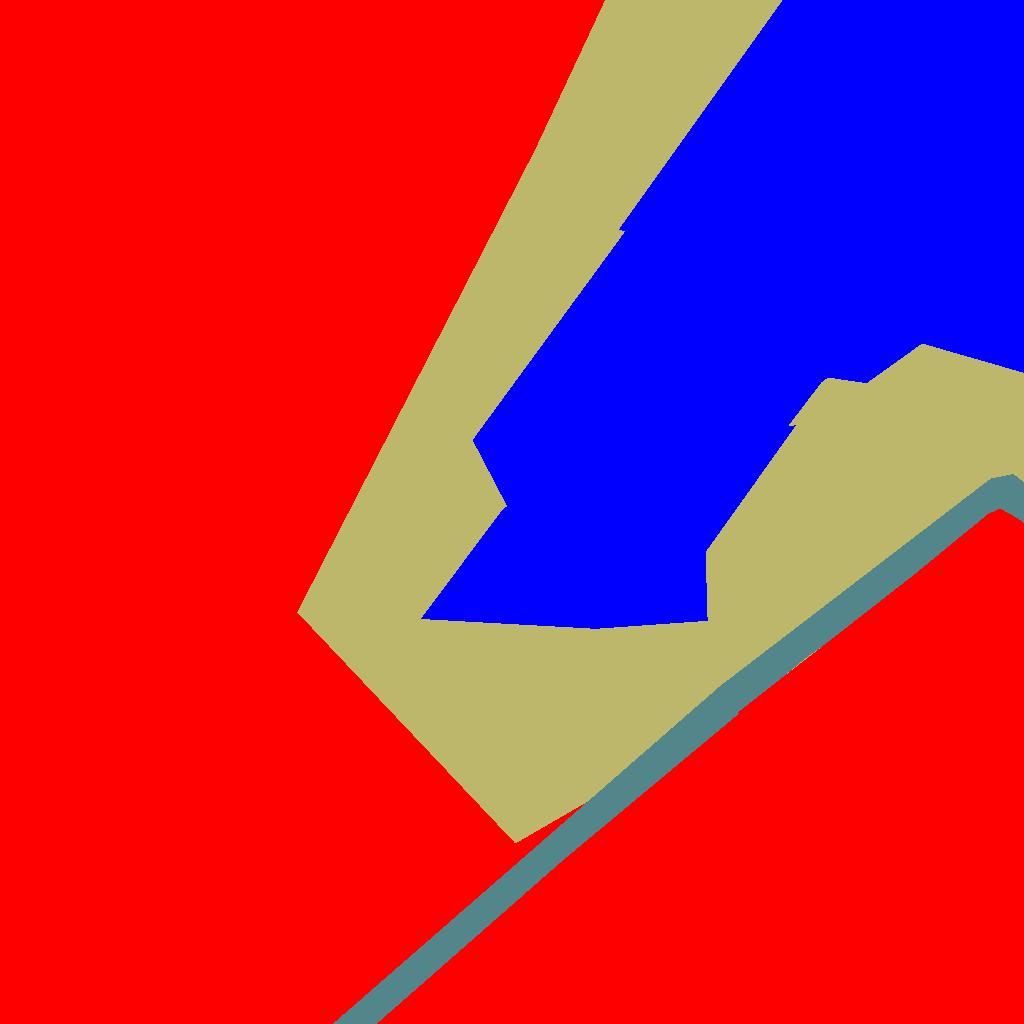}
              \includegraphics[width=.1\linewidth]{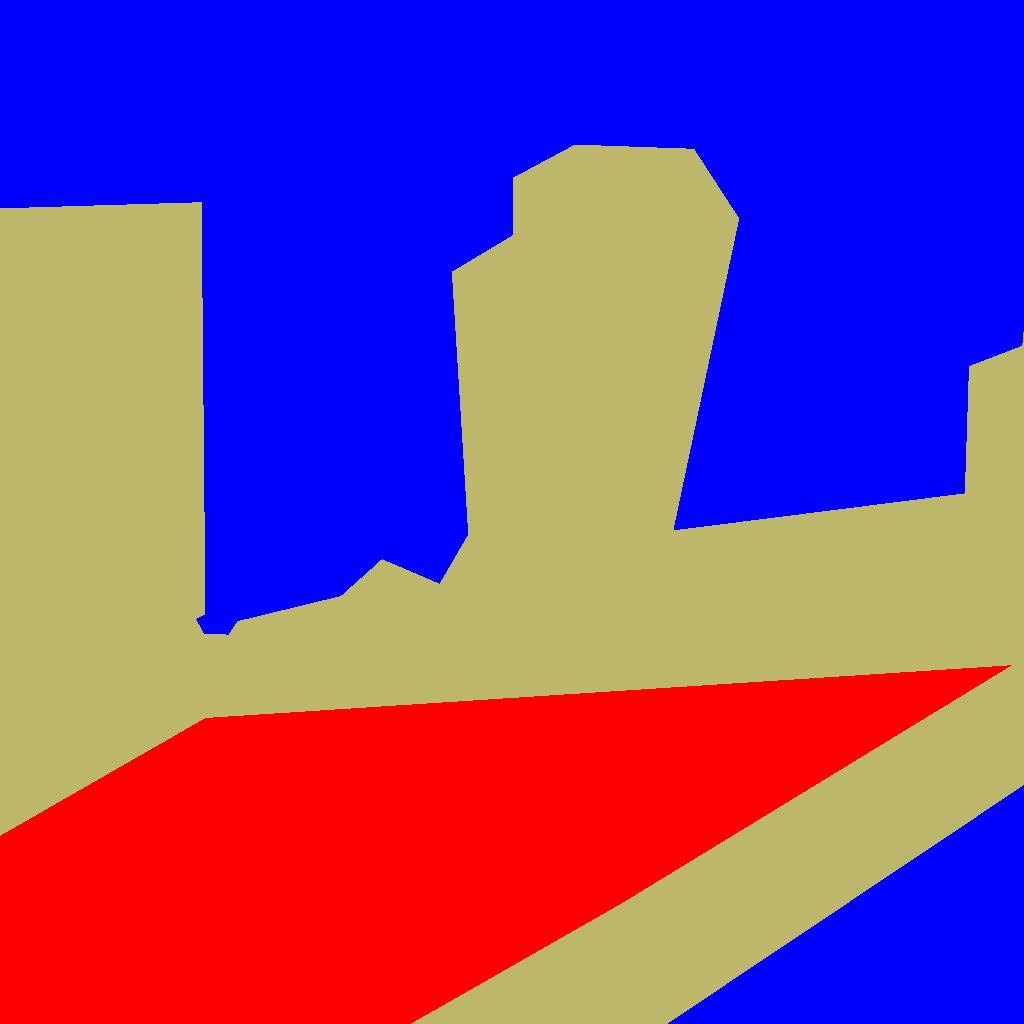}
              \includegraphics[width=.1\linewidth]{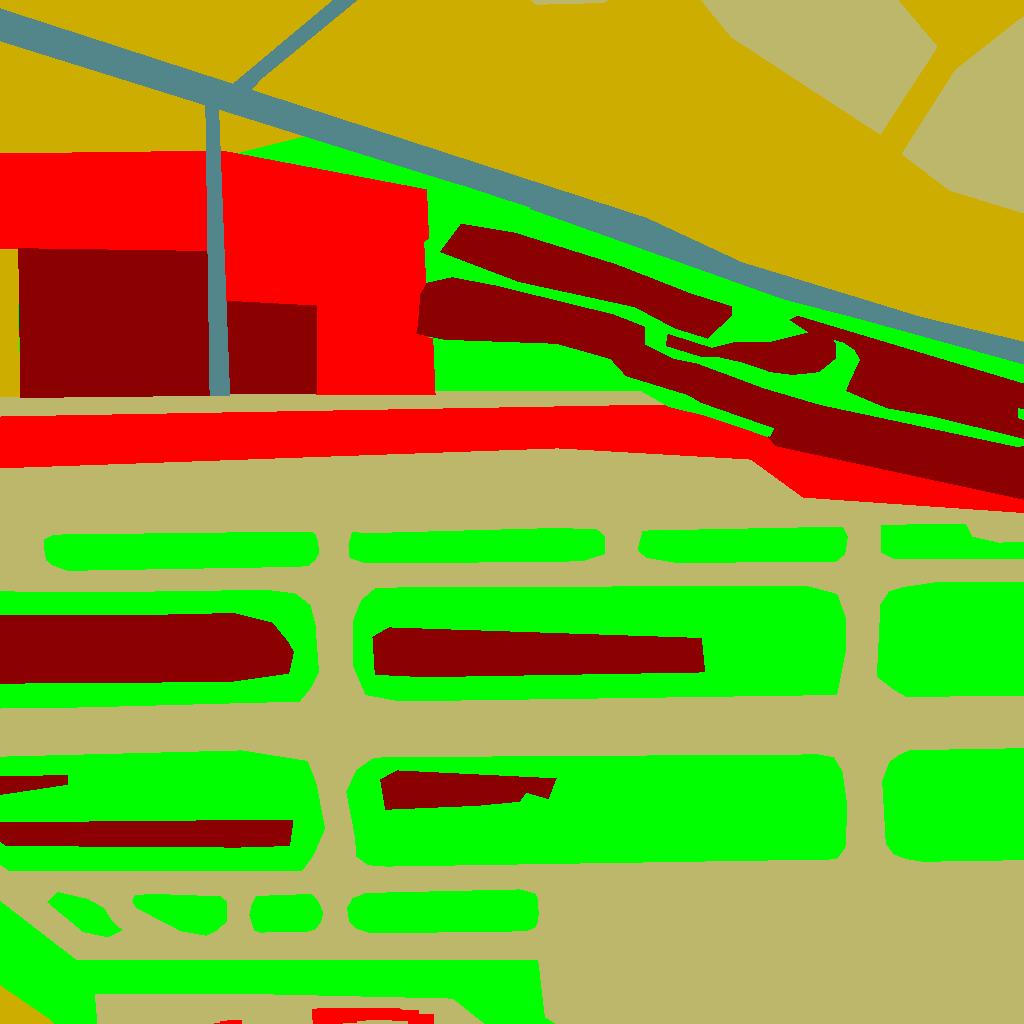}
              \includegraphics[width=.1\linewidth]{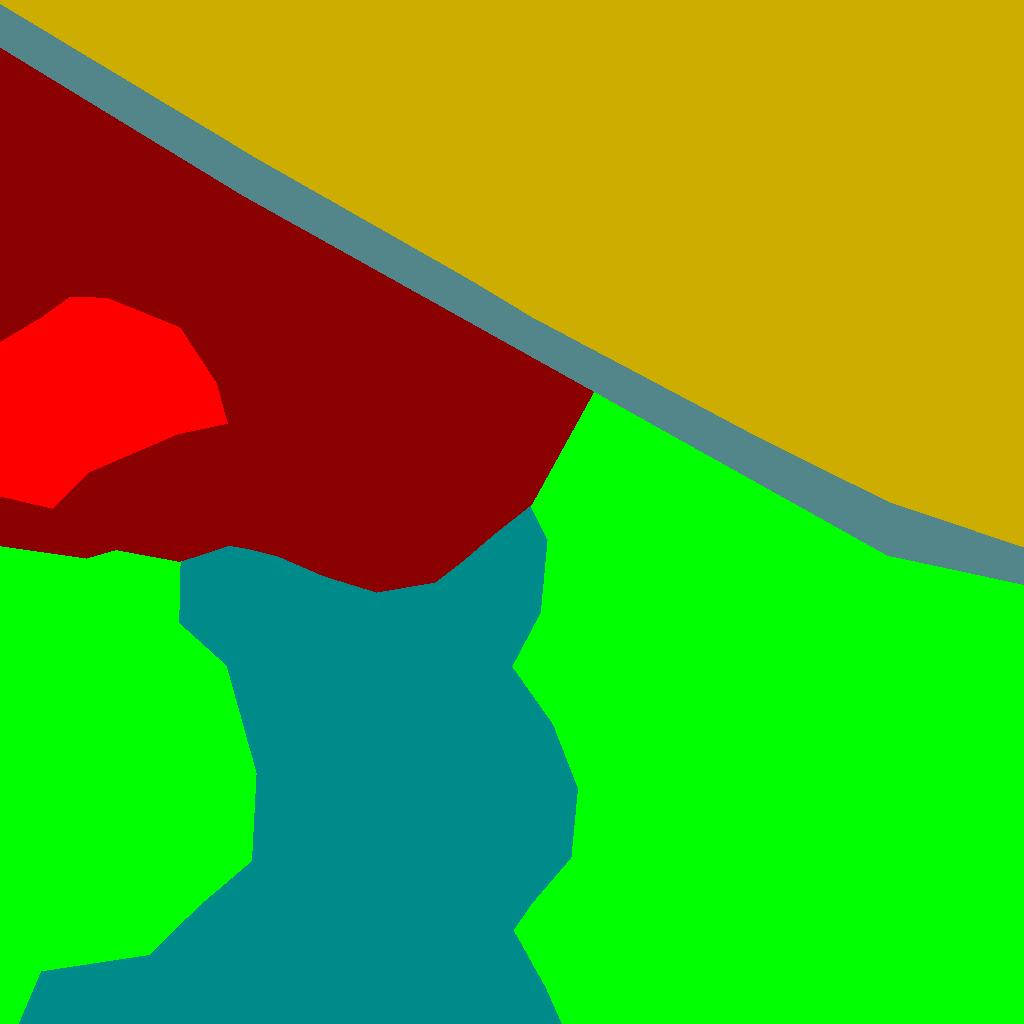}
              \includegraphics[width=.1\linewidth]{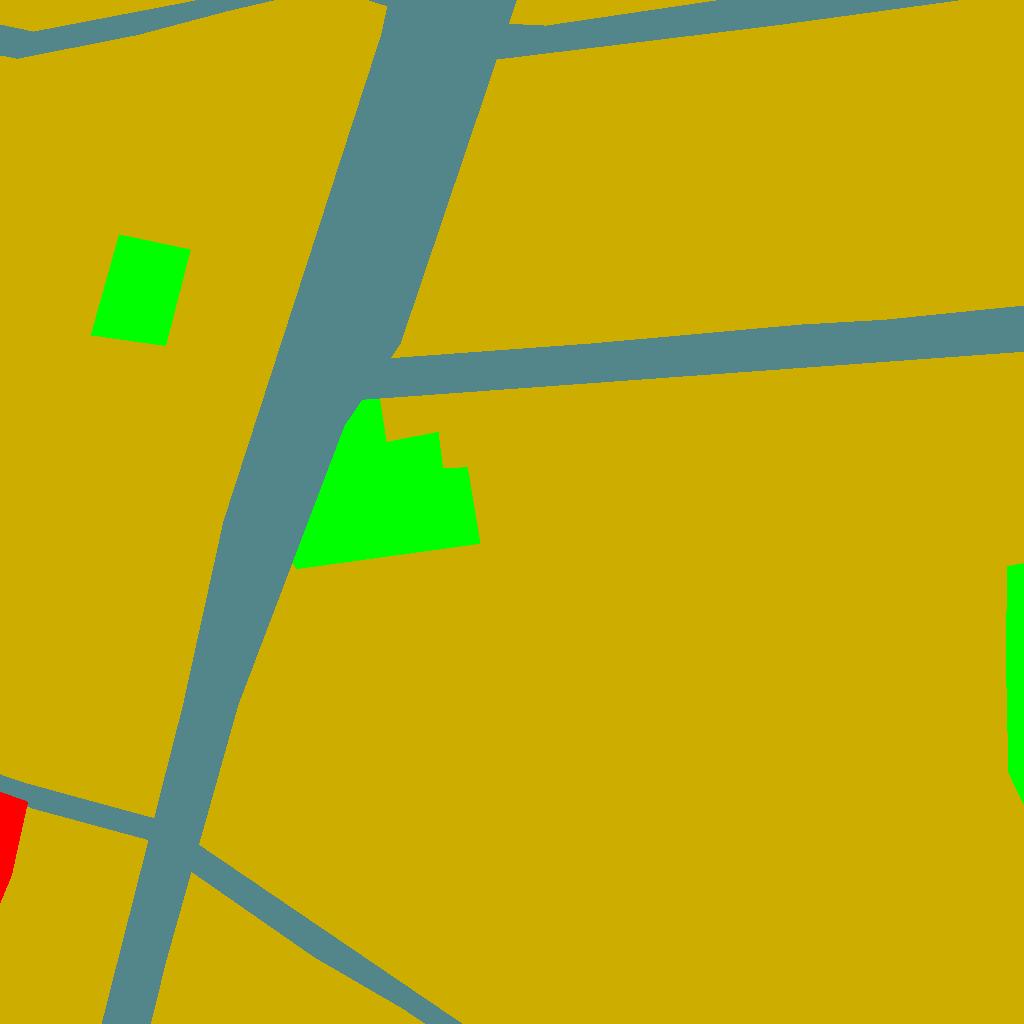}
              \includegraphics[width=.1\linewidth]{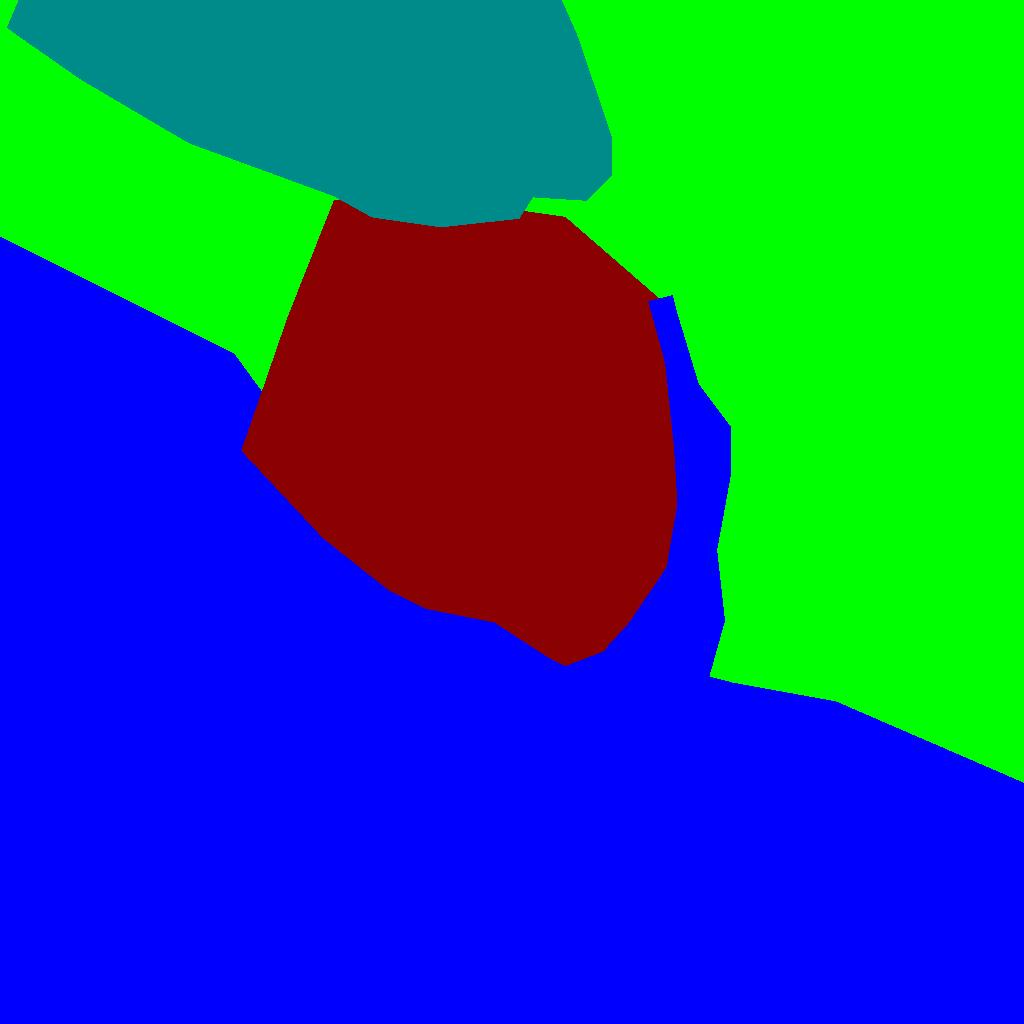}
              \includegraphics[width=.1\linewidth]{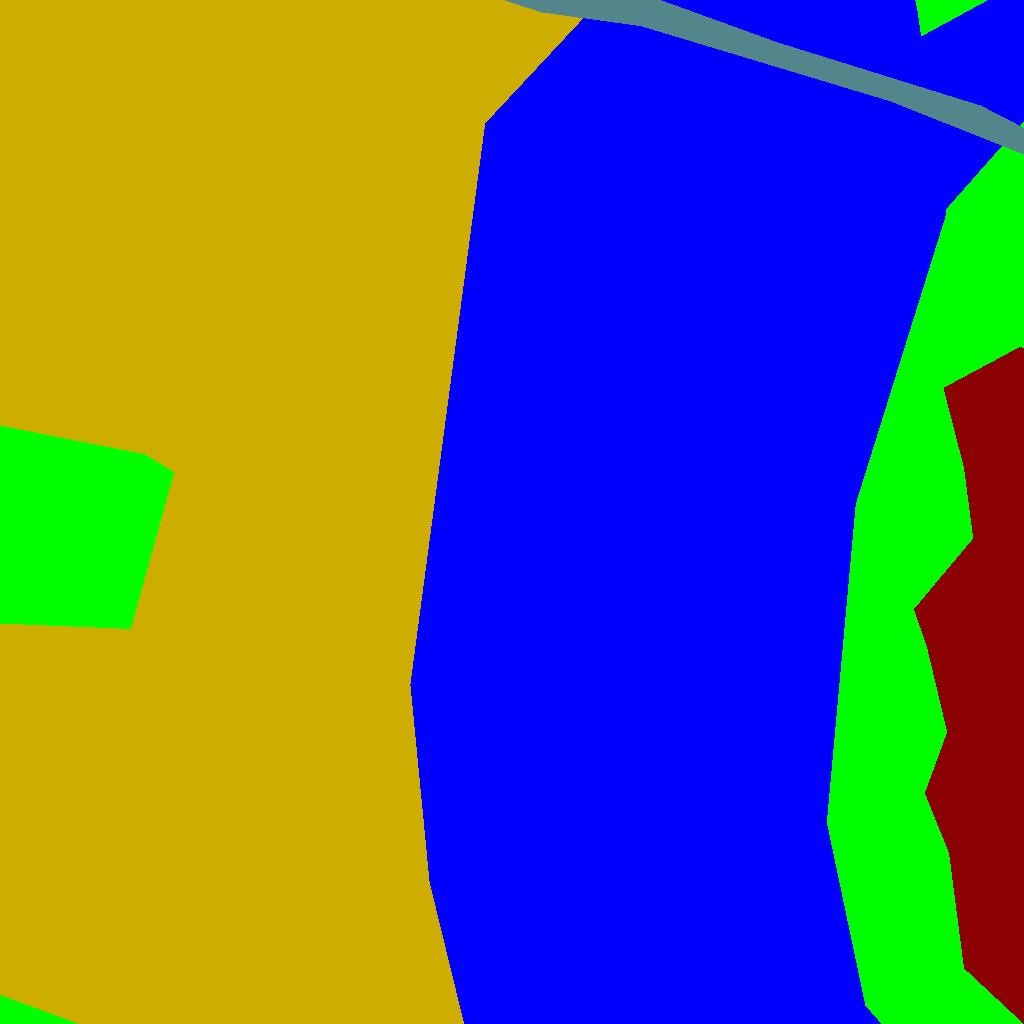}
	\end{minipage}\vspace{5pt}
  \begin{minipage}{0.07\textwidth}
        \centering
        \begin{tikzpicture}
            \fill[blue] (0,0) rectangle (0.5,0.5); 
            \node at (0.9,0.25) {water}; 
        \end{tikzpicture}
    \end{minipage}
    \begin{minipage}{0.105\textwidth}
        \centering
        \begin{tikzpicture}
            \fill[wood] (0,0) rectangle (0.5,0.5); 
            \node at (1.22,0.25) {woodland}; 
        \end{tikzpicture}
    \end{minipage}
    \begin{minipage}{0.105\textwidth}
        \centering
        \begin{tikzpicture}
            \fill[veget] (0,0) rectangle (0.5,0.5); 
            \node at (1.25,0.25) {vegetation}; 
        \end{tikzpicture}
    \end{minipage}
    \begin{minipage}{0.095\textwidth}
        \centering
        \begin{tikzpicture}
            \fill[bare] (0,0) rectangle (0.5,0.5); 
            \node at (1.15,0.25) {bare soil}; 
        \end{tikzpicture}
    \end{minipage}
    \begin{minipage}{0.095\textwidth}
        \centering
        \begin{tikzpicture}
            \fill[indus] (0,0) rectangle (0.5,0.5); 
            \node at (1.12,0.25) {industry}; 
        \end{tikzpicture}
    \end{minipage}
    \begin{minipage}{0.1\textwidth}
        \centering
        \begin{tikzpicture}
            \fill[resid] (0,0) rectangle (0.5,0.5); 
            \node at (1.2,0.25) {residence}; 
        \end{tikzpicture}
    \end{minipage}
    \begin{minipage}{0.065\textwidth}
        \centering
        \begin{tikzpicture}
            \fill[road] (0,0) rectangle (0.5,0.5); 
            \node at (0.85,0.25) {road}; 
        \end{tikzpicture}
    \end{minipage}
    \begin{minipage}{0.075\textwidth}
        \centering
        \begin{tikzpicture}
            \fill[paddy] (0,0) rectangle (0.5,0.5); 
            \node at (0.95,0.25) {paddy}; 
        \end{tikzpicture}
    \end{minipage}
    \begin{minipage}{0.09\textwidth}
        \centering
        \begin{tikzpicture}
            \fill[planting] (0,0) rectangle (0.5,0.5); 
            \node at (1.1,0.25) {planting}; 
        \end{tikzpicture}
    \end{minipage}
    \begin{minipage}{0.09\textwidth}
        \centering
        \begin{tikzpicture}
            \fill[human] (0,0) rectangle (0.5,0.5); 
            \node at (1.4,0.25) {human built}; 
        \end{tikzpicture}
    \end{minipage}

    \caption{Qualitative results of landcover classification masks. The compared visualizations between the proposed method and the best state-of-the-art algorithm HRNet-48 on FUSAR-Map2.0 dataset. The predicted masks in the second and third rows are overlapped with the original SAR image for visualization by using a transparency of 0.5.}
\end{figure*}
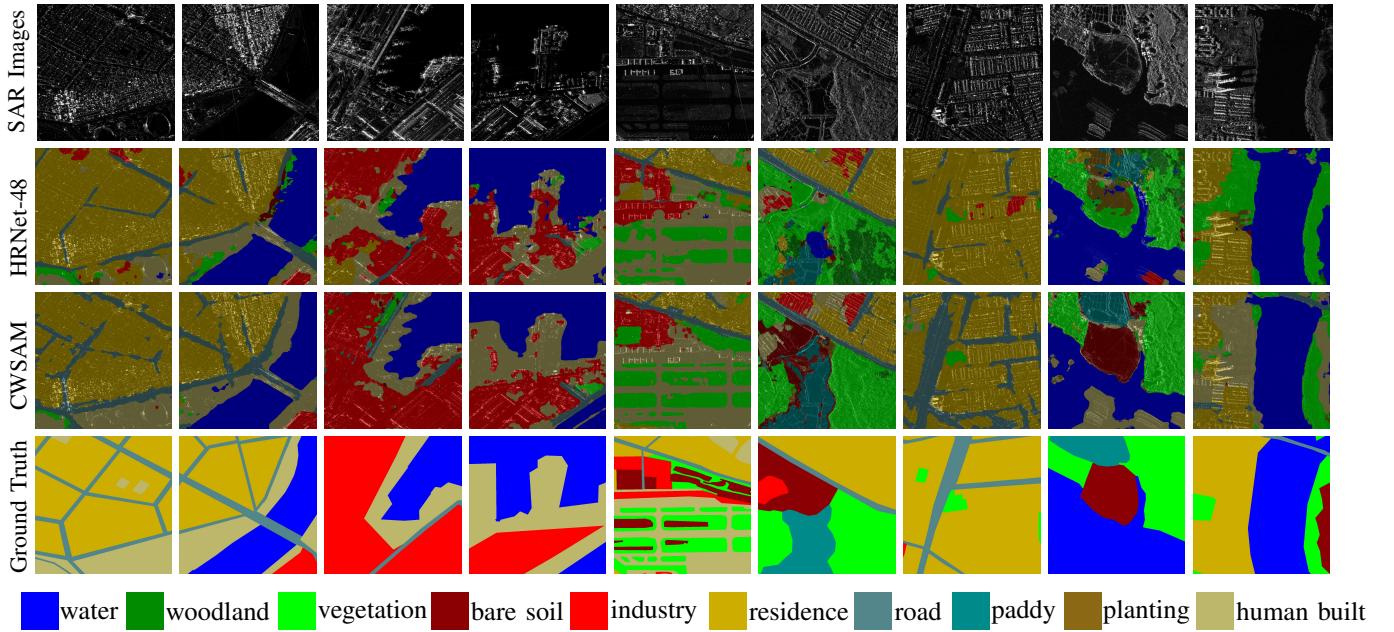

\begin{table}[htbp]
\centering
\caption{The comparison results of the proposed method and the state-of-the-art algorithms on FUSAR-Map1.0 dataset. Evaluation by multiple average standard metrics of semantic segmentation.}
\begin{tabular}{lcccccc}
\toprule
 & mIoU & OA & Accuracy & Precision & mDice \\
\midrule 
Deeplabv3 & 51.05 & 74.21 & 65.37 & 69.95 & 64.85 \\
FE-Deeplabv3 & 56.79 & 76.97 & 70.34 & 70.2 & 69.54 \\
Deeplabv3plus & 53.68 & 77.47 & 65.28 & 70.96 & 66.35 \\
HRNet-18 & 47.51 & 64.83 & 54.05 & 68.48 & 60.08 \\
HRNet-48 & 52.7 & 72.12 & 62.42 & 71.04 & 66.12 \\
Segformer-mitb0 & 57.81 & 80.25 & 69.66 & 72.1 & 69.65 \\
Segformer-mitb3 & 57.25 & 80.69 & 68.91 & 71.68 & 68.17 \\
CWSAM & \pmb{61.48} & \pmb{82.14} & \pmb{73.45} & \pmb{75.65} & \pmb{73.68} \\
\bottomrule
\end{tabular}
\end{table}

As for separate categories, we mainly consider the mIOU and accuracy metrics for evaluation. In Table II, most categories have the best performance among all these algorithms as illustrated  for mIoU. Especially in the challenging category of road, the proposed CWSAM has a significant advantage, outperforming the second highest algorithm by the mIOU of 4.83$\%$. Additionally, it has strong robustness on Accuracy across all categories. In Table III, even though the proposed algorithm doesn't achieve the best performance in any single category, its overall accuracy is the highest. This is because it doesn't have extremely poor performance in any particular category and keep a proper trade-off on accuracy between all categories.

\begin{table}[htbp]
\centering
\caption{The evaluation results of experiments on mIoU for every categories on FUSAR-Map1.0 dataset.}
\begin{tabular}{lccccc}
\toprule
 & mIoU & building & vegetation & water & road \\
\midrule
Deeplabv3 & 51.05 & 51.01 & 43.98 & 85.2 & 24.02 \\
FE-Deeplabv3 & 56.79 & 50.41 & 60.52 & 90.75 & 25.44 \\
Deeplabv3plus & 53.68 & 46.57 & 58.64 & 89.28 & 20.19 \\
HRNet-18 & 47.51 & 32.21 & 51.36 & 89.26 & 17.25 \\
HRNet-48 & 52.7 & 47.84 & 49.73 & 88.82 & 24.42 \\
Segformer-mitb0 & 57.81 & 54.02 & 64.53 & 92.11 & 20.57 \\
Segformer-mitb3 & 57.25 & 56.44 & \pmb{66.92} & 91.46 & 14.16 \\
CWSAM & \pmb{61.48} & \pmb{57.76} & 64.77 & \pmb{93.11} & \pmb{30.27} \\
\bottomrule
\end{tabular}
\end{table}

\begin{table}[htbp]
\centering
\caption{The evaluation results of experiments on Accuarcy for every categories on FUSAR-Map1.0 dataset.}
\begin{tabular}{lccccc}
\toprule
 & Accuarcy & building & vegetation & water & road \\
\midrule
Deeplabv3 & 65.37 & 86.28 & 47.68 & \pmb{96.4} & 31.11 \\
FE-Deeplabv3 & 70.34 & 74.63 & 67.26 & 91.65 & \pmb{47.82} \\
Deeplabv3plus & 65.28 & 57.1 & \pmb{83.52} & 95.95 & 24.55 \\
HRNet-18 & 54.05 & 39.1 & 60.39 & 89.95 & 26.75 \\
HRNet-48 & 62.42 & 62.05 & 60.53 & 94.48 & 32.63 \\
Segformer-mitb0 & 69.66 & 80.84 & 77.28 & 94.49 & 26.04 \\
Segformer-mitb3 & 68.91 & \pmb{86.95} & 76.63 & 94.92 & 17.14 \\
CWSAM & \pmb{73.45} & 84.99 & 74.52 & 95.28 & 39.01 \\
\bottomrule
\end{tabular}
\end{table}
As the Figure 6 presents, We visualizes the segmentation results of both CWSAM and the runner-up algorithm on FUSAR-Map1.0 dataset, SegFormer. In the landcover classification task, the proposed method consistently delivers more precise pixel-level predictions. Notably, its masks exhibit superior edge continuity, and it distinctly outperforms in delineating challenging categories, such as road and building.

(2) Results on FUSAR-Map2.0: The landcover classification task on FUSAR-Map2.0 becomes more challenging with more landcover categories and more extreme data distribution such as woodland, bare soil, road and paddy field. The comprehensive comparison results of several standard metrics are illustrated in Table IV. Similar with FUSAR-Map1.0, the general performance of the proposed method surpasses other compared state-of-the-art algorithms. Our experiments confirms that it is hard to achieve the best performance on all categories of FUSAR-Map2.0 dataset while the proposed method has overall the most outstanding performance. CWSAM outperforms the second highest algorithm by the mIOU of 0.59$\%$, the OA (Overall Accuracy) of 1.81$\%$, the Precision of 2.19$\%$, except mDice is slightly lower than HRNet-48 by 0.24$\%$

\begin{table}[htbp]
\centering
\caption{The evaluation results of experiments on mIoU metric for every categories on FUSAR-Map2.0 dataset.}
\begin{tabular}{lccccc}
\toprule
 & mIoU & OA & Accuracy & Precision & mDice \\
\midrule
Deeplabv3 & 31.43 & 64.56 & 42.04 & 48.3 & 43.2 \\
FE-Deeplabv3 & 30.23 & 62.67 & 42.49 & 48.37 & 41.32 \\
Deeplabv3plus & 32.18 & 64.93 & 44.0 & 49.71 & 43.8 \\
HRNet-18 & 32.91 & 65.48 & 44.55 & 50.2 & 44.77 \\
HRNet-48 & 35.44 & 65.86 & 47.72 & 52.31 & \pmb{48.43} \\
Segformer-mitb0 & 32.21 & 63.17 & 44.27 & 46.28 & 43.73 \\
Segformer-mitb3 & 30.92 & 59.87 & 43.19 & 47.1 & 42.42 \\
CWSAM & \pmb{36.03} & \pmb{67.67} & \pmb{47.92} & \pmb{54.5} & 48.19 \\
\bottomrule
\end{tabular}
\end{table}

As listed in Table V and Table VI, the segment performance of FUSAR-Map2.0 varies between all categories. Several individual category is difficult to be recognized, such as bare soil and planting area. The overall performance of the proposed method is optimal based on evaluation metrics because it obtains the best performance in the majority of categories without extreme deficiencies in any specific categories. Specifically, CWSAM's evaluation results of mIOU and Accuracy metrics are higher on most categories while other methods perform well in only one or two categories. 

\begin{table*}[htbp]
\centering
\caption{The evaluation results on mIoU of comparison experiments between the proposed method and other state-of-the-art algorithms on FUSAR-Map2.0 dataset}
\begin{tabular}{lccccccccccc}
\toprule
 & mIoU & water & woodland & vegetation & bare soil & industry & residence & road & paddy & planting & human built \\
\midrule
Deeplabv3 & 31.43 & 87.39 & 17.8 & 48.47 & 1.01 & 31.88 & 52.87 & 30.19 & 10.39 & 16.25 & \pmb{18.02} \\
FE-Deeplabv3 & 30.23 & 88.8 & 28.37 & 48.22 & 1.36 & 24.36 & 52.7 & 26.39 & 8.99 & 6.82 & 16.24 \\
Deeplabv3plus & 32.18 & 90.27 & 24.59 & 50.94 & 2.8 & 31.98 & 54.64 & 27.89 & 14.42 & 7.37 & 16.94 \\
HRNet-18 & 32.91 & 88.24 & 28.07 & 52.46 & 1.97 & 26.97 & \pmb{55.18} & \pmb{31.88} & 6.23 & 22.75 & 15.38 \\
HRNet-48 & 35.44 & 88.73 & 25.66 & 52.48 & 2.57 & 32.17 & 54.37 & 30.77 & \pmb{20.83} & \pmb{30.06} & 16.75 \\
Segformer-mitb0 & 32.21 & 89.41 & 32.16 & 54.9 & 1.01 & 30.49 & 50.39 & 22.66 & 21.48 & 5.97 & 13.64 \\
Segformer-mitb3 & 30.92 & 89.11 & 34.9 & 53.48 & 2.29 & 22.01 & 45.62 & 25.11 & 14.64 & 9.42 & 12.63 \\
CWSAM & \pmb{36.03} & \pmb{90.45} & \pmb{43.45} & \pmb{57.32} & \pmb{10.0} & \pmb{40.9} & 54.85 & 28.35 & 13.99 & 6.58 & 14.74 \\
\bottomrule
\end{tabular}
\end{table*}

\begin{table*}[htbp]
\centering
\caption{The evaluation results on Accuarcy of comparison experiments between the proposed method and other state-of-the-art algorithms on FUSAR-Map2.0 dataset}
\begin{tabular}{lccccccccccc}
\toprule
 & Accuarcy & water & woodland & vegetation & bare soil & industry & residence & road & paddy & planting & human built \\
\midrule
Deeplabv3 & 42.04 & 91.04 & 31.04 & 67.48 & 1.15 & 36.66 & \pmb{86.16} & 43.51 & 12.32 & 23.52 & 27.52 \\
FE-Deeplabv3 & 42.49 & 91.96 & 61.95 & 69.33 & 1.43 & 27.86 & 74.45 & 42.66 & 9.8 & 8.72 & 37.17 \\
Deeplabv3plus & 44.0 & \pmb{94.48} & 51.98 & 71.97 & 3.16 & 38.05 & 74.7 & 36.93 & 20.75 & 8.11 & 39.89 \\
HRNet-18 & 44.55 & 93.58 & 50.95 & 72.83 & 2.54 & 30.56 & 78.55 & 45.37 & 6.82 & 31.16 & 33.16 \\
HRNet-48 & 47.72 & 93.99 & 48.07 & 71.51 & 3.43 & 37.6 & 74.54 & 45.82 & 2.52 & \pmb{39.37} & 37.61 \\
Segformer-mitb0 & 44.27 & 92.29 & 52.0 & 74.13 & 1.63 & 36.19 & 65.49 & 44.44 & \pmb{29.4} & 7.55 & 39.6 \\
Segformer-mitb3 & 43.19 & 92.13 & 52.43 & 69.44 & 4.0 & 25.88 & 57.78 & 48.33 & 18.54 & 11.96 & \pmb{51.45} \\
CWSAM & \pmb{47.92} & 92.18 & \pmb{64.75} & \pmb{79.77} & \pmb{14.3} & \pmb{48.14} & 68.13 & \pmb{52.52} & 15.58 & 7.46 & 36.39 \\
\bottomrule
\end{tabular}
\end{table*}

Therefore, the evaluation results on both datasets demonstrates that our method is efficient on multiple landcover classification datasets of SAR images and has higher model robustness and generalization.

\subsection{Ablation Study}
In this section, comprehensive experiments of ablation study are conducted on two datasets FUSAR-Map1.0 adn FUSAR-Map2.0. The ablation study disentangles the effect of every individual components and investigate the necessity of all designed modules in the proposed method. All the parameter settings keep constant in the ablation study and previous compared experiments with state-of-the-art methods. 
The proposed method has mainly three independent modules: classwise mask encoder with feature enhancement module, ViT image encoder adapters and task specific input module of SAR image low frequency information, which can be employed in the experiments sequentially and integrally. According to the evaluation results of ablation study shown in Table VII on FUSAR-Map1.0 and Table VIII on FUSAR-Map2.0, our analyses are as follows:

\begin{table}
    \centering
    \caption{ablation Study of three modules on FUSAR-Map1.0 dataset. 'FE-Mask decoder' is feature enhanced classwise mask decoder, 'ViT adapters' is Vision Transformer image encoder with multiple adapters and 'Task specific input' represents the module of task-specific input with low frequency of on SAR images }
    \label{tab:fusar_map1_abl}
    \begin{tabular}{ccccc}
        \toprule
        FE-Mask decoder & ViT adaptor & Task specific input & mIOU & OA \\
        \midrule
        & & & 38.28 & 61.78 \\
        \checkmark & & & 49.45 & 72.98 \\
        & & \checkmark & 57.42 & 79.7 \\
        & \checkmark & & 61.12 & 81.72 \\
        \checkmark & & \checkmark & 56.59 & 79.80 \\
        & \checkmark & \checkmark & 59.95 & 81.97 \\
        \checkmark & \checkmark & & 60.30 & 81.53 \\
        \checkmark & \checkmark & \checkmark & \pmb{61.48} & \pmb{82.14} \\
        \bottomrule
    \end{tabular}
\end{table}

\begin{table}
    \centering
    \caption{ablation Study of three modules on FUSAR-Map2.0 dataset. 'FE-Mask decoder' is feature enhanced classwise mask decoder, 'ViT adapters' is Vision Transformer image encoder with multiple adapters and 'Task specific input' represents the module of task-specific input with low frequency of on SAR images }
    \label{tab:fusar_map2_abl}
    \begin{tabular}{ccccc}
        \toprule
        FE-Mask decoder & ViT adaptor & Task-specific input & mIOU & OA \\
        \midrule
        & & & 18.1 & 41.52 \\
        \checkmark & & & 19.0 & 40.9 \\
        & \checkmark & & 25.6 & 53.09 \\
        & & \checkmark & 26.29 & 55.93 \\
        \checkmark & \checkmark & & 31.55 & 64.37 \\
        & \checkmark & \checkmark & 32.43 & 63.68 \\
        \checkmark & & \checkmark & 33.92 & 65.5 \\
        \checkmark & \checkmark & \checkmark & \pmb{36.03} & \pmb{67.67} \\
        \bottomrule
    \end{tabular}
\end{table}

(1)	Classwise mask encoder with featured enhancement module: the proposed classwise mask decoder is necessary to implement pixel-level landcover classification as a semantic segmentation task. In order to provide premise feature for improving segmentation performance, we introduce a feature enhancement structure on the classwise mask decoder. The ablation study is conducted on both FUSAR-Map1.0 and FUSAR-Map2.0 datasets to explore the effect of the feature enhancement module by adding and removing it. The results illustrate that the feature enhancement module significantly improve the model performance, especially on the harder dataset FUSAR-Map2.0, since more useful semantic information in the feature map is provided to the classification head. 

(2)	ViT adapters in Transformer blocks of image encoder: light-weighted adapters are widely utilized in Transformer fine-tuning task. Considering the characteristic gap between SAR images and natural scene images of SA-1B dataset, exhaustive adapters are inserted to all blocks of Vision Transformer by both series and parallel connection structure. Our experiments verify that ViT adapters play an extremely essential role on the whole architecture of model and enhance the segmentation performance in the landcover classification task on SAR images.

(3)	Task specific input of SAR image low frequency information: The MLP-based feature fusion module of low frequency information provides rich SAR semantic feature to image encoder by 2D Fast Fourier Transform processing on SAR images. Only conducting this module brings similar effect improvements compared to the ViT adapter module only, while the segmentation performance reaches to the best level on both FUSAR-Map1.0 and FUSAR-Map2.0 datasets in our experiments when employing both two modules. 

Through analyzing the ablation study results, we observe that the model achieves its best performance when the three designed modules are used in conjunction. However, the performance of using individual modules varies slightly depending on the dataset. For simpler datasets, the proposed CWSAM can achieve reasonable performance with fewer modules, whereas more complex and challenging tasks require a comprehensive combination of all modules.

\section{Discussions}
For parameter efficient fine-tuning task, there is a trade-off between the quantity of learnable parameters and model performance. Some uncertainty influences still exist depending on specific tasks and datasets. For relatively easy tasks or with a small number of training samples, a small number of learnable parameters can ensure reliable performance. Increasing the number of tuning parameters does not enhance accuracy and can lead to overfitting, where the model excels on training data but fails on new or test data. Overfitting indicates that the model may be memorizing the training data rather than generalizing from it.

On the other hand, increasing the number of parameters can improve the model's capacity to fit complex data. It can help the model capture intricate patterns and relationships.For some difficult tasks such as FUSAR-Map2.0 datasets, increasing the quantity of learnable parameters significantly improves the model performance while requires more computing resources. 

Besides, refer to the Table IX, compared with traditional segmentation networks, the proposed method requires fewer computing memories with fewer learnable parameters and performs better generalization and stability, as it can achieve best segmentation performance on both FUSAR-Map1.0 and FUSAR-Map2.0 datasets while other state-of-the-art algorithms scarcely have satisfactory performance on all data distributions. 

\begin{table}
\centering
\caption{The comparison of the total model parameters, fine-tuning parameters and resource usage of the proposed method and other state-of-the-art algorithms}
\begin{tabular}{ccccc}
\toprule
Pretrained & Backbone & Total  & Learnable & Training \\
 & & params(M) & params (M) & memory\\
\midrule
Deeplabv3 & ResNet50 & 204.43 & 204.43 & 17G \\
FE-Deeplabv3 & ResNet50 & 232.74 & 232.74 & 20G \\
Deeplabv3plus & ResNet50 & 130.88 & 130.88 & 18G \\
HRNet-18 & HRNet-18 & 29.26 & 29.26 & 12G \\
HRNet-48 & HRNet-48 & 197.94 & 197.94 & 14G \\
segformer & MIT-B0 & 11.20 & 11.20 & 10G \\
segformer & MIT-B3 & 133.96 & 133.96 & 24G \\
CWSAM & SAM-BASE & 104.96 & 15.29 & 10G \\
\bottomrule
\end{tabular}
\end{table}

\section{Conclusion}
In this paper, a parameter efficient fine-tuning method named ClassWise-SAM-Adapter (CWSAM) of visual foundation model is proposed to achieve semantic segmentation and landcover classification in SAR image domain. A light-weighted classwise mask decoder is designed to convert the category agnostic of Segment Anything Model (SAM) to distinguishable pixel-level classification module. Adapters are explored in Vision Transformer image encoder to shrink the domain gap between SA-1B natural scenes and SAR images, and task specific input module of low frequency information injects SAR image features into image encoder for improving the segmentation performance. The comprehensive experiments on multiple SAR image landcover classification datasets demonstrate the effectiveness of the proposed method. Our algorithm outperforms other state-of-the-art semantic segmentation methods and utilizes bits of computing resources for fine-tuning visual foundation model. Besides, to our best knowledge, the proposed method is the first trial of leverage Segment Anything Model to achieve semantic segmentation and landcover classification downstream tasks in SAR images by adapter-tuning. It proves the potentiality of foundation model in SAR image interpretation and applications.

\bibliographystyle{IEEEtran}
	\bibliography{reference}
 
\vspace{11pt}

\begin{IEEEbiography}[{\includegraphics[width=1in,height=1.25in,clip,keepaspectratio]{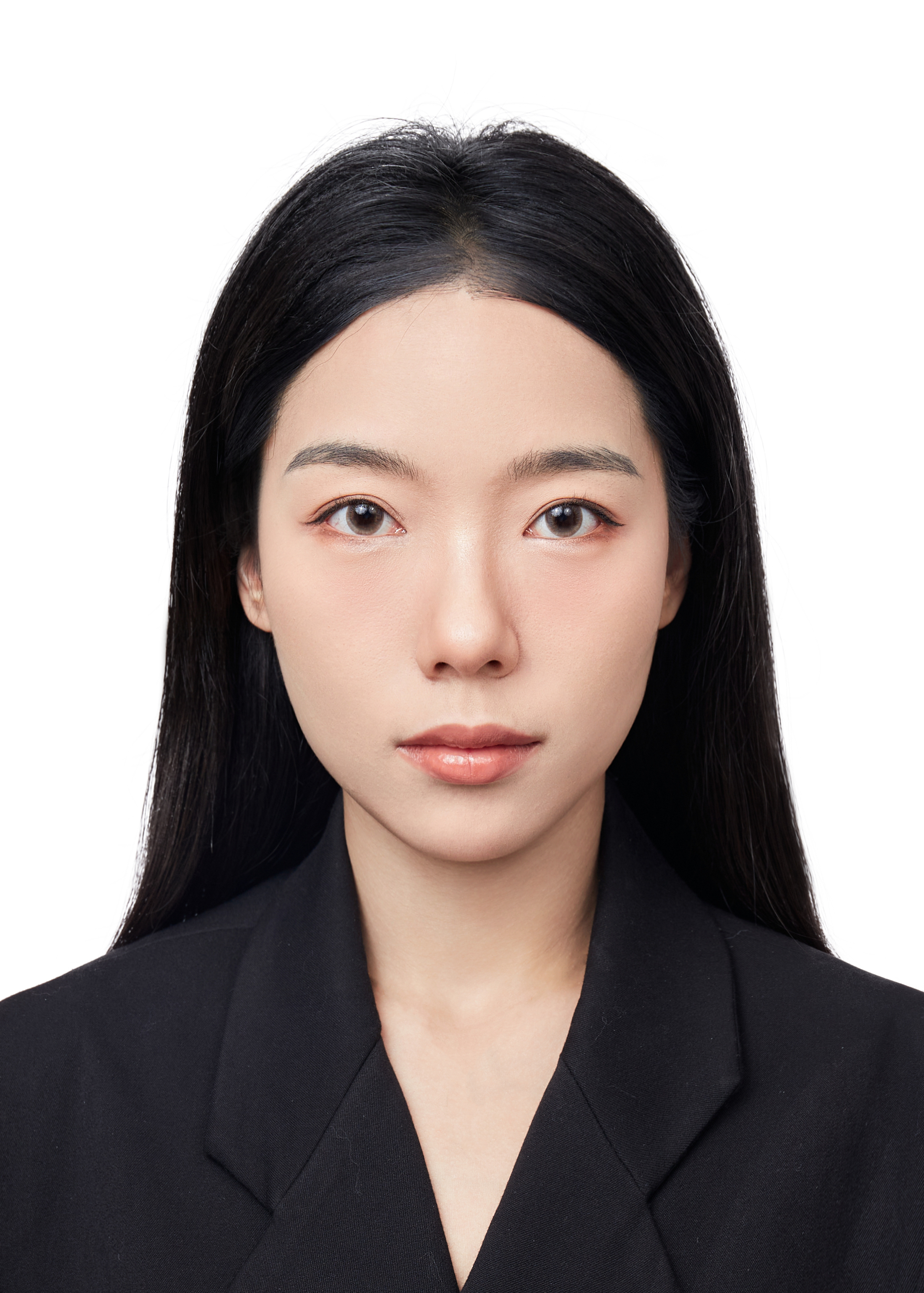}}]{Xinyang Pu} (Graduate Student Member, IEEE) received the B.E. degree in telecommunication engineering from Sun Yat-sen University, Guangzhou, China, in 2020. She is currently pursuing the Ph.D. degree in electronic science and technology with the Key Laboratory for Information Science of Electromagnetic Waves, School of Information Science and Technology, Fudan University, Shanghai, China.

	Her research interests include deep learning in remote sensing applications, intelligent interpretation of SAR images and computer vision.
\end{IEEEbiography}

\vspace{11pt}

\vfill

\end{document}